\documentclass{article}
\pdfoutput=1
\usepackage[utf8]{inputenc}
\usepackage[T1]{fontenc}
\usepackage[margin=0.75in]{geometry}
\usepackage{listings}
\usepackage{hang}
\usepackage{changepage}
\usepackage[russian,USenglish]{babel}
\usepackage{courier}
\usepackage{csquotes}
\usepackage{fancyhdr}
\usepackage{newclude}
\usepackage{gensymb}
\usepackage{subcaption}
\usepackage{float}
\usepackage{wrapfig}
\usepackage{amsmath}
\usepackage{mathtools}
\usepackage{amssymb}
\usepackage{url}
\usepackage{booktabs}
\usepackage{pbox}
\usepackage{hyphenat}
\usepackage{longtable}
\include*{tugboatHyphenation}
\usepackage{multicol}
\usepackage{tabularx}
    \newcolumntype{L}{>{\raggedright\arraybackslash}X}

\usepackage{amsthm}
\newtheorem{definition}{Definition}[]{}

\usepackage{tikz}
\usepackage{tikz-qtree}
\usetikzlibrary{positioning}
\usetikzlibrary{automata}
\usetikzlibrary{shapes,arrows}
\usetikzlibrary{plotmarks}
\tikzstyle{block} = [draw, fill=white, rectangle, 
    minimum height=3em, minimum width=6em]
\def\circledarrow#1#2#3{ 
\draw[#1,->] (#2) +(80:#3) arc(80:-260:#3);
}

\newenvironment{indent1}{%
	\begin{adjustwidth}{1cm}{}
  }{%
	\end{adjustwidth}
  }
\newenvironment{exDesc}{
\noindent
  }{

  }

\newenvironment{clist}		
{ \begin{itemize}
  \setlength\itemsep{-0.4em}
}
{ \end{itemize}
}

\newcommand{\vout}{\small \textbf}
\newcommand{\sub}{\textsubscript}
\newcommand{\nullKS}{[\enspace]}				
\newcommand{\symI}{\ensuremath{\mathbb{I}}}	
\newcommand{\symC}{\ensuremath{\mathbf{C}}}	
\newcommand{\symB}{\ensuremath{\mathbf{B}}}	
\newcommand{\symS}{\ensuremath{\mathbb{S}}}	
\newcommand{\symK}{\ensuremath{\mathbf{K}}}	
\newcommand{\SIGMA}{\ensuremath{\Sigma}}
\newcommand{\SIGMABAR}{\ensuremath{\overline{\Sigma}}}
\newcommand{\DELTA}{\ensuremath{\Delta}}
\newcommand{\PATMINP}{\ensuremath{min_{points}}}
\newcommand{\PATMINO}{\ensuremath{min_{occurs}}}
\newcommand{\DOTS}{\ensuremath{\dots}}

\newcommand{\vp}[1]{\ensuremath{#1}}			

\newcommand{\keywords}[1]{\textbf{\textit{keywords: }} #1}

\lstdefinelanguage{v5}
{
}

\title{Representing and Using Knowledge \\ with the Contextual Evaluation Model}
\author{Victor E Hansen \\ veh@alum.mit.edu}
\date{\today}

\begin{document}
\maketitle
\begin{abstract}
This paper introduces the Contextual Evaluation Model (CEM), a novel method for knowledge representation and manipulation. The CEM differs from existing models in that it integrates facts, patterns and sequences into a single contextual framework. V5, an implementation of the model is presented and demonstrated with multiple annotated examples. The paper includes simulations demonstrating how the model reacts to pleasure/pain stimuli. The \textit{thought} is defined within the model and examples are given converting thoughts to language, converting language to thoughts and how \textit{meaning} arises from thoughts. A pattern learning algorithm is described. The algorithm is applied to multiple problems ranging from recognizing a voice to the autonomous learning of a simplified natural language. 
\end{abstract}
\keywords{knowledge representation, artificial intelligence, context, natural language}
\section{Introduction}

The Contextual Evaluation Model (CEM) is an expansion of Knowledge Representation (KR) models. KR has a long history going back to the beginning of artificial intelligence in the late 1950's and early 1960's with IPL and LISP and continuing into the present with projects such as Cyc and OWL \cite{gps,lisp,wiki:Cyc,cycSyntax,owl,owl2}. In all these models `knowledge' is represented by `facts' with the goal of using the facts in non-trivial ways. A classic example is taking two facts `All  men are mortal' and `Socrates is a man' and deducing that `Socrates is mortal'.
\par It is our view that KR is the foundation upon which \textit{intelligence} is constructed. For the purposes of this paper, \textit{intelligence} is broadly defined as an ongoing process by which an \textbf{intity}\label{def-intity}\footnote{The word `intity' is a contraction of `\underline{int}elligent' and `ent\underline{ity}'. It will be used in this paper to refer to a natural or artificial intelligent entity.} transitions from one moment to the next in such a manner that supports that intity's ability to thrive. These moment-to-moment decisions of what-to-do-next are determined by both the intity's knowledge and the changes to the intity's physical environment as detected by the intity's senses. Intelligence, as so defined, can be encapsulated by an intelligence function~(\symI\label{def-I}) as shown in equation (\ref{equIfunction}). It maps the intity's state (\vp{q_m})\label{def-m} and sensory input (\vp{s_m}) at moment\footnote{The symbol $t$ while usually used to denote time will be later used to represent a \textit{thought}. To avoid confusion, the symbol $m$ will be used to denote a moment in time.} (\vp{m}) to a new a new state (\vp{q_{m+1}}) at moment \vp{m+1}. Function \symI \ is then reevaluated at moment \vp{m+1} with new sensory input (\vp{s_{m+1}}) to get state \vp{q_{m+2}},~\DOTS.
\begin{equation}\label{equIfunction}
\symI(q_{m},s_{m}) \rightarrow q_{m+1} \Longrightarrow \symI(q_{m+1},s_{m+1}) \rightarrow q_{m+2} \Longrightarrow \dots
\end{equation}
In (\ref{equIfunction}) the state encompasses all the know-how, expertise, physical capabilities, actions, etc. required by the intity to exist and thrive. The sensory input reflects a snapshot of the intity's current sensory experience of the physical world. KR, in this framework, is the combination of  the state and sensory input ($q$ and $s$). Given this, the paper investigates two questions:
\begin{enumerate}
\setlength\itemsep{-0.4em}
\item How should the $q$ and $s$ parameters of \symI \ be modeled?
\item What is function \symI \ and how would intelligence arise from its repetitive evaluation?
\end{enumerate}
The first half of this paper is devoted to explaining the Contextual Evaluation Model. The model consists of five components: the point, key set, binding, context and the contextual evaluation operation. By the end of the first half, the reader will be familiar with the components of the model and how the CEM relates to the intelligence function (\symI). Section~\ref{secCEM} of this paper introduces the components the CEM. Section~\ref{secV5LanEng} describes the V5 engine, a virtual machine that implements the CEM. Various examples are presented illustrating important features of the model. Section~\ref{secSeq} describes sequences: the ordering of points and actions. Sequences that sing a song, learn a maze and implement a Turing machine are presented. In section~\ref{secPatterns} an algorithm for learning patterns is given and tested against real world data in an example that recognizes voices. Section~\ref{secMotivate} discusses the modeling of pleasure and pain within the CEM/V5 framework.
\par Singing a song and running a maze, while interesting examples, do not make a convincing argument that the CEM shows intelligence for anything but the simplest forms of intelligence. Demonstrating that the CEM is a fitting model for all forms of intelligence is a daunting task; especially considering that there are no universally accepted tests for intelligence\cite{wiki:Turing_test}. Instead, the remainder of the paper focuses on one form of intelligence that is unique to humans: thought and how it manifests as language. The second half of this paper investigates these questions:
\begin{enumerate}
\setlength\itemsep{-0.4em}
\item What is a \textit{thought} and how is it represented with the CEM?
\item What is the relationship between thought and language?
\item How can an intity \textit{learn} a language? 
\end{enumerate}
Chapter~\ref{secThoughtsLanMean} defines the \textit{thought} and how thoughts relate to language and meaning. Section~\ref{secThought2Lan} describes how thoughts are converted to language. Section~\ref{secLanLearn} deconstructs the language acquisition process into four steps: recognizing words, grounding words, syntax acquisition and thought-to-thought meaning. Each of these four steps is shown to be a variation of pattern learning as previously described in section~\ref{secPatterns}.
\par Appendix~\ref{secGlossary} is a glossary and index of terms used in this paper. Appendix~\ref{secV5Stuff} contains tables detailing V5 commands, registers and instruction set. Appendixes \ref{datapatrecex} and \ref{secParseopRASM} contain supplemental material referenced within the paper.
\section{The Contextual Evaluation Model}\label{secCEM}
The Contextual Evaluation Model (CEM\label{def-CEM}) is a novel method for storing, retrieving and manipulating four classes of data:
\begin{clist}
\item Factual data describing an attribute or component of something such as the name of a person, the temperature of an oven, the reason for an action.
\item Pattern data that collectively can be used to recognize something.
\item Sequence data that describe an ordered collection such as the letters in a word, the notes in a song, the steps required to achieve a goal.
\item Contextual data that influence the interpretation of the above three classes of data.
\end{clist}
The basics of the model and the contextual representation of facts are presented in this section. Sequences and patterns are covered in sections~\ref{secSeq} and \ref{secPatterns} following the introduction of requisite preliminary material.
\subsection{Points, Key Sets, Bindings}
\begin{definition}{}
The \textbf{point}\label{def-point} is the atomic unit of the CEM. A point represents something with no restriction as to what that something can be.
\end{definition}
A point can represent an object, an idea, a number, a quality. It can represent something specific such as a particular person or an abstract concept such as love. A point can represent a class or category of things (dogs) or a specific instance of that class (your dog). A point can represent sensory input: a sound, visual input, an odor. A point may also trigger an action or movement, the generation of a sound, the transmission of a signal.
\begin{definition}{}
Points created by external senses are called \textbf{sensory}\label{def-sensory point}  points while points that trigger external actions are \textbf{control points}\label{def-control point}. All other points are \textbf{internal}\label{def-internal point} points. 
\end{definition}
A point, other than representing something, in and of itself, conveys no additional information about that something (e.g. a point representing your dog does not convey any additional information about your dog or even that it is a dog).
\begin{definition}{}
A \textbf{key set}\label{def-keyset} is a set of zero or more points. A key set is denoted as a list of points enclosed in brackets, [\vp{point_1} \vp{point_2} \DOTS \vp{point_n}]. The \textbf{null key set}\label{def-null keyset} (\nullKS) is an empty key set.
\end{definition}
\begin{definition}{}
A \textbf{binding}\label{def-binding} is a key-value or attribute-value relationship between a non-null key set and a value set (of points).
\end{definition}
A binding, $\mathbf{b}$, is a 3-tuple:
\begin{equation}
\mathbf{b} = (\left\{ k^i \right\}_{i=1}^n, \left\{ v^i \right\}_{i=1}^{n'}, w)
\end{equation}
where $\mathbf{b}_k$ is a key set, $\mathbf{b}_v$ is a set of value points and $\mathbf{b}_w$ is the binding weight.
\begin{definition}{}
The binding \textbf{weight}\label{def-w} is proportional to the number of points in the key set ($|\mathbf{b}_k|$), i.e. the greater the number of points in the key set the greater the binding weight.
\end{definition}
When a value set consists of a single point then $\mathbf{b}_v$ will represent $\mathbf{b}_v^1$. A binding is notated as $keyset = value$: [\vp{point_1} \vp{point_2} \DOTS \vp{point_n}] = \vp{point_{value}}. The binding weight is normally automatically assigned using implementation dependent parameters.
\subsection{Evaluation and Contextual Evaluation}
\begin{definition}{}
A binding set (\symB\label{def-B}) is a set of bindings.
\end{definition}
Initially, only binding sets will be considered having the restriction that no two binding key sets are identical within the set\footnote{This restriction will be relaxed in subsequent sections.}:
\begin{equation}
\forall \mathbf{b}^i, \mathbf{b}^j \in \mathbf{B} \ \textrm{if} \ \mathbf{b}^i_k = \mathbf{b}^j_k \ \textrm{then} \ i = j 
\end{equation}
A simple evaluation is a function that given a binding set (\symB) and key set (\symK) searches the binding set for a binding with the identical key set and returns that binding's value:
\begin{equation}
E_s(\mathbf{B}, \mathbf{K}) = \mathbf{b}_v \ where \ \mathbf{b} = \forall \mathbf{b}^i \in \mathbf{B} \ \textrm{if} \  \mathbf{b}^i_k = \mathbf{K} \ \textrm{then} \ \mathbf{b}^i  
\end{equation}
\begin{definition}{}
A \textbf{context} (\symC\label{def-C}\label{def-context}) is a separate dynamic set of points.
\end{definition}
How points are inserted into the context and how they are removed from the context is implementation dependent. A more precise definition of the context is provided in sections~\ref{moreoncontext} and \ref{secEngine}. Equation (\ref{equ-CE}) defines a contextual evaluation.
\begin{equation}\label{equ-CE}
E_c(\mathbf{B}, \symC, \mathbf{K}) = \mathbf{b}_v \ \textrm{where} \ \mathbf{b} =  \arg\max_{\mathbf{b} \in \mathbf{B}} \left\{ \mathbf{b}_w \ | \  \mathbf{b}_k \in ( \mathbf{K} \cup \symC ) \ \textrm{and} \ \mathbf{K} \in \mathbf{b}_k  \right\}
\end{equation}
and this simplifies to (\ref{equEvalNullKS}) when \symK \ is the null key set:
\begin{equation}\label{equEvalNullKS}
E_c(\mathbf{B}, \symC, \mathbf{K}) = \mathbf{b}_v \ \textrm{where} \ b = \arg\max_{\mathbf{b} \in \mathbf{B}} \left\{ \mathbf{b}_w \ | \  \mathbf{b}_k \in \symC \right\}
\end{equation}
For an analogy, consider a binding as locking a value with its key set points. The value cannot be obtained without all the keys. A simple evaluation is a search through \symB \ attempting to unlock a binding using all the keys in $\mathbf{K}$. A contextual evaluation searches for a binding with the greatest binding weight (largest number of locks) that is unlocked using all the keys in \symK \ and as many additional key points from the context \symC \ as is necessary. For both $E_s$ and $E_c$, if no binding is found the operation fails and the result is undefined.
\begin{definition}{}
Going forward, \textbf{evaluating a key set} ($\mathbf{K})$ means evaluating $E_c(\mathbf{B}, \symC, \mathbf{K})$ where $\mathbf{B}$ and \symC \ are assumed.
\end{definition}
\subsection{Examples}
The statement `the boiling point of water is $100\degree$ Celsius' could be represented with the binding [\vp{p_1} \vp{p_2} \vp{p_3}]=\vp{p_v} where \vp{p_1} represents the attribute of the boiling point of something, \vp{p_2} represents water, \vp{p_3} represents the Celsius scale and \vp{p_v} represents $100\degree$. Greater clarity and understanding is obtained by replacing symbolic points (\vp{p_x}) with descriptive labels. The same example becomes [\vp{boilingPoint} \vp{water} \vp{celsius}]=100. Descriptive labels may optionally be enclosed in double quotes. Below are some binding examples:
\begin{clist}
\item[] `Chocolate tastes good' - [\vp{chocolate} \vp{tastes}]=\vp{good}
\item[] `John thinks chocolate tastes bad' - [\vp{johnThinks} \vp{chocolate} \vp{tastes}]=\vp{bad}
\item[] `The meaning of life is 42 according to the Hitchhiker's Guide to the Galaxy' - [\vp{meaningLife} \vp{HHGuide}]=42
\item[] `the factorial of 5 is 120' - [\vp{factorial} 5]=120
\item[] `representation of the number 22 is ``22''' - [\vp{representation} 22] = `22'
\item[] `representation of the number 22 in base 2' - [\vp{representation} 22 \vp{base2}] =`'10110'
\end{clist}
Given the above bindings and an empty context (\symC = $\varnothing$) then the contextual evaluation of [\vp{chocolate} \vp{tastes}] results in \vp{good}. If the context contains the point \vp{johnThinks} then the evaluation of the same key set, [\vp{chocolate} \vp{tastes}], results in \vp{bad}. In this second evaluation, both bindings ([\vp{chocolate} \vp{tastes}]=\vp{good} and [\vp{johnThinks} \vp{chocolate} \vp{tastes}]=\vp{bad}) satisfy the constraints but since the second binding has a higher binding weight (more points in the binding key set) it is selected.
\subsection{Complete vs Incomplete Contextual Evaluations}
The greater the number of points in a binding key set, the greater the binding weight and the greater the specificity of the binding. Your pet dog Fido might have multiple descriptions. In general if asked what Fido is, you would reply `a dog'. If you met your biology teacher in the park and he or she asked you would reply `a beagle'. But if your biology professor asked you in biology class you would reply `canis lupis familiaris'. Bindings describing these possible answers might be:
\begin{clist}
\item[] [\vp{whatIs} \vp{Fido}]=\vp{dog}
\item[] [\vp{whatIs} \vp{Fido} \vp{profKnowItAll}]=\vp{beagle}
\item[] [\vp{whatIs} \vp{Fido} \vp{profKnowItAll} \vp{biologyLecture}]=\vp{CanisLupusFamiliaris}
\end{clist}
In each instance, the same question, `What is Fido?', is answered by evaluating the same key set [\vp{whatIs} \vp{Fido}]. However, the context varies in each instance. If the context contains none of the points in this example the result is \vp{dog}. But if you meet your professor outside of school  (\vp{profKnowItAll} $\in$ \symC) then the answer would be \vp{beagle} and if you met in biology class (\vp{profKnowItAll}, \vp{biologyLecture} $\in$ \symC) then it would be \vp{CanisLupusFamiliaris}.
\par A \textbf{complete}\label{def-complete} evaluation is when the points in the key set are an identical match for the points in the binding key set (i.e. no points in the context were used). An \textbf{incomplete}\label{def-incomplete} evaluation is where one or more context points are required in finding a matching binding. The ability to perform incomplete evaluations allows for tremendous flexibility. Examples throughout this paper will demonstrate this flexibility.
\subsection{Point Variants and Twines}\label{def-variant}
\begin{definition}{}
Every internal point has two \textbf{variants} notatated as: \vp{point.i} and \vp{point.v}.
\end{definition}
The first variant is for representing class membership or is-a membership, i.e. `x is a y' as in `an apple is a fruit'. The second  variant is for representing a quality of or attribute of something, i.e. `the x of y is z' as in `the color of the apple is red'.
\begin{definition}{}\label{def-twine}
A \textbf{twine} is a binding with a variant point in its key set. Only one point of a binding's key set may be a variant. 
\end{definition}
\par Twines have an optional abbreviated notation. A simple is-a relationship is shown in~(\ref{equEqvIsaTwine}a). The syntax in~(\ref{equEqvIsaTwine}b) is used when other conditional/contextual points (\vp{c_i}) are included:
  \begin{subequations}\label{equEqvIsaTwine}
    \begin{align}
      \vp{p'}<\vp{p} \  &\textrm{is equivalent to} \ [\vp{p.i}] = \vp{p'} \\
      \vp{p'}<\vp{p}|\vp{c_1},\vp{c_2},\DOTS \  &\textrm{is equivalent to} \ [\vp{p.i} \ \vp{c_1} \ \vp{c_2} \DOTS] = \vp{p'}
    \end{align}
  \end{subequations}
A similar notation is used for value-of twines. (\ref{equEqvValTwine}a) is the basic notation, (\ref{equEqvValTwine}b) is with additional points.

  \begin{subequations}\label{equEqvValTwine}
    \begin{align}
     \vp{p}>\vp{p'} \ &\textrm{is equivalent to} \ [\vp{p.v}] = \vp{p'} \\
     \vp{p}>\vp{p'}|\vp{c_1},\vp{c_2},\DOTS \  &\textrm{is equivalent to} \ [\vp{p.v} \ \vp{c_1} \ \vp{c_2} \DOTS] = \vp{p'}
    \end{align}
  \end{subequations}

If two points are doubly twined (\vp{p}>\vp{p'} and \vp{p}<\vp{p'}) then the notation in (\ref{equEqvBothTwine}a/b) can be used:


  \begin{subequations}\label{equEqvBothTwine}
    \begin{align}
     \vp{p}:\vp{p'} \  &\textrm{is equivalent to} \ [\vp{p.v}] = \vp{p'} \ \textrm{and} \ \ [\vp{p'.i}] = \vp{p} \\
     \vp{p}:\vp{p'}|\vp{c_1},\vp{c_2},\DOTS \  &\textrm{is equivalent to} \ [\vp{p.v} \ \vp{c_1} \ \vp{c_2} \DOTS] = \vp{p'} \ \textrm{and} \ \ [\vp{p'.i} \ \vp{c_1} \ \vp{c_2} \DOTS] = \vp{p}
    \end{align}
  \end{subequations}

As a general rule, the is-a twine is used to represent relationships and the value-of twine is evaluated to `retrieve' a relationship value. For example, given \vp{color}<\vp{red} and \vp{red}<\vp{ball} (red is a color and the ball is red) then evaluating [\vp{color.v} \vp{ball}] results in \vp{red} (the color of the ball is red).
\subsection{Time}
Time is an integral component of knowledge representation and should be included in an AI knowledge representation data model. With a couple of minor changes, time and the flow of time can easily and elegantly be represented in the CEM. The first change creates a time point with an associated magnitude that increments at a constant rate (e.g. every 100 milliseconds). The time point is always in the context. Secondly, the contextual evaluator ($E_c$) is modified such that bindings containing a time point will match if the time point in context has a magnitude equal to or greater than the time point in the binding. Thirdly, the binding weight of a binding containing a time point is based, in part, on the magnitude of the time point. Given two bindings each containing a time point and having identical other points then the binding with the greater time point would have the greater binding weight.
\par In the example below time is notated as m(\vp{n}) for  the moment in time with magnitude~\vp{n}.
\begin{lstlisting}[numbers=none]
maritalStatus>single|John,m(100)
maritalStatus>married|John,m(200)
maritalStatus>divorced|John,m(300)
maritalStatus>remarried|John,m(400)
\end{lstlisting}
With a current time of \vp{m(1000)} in the context the evaluation of [\vp{maritalStatus.v} \vp{John}] would return \vp{remarried}. All four bindings would be potential matches but the last one with the largest time magnitude has the greatest binding weight. With time \vp{m(350)} in the context evaluating [\vp{maritalStatus.v} \vp{John}] returns \vp{divorced}.
\subsection{More on Context and Evaluation of is-a Twines}\label{moreoncontext}
The context consists of explicit and implicit points. The explicit points are those directly inserted into the context. The implicit points are obtained by evaluating the is-a twine for each point in the context. The results of successful evaluations are additionally added to the context. These is-a context points are linked to the base (explicit) point such that if the base point is removed from the context then any/all is-a points are also removed.
\par While most contextual evaluations have a single result (the binding with the highest weight), the is-a evaluations return all valid binding results not just the is-a binding with the highest weight. The exception to this is when multiple potential values differ only in the time point (\vp{m}). In these cases the single binding with the highest time point is selected.
\par
\begin{figure}[H]
\centering
\begin{subfigure}[b]{0.3\textwidth}
\centering
\begin{tikzpicture}[scale=0.8, grow'=up]
\node[] { \{ };
\begin{scope}[xshift=1cm,yshift=-1mm]
\Tree [.\vp{Harry}  [.\vp{man} \vp{human} ] \vp{married} \vp{father} ]
\end{scope}
\begin{scope}[xshift=3.25cm]
\node[] {, \enspace $m(1000)$ \enspace \}};
\end{scope}
\end{tikzpicture}
\caption{Explicit context}\label{figexpcon}
\end{subfigure}
\begin{subfigure}[b]{0.6\textwidth}
\centering
\{ \vp{Harry} , \vp{man} , \vp{human} , \vp{married} , $m(1000)$ , \vp{father} \}
\caption{Explicit \& implicit context}\label{figimpcon}
\end{subfigure}
\caption{Explicit and implicit context points}\label{figimpexpcon}
\end{figure}
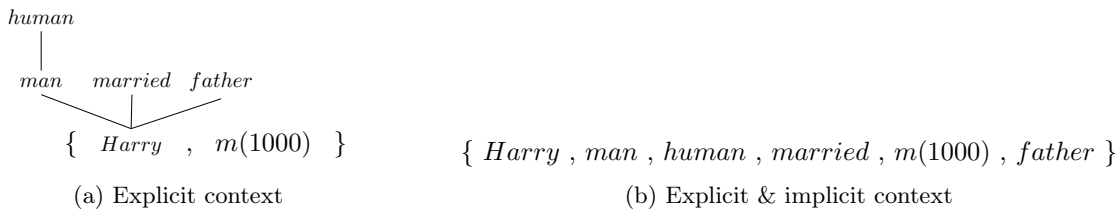

 Figure~\ref{figexpcon} shows two explicit points in the context: \{\vp{Harry},$m(1000)$\}. Point \vp{Harry} has the following is-a twines represented as a tree\footnote{In the tree, nodes are points and branches are is-a twines. See section~\ref{secThoughts} for more on tree representations.}: \vp{man}<\vp{Harry}, \vp{human}<\vp{man}, \vp{married}<\vp{Harry}|$m(100)$, \vp{single}<\vp{Harry}|$m(50)$ and \vp{father}<\vp{Harry}. The full context (explicit + implicit) is as shown in figure~\ref{figimpcon}.
\subsection{Relating the Intelligence Function (\symI) to the CEM}
The introduction posed two questions: what is the intelligence function and what are its arguments? Within the framework of the CEM, the intelligence function is the contextual evaluation function $E_c$ (equation~\ref{equ-CE}). The sensory inputs ($s$) are the sense points injected into the context. The state ($q$) is a combination of the binding set and the non-sensory points in the context.
\par
\begin{figure}[H]
\centering
\begin{tikzpicture}[auto, node distance=3cm,>=latex']
\node [block](bs){$\symB = q$};
\node [block, right of=bs](ce){$\symI = E_c()$ };
\node[cloud, cloud puffs=12, cloud ignores aspect, align=center, draw, right of=ce, minimum width=2cm, minimum height=2cm] (context) {$\symC = s + q$};
\node [above of=context, node distance=1.75cm](sense) {sensory input ($s$)};
\node [below of=context, node distance=1.75cm](control){control output};
\node [below of=ce, node distance=1.5cm] (eval) {eval \nullKS};
\circledarrow{}{eval}{.75cm};
\draw [<->] (bs.east) -- (ce.west);
\draw [<->](ce.east) -- (context.west);
\draw [line width=.1cm,->,color=black!40](sense.south) -- (context.north);
\draw [line width=.1cm,->,color=black!40](context.south) -- (control.north);
\end{tikzpicture}
\caption{The CEM - \symI \enspace Relationship}\label{figce-i-relation}
\end{figure}
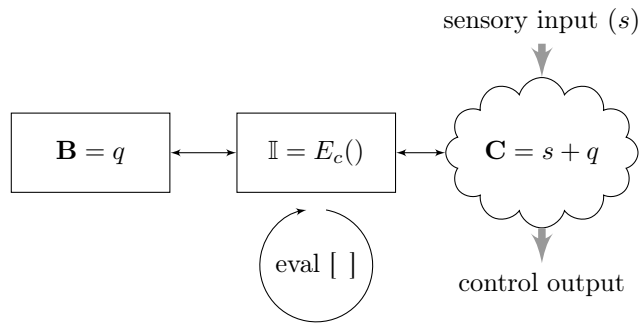

 Figure~\ref{figce-i-relation} illustrates how an intity would utilize the CEM. Knowledge would be stored as bindings within the binding set (\symB). The context (\symC) is both referenced and updated by the contextual evaluation process. The context would also be updated periodically by the intity's sensory input. Context control points dictate any external actions taken by the intity. Each evaluation of the null key set is the transition from one moment to the next.
Remaining to be demonstrated is how the CEM can effectively compute intelligent behavior. This will be accomplished with the help of the V5 language/engine and working examples.
\section{The V5 Language and Engine}\label{secV5LanEng}
V5 is an experimental language and contextual evaluation engine. The low level language is similar to a machine language in that it consists of primitive opcodes that operate within a virtual machine. The machine is accessed through a command interpreter. The components of V5 are its point sets, the opcodes for the virtual machine and a suite of interpreter commands for defining points, loading point sets, running the interpreter and of course debugging. It is written in C and comprises approximately 6,000 lines of code.
\par The CEM consists solely of points. The sole data type of V5 is the point. There are no integers, floating point numbers or character strings and correspondingly no instructions that operate on those data types. There are only instructions to manipulate points, create bindings/twines and evaluate bindings. The minimalist design is intentional to explore the potential of the CEM.
\par Points in V5 can be dynamically created using instructions within V5 or points can be declared through the command interpreter using the `def' command. This command creates a new point and associates a text label to the point. Every point has an internal reference number. When a point is output to the console, V5 checks to see if that point has a label. If so then the label is output. If the point has no label then the point is output as $\#nnn$ where \vp{nnn} is the hexadecimal value of the point's reference number. Variants have `.v' or `.i' suffix. Surrogate points (see section~\ref{def-surrogate}) have a `?' suffix.
\par Many code examples will be presented. In these examples, comments are enclosed in /* \DOTS */. V5 output is displayed in boldface. Every line within an example is numbered starting with line 1. Examples may be interspersed with descriptive text. In this case the line numbers continue. 
\begin{figure}[H]
\begin{lstlisting}[xleftmargin=2cm]
/* this is a comment */
ps a,b,c,d
\end{lstlisting}
\begin{exDesc}
\begin{adjustwidth}{2cm}{}This is descriptive text in the middle of the example\end{adjustwidth}
\end{exDesc}
\begin{lstlisting}[xleftmargin=2cm,firstnumber=3]
/*	The example continues at this line (3). Note that the output from the print queue (PQ)
	(line (*\ref{line:expqout}*)) includes the current time moment (`20' in this example)
*/
run
(*\vout{PQ(20): an example of V5 output}\label{line:expqout}*)
\end{lstlisting}
\caption{V5 example conventions}
\end{figure}
\subsection{The Engine}\label{secEngine}
\begin{figure}[H]
\begin{center}
\begin{tikzpicture}[auto, node distance=3cm,>=latex']

    \node [block, name=as] {AS};
    \node [block, name=ss, below of=as, node distance=1cm] {MR};
    \node [block, name=ps, below of=ss, node distance=1cm]{PS/\symC};
    \node [block, name=pq, below of=ps, node distance=1cm] {PQ};
    \node [block, name=tc, below of=pq, node distance=1cm] {TC};
    \node [block, name=ppu, right of=ps]{PPU};
    \node [block, name=ce, right of=ppu]{E\sub{C}/\symI};
    \node [block, name=bs, above of=ce, node distance=2cm]{BS/\symB};
    \node [block, name=ci, below of=ce, node distance=2cm]{CI};
    
    \draw [-] (1.5,-4)--(1.5,0);
    \draw [-] (1.5,0)--(as.east);
    \draw [-] (1.5,-1)--(ss.east);
    \draw [-] (ps.east)--(ppu.west);
    \draw [-] (1.5,-3)--(pq.east);
    \draw [-] (1.5,-4)--(tc.east);
   
    \draw [-] (ci.west) -| (ppu.south);
    \draw [-] (ppu.east)--(ce.west);
    \draw [-] (ci.north)--(ce.south);
    \draw [-] (bs.west) -| (ppu.north);
    \draw [-] (bs.south)--(ce.north);

\end{tikzpicture}
\caption{Block Diagram of V5 Engine}\label{bdofv5}
\end{center}
\end{figure}
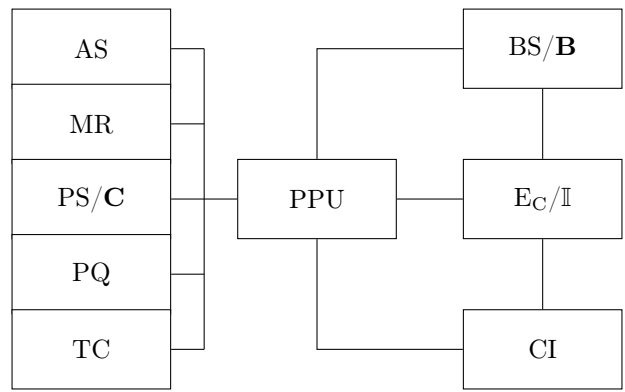

Figure~\ref{bdofv5} shows a block diagram of the V5 engine and its relationship to the intelligence function (\symI) and contextual evaluation function. The left-hand side consists of several register sets. These hold sets of points for various V5 instructions.
\begin{definition}{}
The point set (\textbf{PS})\label{def-PS} serves a dual function as a push down stack and as the context for the engine.
\end{definition}
The context consists of all the points in the PS plus all is-a points off of these points, plus all is-a points off of the is-a points$\dots$ as explained in section~\ref{moreoncontext}. The is-a points for a point \vp{p} are determined by evaluating the is-a variant of that point ([\vp{p}.i]) and then repeating for each resulting point of the evaluation. This continues until the evaluation fails (e.g. there are no more is-a points). Each evaluation is contextual and uses the points in the context at that instance. This is intentionally chaotic. Any given initial set of PS points may result in one of any number of different contexts due to the ordering and timing of the evaluations of all the possible is-a points. The aggregation set (\textbf{AS}\label{def-AS}) holds points used by the aggregate set reduce  (opRAS and opRASM) instructions. Section~\ref{secASReduce} further describes the AS. The print queue (\textbf{PQ}\label{def-PQ}) holds a set of points for output to the console. With the exception of debugging features, the PQ and related instructions are the only output mechanism in V5. The twine context (\textbf{TC}\label{def-TC}) is a set of points that are automatically inserted into a twine binding when the twine is created. The miscellaneous register set (\textbf{MR}\label{def-MR}) is a collection of special registers updated with the running of the engine. A list of these can be found in appendix~\ref{secregisters}.
\par The point-processing-unit (\textbf{PPU}\label{def-PPU}) executes V5 instructions. The contextual evaluator (\textbf{E\sub{C}}\label{def-EC}) performs all contextual evaluations as requested from the PPU. The \textbf{BS} is the binding set for the engine. All bindings are maintained within the BS. Bindings are created and inserted into BS by the PPU. The command interpreter \textbf{CI}\label{def-CI} handles the interface via the console between the V5 engine and the user.
\subsubsection{Operation}
After the engine has been initialized with the necessary points and bindings the `run' command is given to the command interpreter. This initiates the following:
\begin{enumerate}
\item Scan the points in the PS from top to bottom.
\item If the point is an opcode then remove it from the PS and execute it.
\item If the point is a register then replace it with the register's current value.
\item If the point is a value variant then evaluate the key set ([\vp{point}.v]). If the evaluation succeeds then remove the variant point and push the resulting value(s) onto the PS and continue with step \#1.\footnote{Except if the result is a single point null. In this case nothing is pushed onto the PS.} If any point pushed onto the PS is already in the PS then that instance is removed so that only the instance at the top of the PS remains. If the evaluation fails then leave the value variant unchanged in the PS.
\item Leave all other points in the PS unchanged.
\item Stop if there are no remaining instructions or value variants in the PS otherwise continue with step \#1.
\end{enumerate}
\subsection{Commands and the Instruction Set}
The engine interface is command driven. All user input begins with a command followed by command arguments. A list of V5 commands is found in appendix~\ref{secV5Commands}.
\par V5 is an experimental system and many instructions have been implemented for various experiments. Appendix~\ref{secopcodes} lists only the instruction opcodes used in the examples in this document. All instruction opcodes begin with lower case `op' and end with an uppercase mnemonic. If an opcode ends with a lower case `t' then it is identical to the opcode without the suffix. The suffix indicates that trace debugging output is to be performed when the opcode is executed. For example, the instruction opEVAL constructs and evaluates a key set while the instruction opEVALt performs the same action plus outputs trace information to the console. The naming convention for the processing registers (MR) is lower case `r' followed with an upper-case mnemonic. For example, the opNEW instruction creates a new point and updates register rNEW with that new point. A list of the registers is found in appendix~\ref{secregisters}.
\subsubsection{Aggregate Set (AS) Reduction}\label{secASReduce}
The purpose of the AS is to perform evaluations on multiple contiguous pairs of points. It is used primarily for parsing natural language to thoughts and generating natural language from thoughts. The aggregate set (AS) may be loaded with a set of points using the opLASM instruction. The opcode opRASM performs the reduction operation as follows:
\begin{enumerate}
\item Starting at the left side of the AS, add two contiguous points (plus any is-a points) to the PS and evaluate the null key set. If the evaluation succeeds then temporarily hold the result along with the binding weight of the matched binding.
\item Move right one point and repeat the null key set evaluation again holding a success off to the side.
\item When all contiguous pairs have been evaluated then select the held binding with the highest binding weight. In the case of multiple bindings choose the left most one.
\item If any of the points in the binding are prefaced with a minus sign then remove the corresponding point from the AS.
\item Execute the value of the selected binding. This means to take all value points except for the last, insert into the PS and execute. Take the last value point and insert it into the AS at the point of the matched points (unless it is the \vp{null} point).
\end{enumerate}
\par There are extended binding forms that can be used while performing AS reduction. Take the binding found in the example below:
\begin{lstlisting}[numbers=none]
bind +2 [-np/-noun] eoa,noun,noun.v,opEVAL,np.v,rEVAL,opTWISA,np.v
\end{lstlisting}
The weight of a binding is normally determined by the number of points in the binding key set. The `+2' in the above example adds 2 to the weight of the binding giving it a higher precedence over other two-point bindings. The use of the minus sign before a point (-\vp{np}, -\vp{noun}) indicates that the point is to be removed from the AS if this binding is matched. The slash means that the ordering of the points matters. In this example an \vp{np} point followed by a \vp{noun} point would match. The reverse would not.
\par Normally a binding (other than value twine) is only allowed to have one value point. During AS reduction the multiple value points are pushed onto the PS a point at a time. If a point is an opcode then it is immediately executed. Note that pushing the value points from left to right onto a stack effectively reverses the ordering of the arguments.
\par The evaluation of [\vp{p.v}] is a little different. The AS and all of its is-a points is first searched. If \vp{p} is located then [\vp{p.v}] is taken as the root point.
\par After all value points have been handled, the last point (top of the PS) is pulled from the PS. If it is not the \vp{null} point then it is inserted into the AS at the point of the match.

\subsection{Evaluation of [\vp{p.v}] with Respect to the PS}
The evaluation of a value variant ([\vp{p.v}]) within V5 is not treated strictly as a CEM evaluation. V5 first examines the is-a points in the PS for an instance of \vp{p}. If found then the value of [\vp{p.v}] is taken as the root value of that is-a tree in the PS as the value. Otherwise the evaluation is performed as a normal CEM evaluation. Note that this only applies with the evaluation of [\vp{p.v}] where \vp{p.v} is the only point of the key set. For example, if the PS contains \{\vp{a},\vp{b},\vp{c}\} and \vp{p}<\vp{c} then the evaluation of [\vp{p.v}] results in \vp{c} (\vp{c} is-a an instance of \vp{p} within the current context).
\subsection{Simple Examples}
This example is about horses and four specific horses. Three of the horses: Pegasus, Mr. Ed and Seabiscuit\footnote{Pegasus is a mytholgical winged horse\cite{pegasus}, Mr. Ed is a talking horse featured on a TV series of the same name\cite{wiki:Mister_Ed}, Seabiscuit is a famous race horse\cite{seabiscuit}.} are defined as horses on line~\ref{line:exHorses}. The example examines whether or not horses can talk and fly. In general, they cannot as defined on line~\ref{line:exhorse1}.  But Mr. Ed is able to talk (\ref{line:exhorse2}) and Pegasus can fly (\ref{line:exhorse3}). Seabiscuit, while being very fast, can neither talk nor fly.
\begin{lstlisting}
def horse, Pegasus, MrEd, Seabiscuit
def canTalk, canFly
def yes, no

twine horse<Pegasus,MrEd,Seabiscuit(*\label{line:exHorses}*)

twine canTalk>no|horse ; canFly>no|horse(*\label{line:exhorse1}*)
twine canTalk>yes|MrEd,horse(*\label{line:exhorse2}*)
twine canFly>yes|Pegasus,horse(*\label{line:exhorse3}*)
\end{lstlisting}
\begin{exDesc}
The evaluations below test the horses for special qualities.
\end{exDesc}
\begin{lstlisting}[firstnumber=10]
eval [canFly.v Seabiscuit]
(*\vout{    result = no}*)
eval [canTalk.v MrEd]
(*\vout{    result = yes}*)
eval [canFly.v Pegasus]
(*\vout{    result = yes}*)
eval [canTalk.v Pegasus]
(*\vout{    result = no}*)
\end{lstlisting}
\begin{exDesc}
A new point is defined for Secretariat\footnote{A more recent racehorse without any special abilities other than to win races.\cite{wiki:Secretariat_(horse)}.}. The evaluation on line~\ref{line:exSecFail} fails. Why? Because the only knowledge about whether or not something can talk applies to horses and only if Secretariat is declared as a horse (line~\ref{line:exSecHorse}) will the evaluation succeed (line~\ref{line:exSecOK}).
\end{exDesc}
\begin{lstlisting}[firstnumber=18]
def Secretariat
eval [canTalk.v Secretariat](*\label{line:exSecFail}*)
(*\vout{ ? No result found}*)
Twine horse<Secretariat(*\label{line:exSecHorse}*)
eval [canTalk.v Secretariat](*\label{line:exSecOK}*)
(*\vout{    result = no}*)
\end{lstlisting}
\section{Sequences}\label{secSeq}
\begin{definition}{}
A \textbf{sequence}\label{def-sequence} is collection of bindings that define an ordering of points or a set of steps to accomplish a goal.
\end{definition}
A sequence may be \textit{tightly coupled} where the \textit{next} point of the sequence directly follows the prior. Examples of tightly coupled sequences are: the letters of the alphabet; the letters in a word; the words in a sentence; the digits of $\pi$; the instructions in a computer program. Sequences that are \textit{loosely coupled} are a collection of bindings defining steps/actions where the \textit{next} step does not necessarily directly follow the current but from some other triggering context. Loosely coupled sequences are also known as \textit{goals}. An example would be the goal of driving from home to the grocery store.
\par Sequences, being solely a collection of bindings do nothing in and of themselves. There is no necessary correlation between the ordering of bindings in a sequence and the resulting ordering of the sequence points. Performing a sequence requires an executing engine. Multiple sequences can be running concurrently. The following examples demonstrate how two concurrent sequences interact to print the spelling of a word. A step within a sequence may itself be the start of another sequence. And everything is contextual. There is tremendous flexibility in the definition and execution of sequences in the ever changing context of the real world.
\subsection{Implementing and Running Sequences}
There are many ways to represent sequences within the CEM. Suppose we want a sequence for the spelling of `book'. The first example below shows a sequence of four points (\vp{sb1}, \vp{sb2}, \vp{sb3}, \vp{sb4}). The twined value of \vp{sb1} are the two points "b" and \vp{sb2.v} (line~\ref{line:sb1line}). Evaluating [\vp{sb1.v}] results in both "b" and \vp{sb2.v} being added to the PS. The V5 engine sees \vp{sb2.v}, evaluates it and replaces it with "o" and \vp{sb3.v}. This continues through \vp{sb4}. Its evaluation, [\vp{sb4.v}], is only "k" so after adding "k" to the PS the run is finished. The result is that the PS contains the points "k", "o" and "b". Recall that a given point may only occur once in the PS. That is why only one instance of "o" (line~\ref{line:exBookSeq1}) is given. This sequence iterates through the spelling of `book', but not in a particularly useful way.
\begin{lstlisting}
def sb1, sb2, sb3, sb4
twine sb1>"b",sb2.v(*\label{line:sb1line}*)
twine sb2>"o",sb3.v
twine sb3>"o",sb4.v
twine sb4>"k"
ps sb1.v
run
show ps
(*\vout{ps: "k", "o", "b"}\label{line:exBookSeq1}*)
\end{lstlisting}
The second example outputs the spelling of `book'. The evaluation starts with \vp{sb1.v} in the PS. It is evaluated and the letter `b' is added to the print queue (line~\ref{line:exsb1}). It continues until [\vp{sb4}.v] is evaluated and `k' is added to the print queue and the contents of PQ are output to the console. The problem with this sequence is that it is only useful for spelling book and that the spelling is only output to the console. The spelling cannot be referenced in any other way.
\begin{lstlisting}
def sb1, sb2, sb3, sb4(*\label{line:exsb1}*)
twine sb1>"b",opADDPQ,sb2.v
twine sb2>"o",opADDPQ,sb3.v
twine sb3>"o",opADDPQ,sb4.v
twine sb4>"k",opADDPQ,opOUTPQ
ps sb1.v
run
(* \vout{PQ(2): "b" "o" "o" "k"} *)
\end{lstlisting}
The third example in this series demonstrates a more generalized and useful approach by having two sequences. One sequence is defined for the spelling of `book'. Then another sequence is defined that takes the first sequence (or any similar sequence) and outputs the spelling of the word to the console. The sequence points for the spelling are \vp{b1} through \vp{b4}. They are linked through value twines (line~\ref{line:spellword}). The sequence works/runs as follows: [\vp{spell} `book'] is bound to the first point (\vp{b1}) of the sequence (line~\ref{line:exBookSeq2a}). The evaluation of [\vp{b1.v}]) gives the next point (\vp{b2}) of the sequence. Each of the points \vp{b1} through \vp{b4} has an is-a twine of the corresponding letter. (V5 automatically defines the letters of the alphabet and twines each letter to the \vp{letter} point: \vp{letter}<`a', \vp{letter}<`b', \DOTS). Therefore the evaluation of [\vp{letter.v}] with \vp{b1} in the context results in `b', with \vp{b2} in the context `o', etc.
\begin{lstlisting}
def word,"book",spell,pEnd
twine word<"book"
def b1,b2,b3,b4
bind [spell "book"] b1(*\label{line:exBookSeq2a}*)
twine b1>b2 ; b2>b3 ; b3>b4 ; b4>pEnd(*\label{line:spellword}*)
twine "b"<b1 ; "o"<b2 ; "o"<b3 ; "k"<b4
twine spell<b1,b2,b3,b4
\end{lstlisting}
\begin{exDesc}
The second sequence is defined by points \vp{s} through \vp{s4}. The start of the sequence evaluates [\vp{spell} \textit{word}] to get the starting sequence point for the spelling of the word. Line~\ref{line:spellisas} uses the opcode opPSISAS to re-evaluate all of the is-a points in the PS so that the evaluation of \vp{letter.v} results in the letter for the current step in the word sequence. That letter is added to the print queue. Line~\ref{line:spellnext} attempts to get the next sequence point of the word [\vp{spell.v}]. If the evaluation results in the next sequence point of the word the opVAL instruction converts the point to its value variant which is then evaluated by the V5 engine. If that result is another \vp{b_n} point then \vp{s4.v} evaluates to \vp{s2.v}. If it is \vp{pEnd} then the twine on line~\ref{line:spelldone} is run, the spelling is output and the sequence terminates.
\end{exDesc}
\begin{lstlisting}[firstnumber=8]
def s,s2,s3,s4
twine s>word.v,spell,eoa,opEVAL,rEVAL,s2.v
twine s2>opPSISAS,letter.v,opADDPQ,s3.v(*\label{line:spellisas}*)
twine s3>spell.v,opVAL,s4.v(*\label{line:spellnext}*)
twine s4>s2.v
twine s4>opOUTPQ | pEnd(*\label{line:spelldone}*)
ps "book",opPSISAS,s.v
run
(*\vout{PQ(2): "b" "o" "o" "k"}*)
\end{lstlisting}
\begin{exDesc}
The next section shows how context can be used to add a plural ending to a word. The spelling for `spy' is defined just as `book' was above using points \vp{t1}, \vp{t2} and \vp{t3}. At line~\ref{line:spellplural} an alternative value for \vp{t2.v} is given in the context of \vp{plural}. Instead of linking to \vp{t3} (`y') it links to \vp{pEnd2}. The value of \vp{s4} is also redefined in the context of \vp{plural} and \vp{pEnd2} to add `ies' to the PQ and then output the PQ to the console. Line~\ref{line:spellies} show the output for spelling the word `spy' with the addition of plural to the context.
\end{exDesc}
\begin{lstlisting}[firstnumber=17]
twine word<"spy"
def t1,t2,t3
bind [spell "spy"] t1
twine t1>t2 ; t2>t3 ; t3>pEnd
twine "s"<t1 ; "p"<t2 ; "y"<t3
twine spell<t1 ; spell<t2 ; spell<t3

def pEnd2,"ies"
twine t2>pEnd2 | plural(*\label{line:spellplural}*)
twine s4>"ies",opADDPQ,opOUTPQ | pEnd2,plural
ps "spy",plural,opPSISAS,s.v
run
(*\vout{PQ(2): "s" "p" "ies"}\label{line:spellies} *)
\end{lstlisting}
\subsection{Singing a Song}
Singing the first few notes of the song `Mary had a little lamb' is the purpose of this example. Currently, V5 output is limited to console text so the singing is emulated by the outputting of pitch and lyrics with the proper timing. The example uses future time points to control the timing of the console output. If the PS contains a value variant and the twine defining the value contains a future moment (time) point (\vp{p}>\vp{value}|\vp{m}(\textit{future-time})) then it will remain (unevaluated) in the PS until the current time is equal to or greater than the \textit{future-time} in the twine binding.
\par This first section below defines the basic points of the song and song sequence. The points \vp{n1} through \vp{n13} are for the first 13 notes of `Mary had a little lamb'. Line~\ref{line:marynotes} twines all the notes to the note point. Line~\ref{line:marynoteseq} links each note to the subsequent note ($note_n.v$ $\rightarrow$ $note_{n+1}$). Line~\ref{line:marynoteword} associates the words/lyrics with each of the notes.
\begin{lstlisting}[firstnumber=1]
def song,lyrics,note,songEnd
twine songEnd>null
def MarySong,Mar,ee,had,a,lit,tle,lamb
twine song<MarySong
twine lyrics<Mar,ee,had,a,lit,tle,lamb
def n1,n2,n3,n4,n5,n6,n7,n8,n9,n10,n11,n12,n13
twine note<n1,n2,n3,n4,n5,n6,n7,n8,n9,n10,n11,n12,n13(*\label{line:marynotes}*)
twine n1>n2 ; n2>n3 ; n3>n4 ; n4>n5 ; n5>n6 ; n6>n7 ; n7>n8 ;(*\label{line:marynoteseq}*)
  n8>n9 ; n9>n10 ; n10>n11 ; n11>n12 ; n12>n13 ; n13>songEnd
twine Mar<n1 ; ee<n2 ; had<n3 ; a<n4 ; lit<n5 ; tle<n6 ; lamb<n7 ;(*\label{line:marynoteword}*)
  lit<n8 ; tle<n9 ; lamb<n10 ;  lit<n11 ; tle<n12 ; lamb<n13
twine MarySong>n1	/* n1 is first point in sequence */
\end{lstlisting}
\begin{exDesc}
The timings of the notes are specified below. The timing point is twined to different numbers of opINCT instructions based on the tempo and note type (\vp{half} or \vp{qtr}/quarter notes). Line~\ref{line:marytime} twines \vp{half} or \vp{qtr} to each of the notes. The opINCT instruction increases the time point within the twine context set (TC).
\end{exDesc}
\begin{lstlisting}[firstnumber=13]
def timing,tempofast,temposlow,qtr,half
twine timing>opINCT,opINCT,opINCT|qtr,tempofast
twine timing>opINCT,opINCT,opINCT,opINCT,opINCT,
  opINCT,opINCT,opINCT,opINCT,opINCT|half,tempofast
twine timing>opINCT,opINCT,opINCT,opINCT,opINCT,
  opINCT|qtr,temposlow
twine timing>opINCT,opINCT,opINCT,opINCT,opINCT,
  opINCT,opINCT,opINCT,opINCT,opINCT,opINCT,opINCT,
  opINCT,opINCT,opINCT,opINCT,opINCT,opINCT,opINCT,
  opINCT|half,temposlow
twine qtr<n1 ; qtr<n2 ; qtr<n3 ; qtr<n4 ; qtr<n5 ; qtr<n6 ; half<n7 ;(*\label{line:marytime}*)
  qtr<n8 ; qtr<n9 ; half<n10 ; qtr<n11 ; qtr<n12 ; half<n13
\end{lstlisting}
\begin{exDesc}
Similarly, pitches are associated with each note. Each note now has two is-a twines: one for timing and one for pitch. The value (\textit{note}.v) of each note links to the next note. Pitch is represented with a letter-number pair. Each letter (A-G) represents a note on the scale and the number corresponds to the octave above middle C. For example, the note B1 represents the B above middle C (C0).
\end{exDesc}
\begin{lstlisting}[firstnumber=25]
def pitch,a0,b0,c0,d0,e0,f0,g0,a1,b1,c1,d1,e1,f1,g1
twine pitch<a0,b0,c0,d0,e0,f0,g0,a1,b1,c1,d1,e1,f1,g1
twine b1<n1 ; a1<n2 ; g0<n3 ; a1<n4 ; b1<n5 ; b1<n6 ; b1<n7 ;
  a1<n8 ; a1<n9 ; a1<n10 ; b1<n11 ; d1<n12 ; d1<n13
\end{lstlisting}
\begin{exDesc}
Below sets up another sequence to iterate through all the notes of a song (points \vp{swt} through \vp{swt5}). Line~\ref{line:maryfirstnote} gets the first note of the sequence. This line also creates a new point to be used to distinguish one execution of this sequence from another. Line~\ref{line:maryfirstlp} gets the lyric and pitch associated with the current note and places each into the print queue. Then it outputs the queue and clears it. On line~\ref{line:marynextnote}, \vp{note.v} results in the note, opVAL converts the note to its value variant. The evaluation of the value variant gives the next note. The next/new note is saved in working memory register 1 (opLWM1). If the next note is \vp{songEnd} then the next step is at line~\ref{line:maryend}, otherwise at line~\ref{line:marycontinue} where \vp{swt5} is twined to \vp{swt2.v} to continue with the next note. This twine contains a future time point. Evaluations fail until the future time is reached (twine \vp{swt5}>\vp{swt2.v}|\vp{instancepoint,m(future),note}).
\end{exDesc}
\begin{lstlisting}[firstnumber=29]
def swt,swt2,swt3,swt4,swt5
twine swt>song.v,opVAL,opNEW,rNEW,swt2.v(*\label{line:maryfirstnote}*)
twine swt2>opPSISAS,lyrics.v,opADDPQ,pitch.v,opADDPQ,opOUTPQ,opCLRPQ,swt3.v(*\label{line:maryfirstlp}*)
twine swt3>opRCTX,timing.v,note.v,opVAL,opLWM1,rWM1,swt4.v(*\label{line:marynextnote}*)
twine swt4>opPSISAS,rWM1,opACTX,rNEW,opACTX,swt5,@swt2.v,opTWVAL,swt5.v(*\label{line:marycontinue}*)
twine swt4>opSTATE,songEnd,opVAL|songEnd(*\label{line:maryend}*)
\end{lstlisting}
\begin{exDesc}
All the components of the sequences are now ready to run. The song is started by placing the song point, \vp{MarySong}, into the PS along with the tempo point \vp{tempSlow} and the starting point for the sequence (\vp{swt.v}). The output below shows the current moment (time) point enclosed in parentheses. The time delay between lyric-note output can be determined by taking the difference between two moment points. The delay between outputs is 6 (quarter notes) until `lamb' (half notes) with a delay is 20. At the end of the sequence the opSTATE instruction (line~\ref{line:maryend}) results in a debugging output line (\ref{line:marystateout}).
\end{exDesc}
\begin{lstlisting}[firstnumber=35]
ps MarySong,opPSISAS,temposlow,swt.v
run
(*\vout{PQ(2): Mar b1}*)
(*\vout{PQ(8): ee a1}*)
(*\vout{PQ(14): had g0}*)
(*\vout{PQ(20): a a1}*)
(*\vout{PQ(26): lit b1}*)
(*\vout{PQ(32): tle b1}*)
(*\vout{PQ(38): lamb b1}*)
(*\vout{PQ(58): lit a1}*)
(*\vout{PQ(64): tle a1}*)
(*\vout{PQ(70): lamb a1}*)
(*\vout{PQ(90): lit b1}*)
(*\vout{PQ(96): tle d1}*)
(*\vout{PQ(102): lamb d1}*)
(*\vout{state: ps: songEnd, opVAL, \#338, temposlow}\label{line:marystateout}*)
\end{lstlisting}

\subsection{Running a Maze}
\begin{figure}[H]
\centering
\begin{tikzpicture}[scale=0.4, every node/.style={scale=0.75}]
\draw[very thick] (3,0)--(3,2)--(1,2)--(1,1)--(0,1)--(0,4)--(1,4)--(1,3)--(6,3)--(6,5)--(5,5)--(5,6)--(9,6)--(9,7);
\draw[very thick] (4,0)--(4,2)--(6,2)--(6,2)--(7,2)--(7,5)--(9,5)--(9,4)--(10,4)--(10,7);
\draw[below] (3.5,0) node(s){start};
\draw[above] (9.5,7) node(f){finish};
\draw[above] (3.5,2) node(n1) {n1};
\draw[above] (0.5,2) node(n2) {n2};
\draw [above] (6.5,2) node(n3) {n3};
\draw [above] (6.5,5) node(n4) {n4};
\draw [above] (9.5,5) node(n5) {n5};
\end{tikzpicture}
\caption{Maze modeled in example}\label{figmaze}
\end{figure}
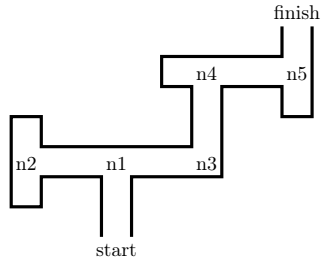

A V5 sequence for running a simple T-maze is presented in this next example. The maze, shown in figure~\ref{figmaze}, is represented in V5 as a tree. 
The section below defines the layout of the maze. In particular, the bindings on lines~\ref{line:mazedef1}-\ref{line:mazedef2} define the linked right-hand and left-hand nodes at any given node. If a node has no left or right binding then going that direction at that node leads to a dead end.
\begin{lstlisting}
def node,success,isSuccess, n1,n2,n3,n4,n5, direction,left,right,nextDirection
twine node<n1,n2,n3,n4,n5
bind [success isSuccess] success
bind [nextDirection left] right
bind [n1 left] n2(*\label{line:mazedef1}*)
bind [n1 right] n3
bind [n3 left] n4
bind [n4 right] n5
bind [n5 left] success(*\label{line:mazedef2}*)
\end{lstlisting}
\begin{exDesc}
The maze running strategy is to begin with the left-hand side of a node (turn left). If that fails then try the right-hand side. If that fails then backtrack to the prior node and continue. If the maze is considered a tree then running the maze begins with the left most leaf and moves clockwise from leaf to leaf. Eventually it will hit the successful leaf and exit the maze.
\par The value of \vp{pos} (\vp{pos.v}) is used to track the position within the maze. The time point (\vp{m}) is included in the twine as the value will be changing. The twine on line~\ref{line:mazepos1} sets the initial position at node 1. Bindings will be created to keep track of the direction taken at a node [turn \textit{node}] and to keep track of the prior node [prior \textit{node}]. The latter is for use in backtracking.
\end{exDesc}
\begin{lstlisting}[firstnumber=10]
def pos, turn, prior
twine pos>n1|rCTP(*\label{line:mazepos1}*)
\end{lstlisting}
\begin{exDesc}
The logic for running the maze is below. Points \vp{m1} through \vp{m7} are for the main sequence loop, \vp{mf1} through \vp{mf3} are when a node fails, \vp{mb1} and \vp{mb2} are for backtracking to a prior node and \vp{ms1} through \vp{ms5} are for outputting the correct path on success. The first step is to determine the direction to turn at the current node (\vp{pos.v}). The binding [\textit{node} \vp{pos}] is evaluated. If it succeeds then continue with line~\ref{line:m2ok}, otherwise continue at line~\ref{line:m2fail} and perform the binding [\textit{node} \vp{pos}]=\vp{left} and start from the top. At line~\ref{line:m2ok} the register rEVAL has the direction to turn (left or right). Evaluate [\textit{node} \textit{direction}] to get the next node. If that fails continue at \vp{mf1}. If it succeeds and done with the maze then continue with \vp{ms1} otherwise set \vp{pos.v} to the new node (line~\ref{line:m6}) and continue back at \vp{m1}.
\end{exDesc}
\begin{lstlisting}[firstnumber=12]
def m1,m2,m2a,m3,m4,m5,m6,m7, mf1,mf2,mf3, mb1,mb2, ms1,ms2,ms3,ms4,ms5
twine m1>opINCCTP,pos.v,turn,eoa,opEVAL,m2.v	
twine m2>evalFail,opVAL,m2a.v|evalFail(*\label{line:m2fail}*)
twine m2a>pos.v,turn,eoa,left,opBIND,m1.v
twine m2>pos.v,rEVAL,eoa,opEVAL,m4.v(*\label{line:m2ok}*)
twine m4>evalFail,opVAL,mf1.v|evalFail
twine m4>rEVAL,opLWM1,isSuccess,rEVAL,eoa,opEVAL,m5.v
twine m5>ms1.v
twine m5>evalFail,opVAL,m6.v|evalFail
twine m6>rWM1,prior,eoa,pos.v,opBIND,m7.v(*\label{line:m6}*)
twine m7>pos,rWM1,opTWVAL,m1.v
\end{lstlisting}
\begin{exDesc}
The section below is executed when the evaluation of [\textit{node direction}] fails. It attempts to get the direction currently associated with the node (line~\ref{line:mf1} and evaluate [nextDirection \textit{direction}]. If that succeeds (i.e. [nextDirection left]=right) continue with \vp{mf3} and bind [turn \textit{node}] to the new direction and continue with \vp{m1}. If it fails then both directions failed at the node and backtracking needs to be done (\vp{mb1}).
\end{exDesc}
\begin{lstlisting}[firstnumber=23]
twine mf1>pos.v,turn,eoa,opEVAL,mf2.v(*\label{line:mf1}*)
twine mf2>rEVAL,nextDirection,eoa,opEVAL,mf3.v
twine mf3>pos.v,turn,rCTP,eoa,rEVAL,opBIND,m1.v
twine mf3>evalFail,opVAL,mb1.v|evalFail
\end{lstlisting}
\begin{exDesc}
Backtracking is simple. Evaluation of [\textit{node} \vp{prior}] gives the prior node (\vp{mb1}). Set the current position to this node (\vp{mb2}) and continue back at \vp{mf1}.
\end{exDesc}
\begin{lstlisting}[firstnumber=27]
twine mb1>pos.v,prior,eoa,opEVAL,mb2.v
twine mb2>pos,rEVAL,opTWVALt,mf1.v
\end{lstlisting}
\begin{exDesc}
The next portion of the sequence outputs the correct path through the maze. Set the current node (\vp{pos.v}) back to the first node (\vp{n1}). Evaluate the correct turn for the node, output the node and turn (\vp{ms3}). Then continue with the next node. If there is no next node then the traversal through the maze is complete and the sequence is finished.
\end{exDesc}
\begin{lstlisting}[firstnumber=29]
twine ms1>pos,n1,opTWVAL,ms2.v
twine ms2>pos.v,turn,eoa,opEVAL,ms3.v
twine ms3>opCLRPQ,pos.v,opADDPQ,rEVAL,opADDPQ,opOUTPQ,ms4.v
twine ms3>evalFail,opVAL|evalFail
twine ms4>pos.v,rEVAL,eoa,opEVAL,ms5.v
twine ms5>opINCCTP,pos,rEVAL,opTWVAL,ms2.v
\end{lstlisting}
\begin{exDesc}
Running the sequence is done by initializing the PS with the current position (\vp{pos.v}) to \vp{n1} and then running the V5 engine. The solution to the maze is then output.
\end{exDesc}
\begin{lstlisting}[firstnumber=35]
ps pos,n1,opTWVAL,m1.v
run
(*\vout{PQ(15): n1 right}*)
(*\vout{PQ(16): n3 left}*)
(*\vout{PQ(17): n4 right}*)
(*\vout{PQ(18): n5 left}*)
\end{lstlisting}
\subsection{A Turing Machine}\label{secTM}
This example constructs a Turing machine\cite{turingmachine} to demonstrate that V5 is Turing complete and that, in theory, any computable function can be implemented with V5.\footnote{This is not to suggest that V5 is suitable for general purpose computing\cite{wiki:Turing_tarpit}}. An overview of the example is below:
\begin{clist}
\item The infinite tape is implemented as a series of linked points as described in figure~\ref{figTMTape}. The tape, at any given point is finite in size. When the tape is positioned past the last tape cell on the right, a new tape cell is automatically allocated and appended to the tape.
\item The states of this (simple) machine are declared points: \vp{sInit} and \vp{sHalt}.
\item Actions are defined with bindings with \textit{sdcn} points as shown in figure~\ref{figTMStates}.
\item The machine operates much like any other Turing machine. It repeatedly reads the contents of the current cell, uses that value with the current state to determine the new tape cell contents, new tape position and new state. The machine halts when its current state is \vp{sHalt}.
\end{clist}
\par The V5 commands to define points will no longer be included in the examples. The `set autodef' command below instructs the V5 command interpreter to automatically define any new point it encounters.
\begin{lstlisting}
set autodef on
\end{lstlisting}
\begin{exDesc}
The machine runs as follows:
\begin{clist}
\item[] \vp{tm1}- get the contents of the current tape position in rEVAL
\item[] \vp{tm2} - eval [\textit{tape content} \textit{current state}] to get the next sdc (new state, tape direction, new tape contents) point
\item[] \vp{tm3} - twine \vp{curSDC}>\textit{new \vp{sdc}} (in rEVAL)
\item[] \vp{tm4} - eval [\vp{sdc} \vp{newTapeContent}] for new tape contents
\item[] \vp{tm5} - set the new tape contents
\item[] \vp{tm6} - evaluate [\vp{sdc} \vp{tapeMove}] to see which direction to move
\item[] \vp{tm7} - move the tape to \vp{tm8} on line~\ref{line:tm8ok} if the tape cell exists, or line~\ref{line:tm8fail} if the tape cell does not exist and needs to be created
\end{clist}
\end{exDesc}
\begin{lstlisting}[firstnumber=2]
twine tm1>curPos.v,contents,eoa,opEVAL,tm2.v
twine tm2>rEVAL,curState.v,sdc,eoa,opEVAL,tm3.v
twine tm3>curSDC,rEVAL,opTWVAL,tm4.v
twine tm4>curSDC.v,newTapeContent,eoa,opEVAL,tm5.v
twine tm5>contents,curPos.v,rCTP,eoa,rEVAL,opBIND,tm6.v
twine tm6>curSDC.v,tapeMove,eoa,opEVAL,tm7.v
twine tm7>curPos.v,rEVAL,eoa,opEVAL,tm8.v
\end{lstlisting}
\begin{exDesc}
\begin{clist}
\vspace{-15pt}
\item[] \vp{tm8} - twine \vp{curPos}>\textit{new tape position}
\item[] \vp{tm9} - eval [\textit{sdc} \vp{newState}] for the new \vp{sdc}
\item[] \vp{tm10} - twine \vp{curState}>\textit{new state}
\item[] \vp{tm11} - force an increment of the time point
\item[] \vp{tm12} - check to see if the new state is \vp{sHalt}, goto line~\ref{line:tm13halt} if yes, line~\ref{line:tm13cont} if not
\item[] \vp{tm13} - either resume loop at \vp{tm1} or leave with \vp{sHalt} in the PS if done
\end{clist}
\end{exDesc}
\begin{lstlisting}[firstnumber=9]
twine tm8>curPos,rEVAL,opTWVAL,tm9.v(*\label{line:tm8ok}*)
twine tm9>curSDC.v,newState,eoa,opEVAL,tm10.v
twine tm10>curState,rEVAL,opTWVAL,tm11.v
twine tm11>opINCCTP,tm12.v
twine tm12>curState.v,checkHalt,eoa,opEVAL,tm13.v
twine tm13>evalFail,opVAL,tm1.v|evalFail(*\label{line:tm13cont}*)
twine tm13>sHalt(*\label{line:tm13halt}*)
\end{lstlisting}
\begin{exDesc}
When a tape move to the right fails the section below is run.
\begin{clist}
\item[] \vp{tm8} - clear out the \vp{evalFail} point and create a new tape cell point
\item[] \vp{tm8a} - link the current last tape cell to the new point
\item[] \vp{tm8b} - left link the new cell to prior last cell
\item[] \vp{tm8c} - define \vp{moveNone} to remain on current new cell
\item[] \vp{tm8d} - set the contents of new cell to \vp{tBlank}
\item[] \vp{tm8e} - set the current tape position to the new cell
\item[] \vp{tm8f} - resume at \vp{tm9} above
\end{clist}
\end{exDesc}
\begin{lstlisting}[firstnumber=16]
twine tm8>evalFail,opVAL,opNEW,tm8a.v|evalFail(*\label{line:tm8fail}*)
twine tm8a>curPos.v,moveRight,eoa,rNEW,opBIND,tm8b.v
twine tm8b>rNEW,moveLeft,eoa,curPos.v,opBIND,tm8c.v
twine tm8c>rNEW,moveNone,eoa,rNEW,opBIND,tm8d.v
twine tm8d>rNEW,contents,eoa,tBlank,opBIND,tm8e.v
twine tm8e>curPos,rNEW,opTWVAL,tm8f.v
twine tm8f>tm9.v(*\label{line:tm8failb}*)
\end{lstlisting}
\begin{exDesc}
The points \vp{sdc1} and \vp{sdc2} are used to define the new state, tape move direction and new tape contents based on the current state and tape contents. Line~\ref{line:xxx1} defines the sdc (\vp{sdc1}) if state is \vp{SInit} and tape contents is \vp{tBlank}.
\begin{figure}[H]
\begin{center}
\begin{tabular}{  c c | c c c | c  }
\multicolumn{2}{c |}{Current} & \multicolumn{3}{|c|}{Operation} \\
State & Tape & State & Tape & Move & SDC \\
\hline \hline
sInit & tBlank & sInit & tX & none & sdc \\
sInit & tX & sHalt & tX & right & sdc2 \\
\end{tabular}
\end{center}
\caption{State Tables for Turing Machine}\label{figTMStates}
\end{figure}
\end{exDesc}
\begin{lstlisting}[firstnumber=23]
bind [sInit tBlank sdc] sdc1(*\label{line:xxx1}*)
bind [sdc1 newTapeContent] tX	
bind [sdc1 tapeMove] moveNone
bind [sdc1 newState] sInit

bind [sInit tX sdc] sdc2
bind [sdc2 newTapeContent] tX	
bind [sdc2 tapeMove] moveRight
bind [sdc2 newState] sHalt
\end{lstlisting}
\begin{exDesc}
The following lines define the initial tape. It consists of one cell containing \vp{tBlank}. Each cell is left and right linked to its previous/next cell. If there is no right link then that cell is last cell to the right (and a new cell will be automatically appended if necessary).
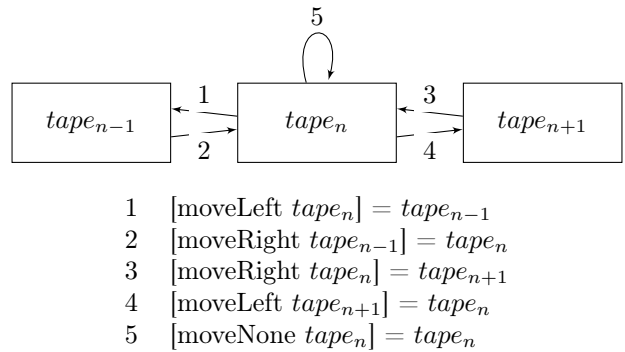
\begin{figure}[H]
\begin{center}
\begin{tikzpicture}[scale=0.5, auto, node distance=3cm,>=latex']
    \node [block, name=tapePrior] {$tape_{n-1}$};
    \node [block, name=tapeCur, right of=tapePrior] {$tape_n$};
    \node [block, name=tapeNext, right of=tapeCur] {$tape_{n+1}$};
    \draw [<-] (tapePrior.10pt)-- node[above, fill=white]{1} ([yshift=5pt] tapeCur.west);
    \draw [->] (tapePrior.-10pt)-- node[below, fill=white]{2} ([yshift=-5pt] tapeCur.west);
    \draw [<-] (tapeCur.10pt)-- node[above, fill=white]{3} ([yshift=5pt] tapeNext.west);
    \draw [->] (tapeCur.-10pt)-- node[below, fill=white]{4} ([yshift=-5pt] tapeNext.west);
    \path (tapeCur) edge [loop above] node {5} (tapeCur);
    \node [below of=tapePrior, node distance = 2cm, xshift=3cm] {
\begin{tabular}{ r l }
1 & [moveLeft $tape_n$] = $tape_{n-1}$ \\
2 & [moveRight $tape_{n-1}$] = $tape_n$ \\
3 & [moveRight $tape_n$] = $tape_{n+1}$ \\
4 & [moveLeft $tape_{n+1}$] = $tape_n$ \\
5 & [moveNone $tape_{n}$] = $tape_n$ \\
\end{tabular}
     };
\end{tikzpicture}
\end{center}
\caption{Layout of Tape in Turing Machine}\label{figTMTape}
\end{figure}
\end{exDesc}
\begin{lstlisting}[firstnumber=32]
bind [tPos1 contents rCTP] tBlank
bind [tPos1 moveLeft] tPos1
bind [tPos1 moveNone] tPos1
\end{lstlisting}
\begin{exDesc}
The sequence is now ready to execute. First define the current state as \vp{sInit} (line~\ref{line:tmcurstate}), then twine the current tape position to \vp{tPos1} (line~\ref{line:tmcurpos}), add the staring sequence point (\vp{tm1}) to the PS and run.
\end{exDesc}
\begin{lstlisting}[firstnumber=35]
ps curState,sInit,opTWVAL(*\label{line:tmcurstate}*)
run
ps curPos,tPos1,opTWVAL(*\label{line:tmcurpos}*)
run
ps tm1.v							/* Start running with tm1 */
run
show ps
(*\vout{ps: sHalt}*)
\end{lstlisting}
\subsection{Recapping Tightly Coupled Sequences}
All of the sequences described thus far have been tightly coupled sequences, i.e. the next point in a sequence is determined by the prior point. Defining tightly coupled sequences with contextual twines has several benefits:
\begin{indent1}
\begin{enumerate}
\item Multiple sequences can be \textit{running} simultaneously as demonstrated with the prior examples.
\item The order in which sequence twines are created is independent of the order in which they are \textit{executed}. This may become useful with the incremental learning of sequences.
\item All evaluations are contextual and that offers tremendous flexibility when running sequences in a variety of contexts.
\item The V5 engine is always doing \textit{next} whatever is most appropriate. It is not explicitly running a sequence as a traditional computer would run a subroutine or function.
\end{enumerate}
\end{indent1}
\subsection{Loosely Coupled Sequences (aka Goals)}
A loosely coupled sequence is a collections of bindings that collectively work to achieve a (long term) goal. Sequences that operate (apply) over long time periods and usually invoke other (sub)sequences are loosely coupled. An example would be the sequence/goal of grocery shopping. The steps to achieve this goal are:
\vspace{10pt}
\begin{indent1}
\begin{enumerate}
  \setlength\itemsep{-0.4em}
\item Get the shopping list
\item Go to the grocery store
\item Purchase the items on the list
\item Go home
\item Put the groceries away
\end{enumerate}
\end{indent1}
\vspace{10pt}
The twines and bindings to achieve this goal will differ from the twines used in the tightly coupled sequences. Patterns will be used instead of linked twine points. The V5 code below gives a possible example of how this might be done. The goal point used is \vp{doShopping}. Line~\ref{line:groceryStart} is the binding to start the shopping. It invokes the sequence \vp{getShoppingList}. The actions for this sequence would be defined separately and would depend on many factors such as, does the list already exist or not. The step in achieving the goal would be triggered by recognizing the pattern (see section~\ref{secPatterns}) [\vp{haveShoppingList} \vp{doShopping}]. When this is recognized the sequence \vp{gotoGroceryStore} is triggered. These patterns $\Rightarrow$ sub-action continue until the goal is satisfied.
\begin{lstlisting}
bind [start doShopping] getShoppingList.v(*\label{line:groceryStart}*)
bind [haveShoppingList doShopping] gotoGroceryStore.v
bind [atGroceryStore doShopping] purchaseItems.v
bind [haveGroceryItems doShopping] goHome.v
bind [atHome withGroceries doShopping] putAwayGroceries.v
bind [groceriesPutAway doShoping] shoppingGoalFinished.v
\end{lstlisting}
\section{Patterns}\label{secPatterns}
A pattern is a binding that is used to recognize something. In the CEM framework a pattern would consist primarily of sensory points: auditory to recognize a sound, visual to recognize an object, tactile to recognize a touch, etc. A pattern is represented as a binding: [\vp{r_1} \vp{r_2} \DOTS] = \vp{r_{thing}} where \vp{r_n} are the recognition descriptor points and \vp{r_{thing}} is the point that is recognized.
\par In the CEM, everything is a point and all points are treated identically. Therefore it is possible to mix sensory modes. For example visual and smell points could be used in combination. Sensory and internal points may be combined.
\subsection{Learning Patterns}\label{subseclearnpat}
A sample (\symS\label{def-S}) is a set of sets of points, typically sense points. A pattern (\vp{a}) within \symS \  is a set of points satisfying the following two conditions: the cardinality of \vp{a} is greater than or equal to a minimum number of points \PATMINP\label{def-minpoints} (equation~\ref{equpat2}); and that \vp{a} is a subset of at least \PATMINO\label{def-minoccurs} sets within \symS (equation~\ref{equpat2}).
\begin{equation}
|a| \geq \PATMINP
\end{equation}\label{equpat1}
\begin{equation}
\sum_{i=1}^{\left | \symS \right |} (\textrm{if} \ a \subset \symS_i \ \textrm{then} \ 1 \ \textrm{else} \ 0) \geq \PATMINO
\end{equation}\label{equpat2}
Patterns are found within \symS \ with the following algorithm: Intersect each set in \symS \ with all the other sets within \symS \ saving all resulting sets with a cardinality greater than or equal to \PATMINP \ as new set of sets, $\mathbb{A}$. Then count the number of times any given set occurs within $\mathbb{A}$.  Any set within $\mathbb{A}$ that occurs at least \PATMINO \ times is considered a pattern and a binding is created with the key set consisting of points in the set and the value being a new internal point.
\par It may be necessary to perform the pattern learning process multiple times. If none of the saved intersected sets occur at least \PATMINO \ times then repeating the learning process on the intersected sets will result in one or more patterns, if any patterns are to be found.
\par As an example, appendix~\ref{datapatrecex} shows a set of sets. Each set contains approximately 20 points with each point represented as an integer number between 0 and 999. There are 40 rows of points. The challenge is to determine whether or not a pattern exists within these 40 rows where, for this example, a pattern is defined as a set of at least 5 points (\PATMINP) found in at least 5 (\PATMINO) rows.
\par The first step is to intersect each row with all of the others. With 40 rows this requires $_{n}C_{r}$ =$_{40}C_{2}$ = $\frac{40!}{(40 - 2)! \times 2!)}$ = $780$ intersections. Figure~\ref{patlearn2} shows the results of the intersections where the cardinality of the result is at least 3 points. Figure~\ref{patlearn3} shows a consolidated result by set. The first two columns are the lines within the data, the third column shows the points resulting from the intersection of the two lines. There are 7 occurrences of \{101 211 307 401 503 601\}. While this count exceeds the \PATMINO (5), it is not the count of occurrences in the first set. If a pattern occurs $n$ times then the number of intersections after the first pass would be $\frac{n\times(n-1)}{2}$, giving 10 for $n = 5$.\footnote{Or given the number of sets after the intersections ($x$) then $\frac{1\pm \sqrt{1+8x}}{2}$ is the maximum value of \PATMINO possible for a pattern.} The additional pass through the aggregated sets adding in for any set that is a super-set gives a final count of 10 (the pattern occurs 5 times in the original data set). Thus a new binding would be created [101 211 307 401 503 601]=\vp{p_{new}} and those points are now recognized as point \vp{p_{new}}.
\begin{figure}[H]
\centering
\begin{subfigure}[b]{0.4\textwidth}
\centering
\begingroup \fontsize{8}{8}\selectfont
\begin{tabular}{r r l}
First&Second&Intersected Points\\
\toprule
4&30&121 246 647\\
7&28&146 211 285\\
14&19&307 326 457\\
16&32&218 275 495\\
19&20&101 211 307 401 503 601\\
19&25&101 211 307 401 503 528 563 601\\
19&29&101 211 307 401 503 601\\
19&33&101 211 307 401 503 591 601\\
20&25&101 211 307 401 503 548 601\\
20&29&101 211 307 401 503 601\\
20&33&101 211 307 401 503 601\\
25&29&101 211 307 401 503 601\\
25&33&101 211 307 401 503 601\\
29&33&101 211 307 401 503 601\\
\bottomrule
\end{tabular}
\endgroup 
\caption{Results of first pass through learning}\label{patlearn2}
\end{subfigure}
\begin{subfigure}[b]{0.4\textwidth}
\centering
\begingroup \fontsize{8}{8}\selectfont
\begin{tabular}{c l}
Times & Points\\
\toprule
7 & 101 211 307 401 503 601\\
1 & 101 211 307 401 503 528 563 601\\
1 & 101 211 307 401 503 591 601\\
1 & 101 211 307 401 503 548 601\\
\bottomrule
\end{tabular}
\endgroup
\caption{Sets with minimum points}\label{patlearn3}
\end{subfigure}
\caption{After passing through intersections}
\end{figure}
\subsection{Patterns in Sequences}
The patterns mentioned so far have not depended on any type of point ordering. But in many cases, the ordering of points does matter. A pattern to recognize the word `tea' cannot be represented with the binding [\vp{t} \vp{e} \vp{a}] = \vp{tea} because the ordering of points in a binding is irrelevant. This binding for `ate' would be [\vp{a} \vp{t} \vp{e}] = \vp{ate}. The key sets [\vp{t} \vp{e} \vp{a}] and [\vp{a} \vp{t} \vp{e}] are equivalent.
\par One way to get around this problem and be able to recognize a specific sequence is to have incremental bindings for each subsequent point in a sequence. For example `tea' would be represented with the points $\vdash \vp{t} \ \vp{e} \ \vp{a} \dashv$ ($\vdash$ and $\dashv$ indicate begin and end of word). The following bindings would be used: [$\vdash$ \vp{t}]=\vp{r_1}, [\vp{r_1} \vp{e}]=\vp{r_2}, [\vp{r_2} \vp{a}]=\vp{r_3} and finally [\vp{r_3} $\dashv$]=\vp{tea}.
\par Another approach is to encode the points with additional positional information. For instance encoding `tea' would be with [\vp{t} \vp{pos1}]=\vp{t_1}, [\vp{e} \vp{pos_2}]=\vp{e_2} and [\vp{a} \vp{pos_3}]=\vp{a_3} and [\vp{t_1} \vp{e_2} \vp{a_3}]=\vp{tea}. The recognition of `eat' would be [\vp{e_1} \vp{a_2} \vp{t_3}]=\vp{eat}.
\par Some sequences may be recognized by a combination of points of different modalities. A melody consists of a sequence of notes with each note having a pitch and duration. The points representing the beginning of a melody would be the set of points {\vp{p_1}, \vp{d_1}, \vp{p_2}, \vp{d_2}, \DOTS } where \vp{p_x} is the pitch for the \vp{x^{th}} note and \vp{d_x} is the duration. A recognition binding for the (beginning) of the melody would be [\vp{p_1} \vp{d_1} \vp{p_2} \vp{d_2} \DOTS] = \vp{melody} where \vp{melody} is the point recognized as the beginning of the melody. The start of the recognition pattern need not necessarily be only the first few notes, it could anywhere throughout the melody. Pitch and duration are just two qualities that are used to recognize a melody. Orchestration and harmonic structure might also be included in the recognition binding.
\subsection{Embedded Patterns within Patterns}\label{secTPwP}
A set of points that represent a pattern may also contain the points of one or more other patterns. Consider the two patterns [\vp{p_1} \vp{p_2} \DOTS \vp{p_n} \vp{q_1} \vp{q_2} \DOTS \vp{q_{n'}}] = \vp{a_{main}} and [\vp{p_1} \vp{p_2} \DOTS \vp{p_n}] = \vp{a_{sub}}. The points of the second pattern are found within the first. When this situation occurs the resulting pattern points are twined. In this case, the twine \vp{a_{sub}}:\vp{a_{main}} would be created.
\par An example of this is the recognition of a visual object as a red ball (i.e. a set of visual points is recognized as a discrete object point \vp{obj}). Within the points recognizing \vp{obj} are the points recognizing the properties of \vp{red} and \vp{ball}, therefore \vp{red}:\vp{obj} and \vp{ball}:\vp{obj}.
\subsection{The Recognition of a Pattern}\label{secRecPat}
When a pattern is recognized from a set of points the resulting pattern point is inserted into the PS with its value variant. For example if the pattern [\vp{p_1} \vp{p_2} \DOTS \vp{p_n}] = \vp{a_r} is recognized then \vp{a_{r}.v} is added to the PS, not \vp{a_r}.
\subsection{An Experiment with Voice Recognition}
An experiment was performed to determine if the pattern recognition process could be used in a real world example. A voice sample recorded as a standard WAV file\cite{wavFormat} was converted to a set of points and put through the pattern learning process. Bindings were created for all learned patterns. Then other test WAV files were similarly converted to sets of points. These test points were inserted into the context and the null key set was evaluated. A successful evaluation indicated that the test points were matched by a previously learned pattern. A failure indicated no pattern.
\par The WAV files used in this experiment were first pre-processed using the FFT algorithm\footnote{FFT or Fast Fourier Transform is an efficient algorithm for converting a time domain signal into the frequency domain\cite[Chapters~12-13]{press}} into encoded \textit{fft} files, a text file consisting of one line per sample of the WAV file. In this experiment the sample duration is $\frac{1}{8}$ of a second. The preprocessing requires two passes through the file. In the first pass, each sample of the WAV file was processed with the FFT. The resulting frequency/magnitude buckets were converted to four digit numbers as shown in (\ref{eqnWav2FFT}).
\begin{equation}\label{eqnWav2FFT}
\left ( \log\left ( f \right )-5 \times 10+0.5 \right ) \times 100 + (\log(mag\times 1000/ \sum mag) + 0.05)
\end{equation}
where \vp{f} is the frequency (0-99), \vp{mag} is the magnitude (0-99). Per FFT convention the magnitude is calculated from the real ($mag_r$) and imaginary ($mag_i$) parts: $\sqrt{(mag_r)^2 + (mag_i)^2)}$
This four digit number is then saved in a bin along with the count of the number of times that number was seen.
\par The second pass is similar to the first except that an output file was created for all samples that occurred more than once. Below is a sample of an fft file. Each line represents an encoding of the spectral analysis of a $\frac{1}{8}$ second sample. Each four digit number represents a frequency and magnitude ($f \times 100 + m$). Figure~\ref{figFFTprep} shows the first few lines of a preprocessed WAV file.
\begin{figure}[H]
\begin{lstlisting}
!\fft\samples\see_ya.wav
1906,2507,2608,2609,2709,2808,2907,2906,3008,3007,3106,3107,3207,3306,3307,3407,3406,3507,3607,3608,
	3609,3709,3708,3808,3907,3908,4008,4007,4108,4107,4209,4210,4310,4309,4308,4408,4407,4406,4507,
	4506,4606,4607,4707,4706,4806,4906,5006,5206
0006,2306,2406,2407,2508,2608,2610,2708,2706,2807,2806,2906,3006,3106,3206,3306,3307,3406,3506,3507,
	3607,3609,	3610,3608,3708,3707,3706,3806,3906,4006,4107,4106,4209,4210,4207,4208,4308,4306,4307,
	4407,4406,4506,4607,4608,4606,4707,4706,4906,4907,5206,5606,6206,6306,6307
0008,1906,2208,2307,2409,2509,2610,2609,2709,2708,2808,2906,2908,2907,3007,3008,3106,3108,3207,3307,
	3408,3409,3509,3510,3610,3609,3709,3708,3707,3808,3807,3806,3908,3907,4007,4008,4006,4108,4109,
	4209,4208,4207,4307,4306,4406,4506,4607,4606,4706,4806,4906
1906,2308,2409,2509,2608,2706,2806,2807,2908,2906,3008,3009,3108,3207,3307,3408,3409,3508,3507,3506,
	3606,3708,3706,3808,3807,3806,3906,3907,4006,4007,4009,4010,4108,4208,4209,4207,4210,4309,4310,
	4308,4307,4306,4407,4406,4408,4409,4509,4508,4506,4507,4608,4607,4609,4707,4706,4708,4808,4807,
	4806,4906,4908,4907,5006,5007,5008,5107,5106,5206,5506,5508,5607,5606,5608,5706,5707,5708,5806,
	5907
\end{lstlisting}
\caption{Sample output from WAV preprocessing}\label{figFFTprep}
\end{figure}
\par The second part of this experiment was finding any patterns in a test \textit{fft} file by first converting each four digit frequency-magnitude into a point using a simple dictionary mapping four-digit numbers to points. Then the pattern learning process as described in section~\ref{subseclearnpat} was performed. The resulting sets of points were bound to a single recognition point. Below is a sample output of the recognition process. Lines 1-\ref{line:ffttrainend} show the learning phase with the training file (FV1MessageMenu.fft).
\par After any patterns were found, tests were performed using WAV files from the same speaker and different speakers. Lines \ref{line:ffttest1a}-\ref{line:ffttest1b} show the result of the first test: 96\% of the lines resulted in a successful evaluation indicating a strong resemblance between the file and the training file. Compare this with lines \ref{line:ffttest2a}-\ref{line:ffttest2b} showing very few successful evaluations.
\begin{lstlisting}
w:  0.0 c:  0.0 - Starting pattern search phase
w: 68.6 c: 68.6 - Created 2744 bindings(*\label{line:ffttrainend}*)
w: 68.6 c: 68.6 - Starting scan of test file (\vic\fft\samples\FV1MessagesUndeleted.fft)(*\label{line:ffttest1a}*)
w: 68.7 c: 68.6 -   26 matches (96%) in 27 lines(*\label{line:ffttest1b}*)
w: 68.7 c: 68.6 - Starting scan of test file (\vic\fft\samples\see_ya.fft)
w: 68.7 c: 68.6 -   0 matches (0%) in 11 lines
w: 68.7 c: 68.6 - Starting scan of test file (\vic\fft\samples\FV1recordAfterTone.fft)
w: 68.7 c: 68.6 -   33 matches (91%) in 36 lines
w: 68.7 c: 68.6 - Starting scan of test file (\vic\fft\samples\VN_UrgentPager_3.fft)(*\label{line:ffttest2a}*)
w: 68.7 c: 68.6 -   3 matches (3%) in 83 lines(*\label{line:ffttest2b}*)
\end{lstlisting}
\subsection{Focus}
In the `real world' all senses are simultaneously and continuously updating the context. It would be desirable to have the ability to limit recognition to only one sensory mode (e.g. visual versus auditory) or a subset of a mode (e.g. words on a page rather than the page itself). This is trivially achieved with the addition of discriminating or focus points. For example, assume a visual object (\vp{o_v}) is recognized by the visual points \vp{v_1}, \vp{v_2}, \DOTS ([\vp{v_1} \vp{v_2} \DOTS] = \vp{o_v}) and a sound (\vp{o_s}) is recognized with the binding [\vp{s_1} \vp{s_2} \DOTS] = \vp{o_s}. If the context includes both sets of points \{\vp{v_1},\vp{v_2},\DOTS,\vp{s_1},\vp{s_2},\DOTS\} then either/both \vp{o_v} or \vp{o_s} could be recognized. However, if we included focus points, \textbf{\vp{f_x}}\label{def-f} in each of these bindings giving [\vp{f_v} \vp{v_1} \vp{v_2} \DOTS] = \vp{o_v} and [\vp{f_s} \vp{s_1} \vp{s_2} \DOTS] = \vp{o_s} then it would be possible to control or focus on visuals or sounds with the inclusion of either \vp{f_v} or \vp{f_s} into the context.
\par A practical example of sensory focus would be the focus of attention to words spoken by a specific individual in a room with others speaking simultaneously. If the patterns that recognized phonemes/words included a focus point (\vp{f_w}) then words would only be recognized when that focus point was in the context. Assume that the particular individual's voice is recognized (as in the prior example) as \vp{p_{voice}} and that point was twined with point \vp{f_w} (\vp{p_{voice}} > \vp{f_w}) then word recognition would only occur when the individual's voice was detected/recognized.
\section{Motivation: The Good and the Bad}\label{secMotivate}
Some points may have one of two additional attributes: `+' and `-'. This optional attribute is used to ascribe good/pleasure or bad/pain to a point. Good points are tagged with the positive \vp{p^+} and bad points are tagged with the negative \vp{p^-}. Most points are neither good nor bad and thus have neither. The V5 engine continuously tracks the number of good/bad points in the PS and sums them by assigning +1 to good points and -1 to bad points. This sum is referred to as \textbf{\SIGMA}\label{def-Sigma}; \SIGMABAR\label{def-SigmaBar} represents a smoothed value of \SIGMA; \DELTA\label{def-Delta} is the difference between the two. These relationships are summarized in equations (\ref{equSigDelta}a/b/c) for moment (int time) $m$ and where $s$ is a smoothing factor between 0.0 and 1.0.
\begin{subequations}\label{equSigDelta}
\begin{align}
\SIGMA_m &= \sum_{i}^{|\{PS\}|} \{PS\}_i \\
\SIGMABAR_{m} &= s * \SIGMA_m + (1-s) * \SIGMABAR_{m-1}  \\
\DELTA_m &= \SIGMA_m - \SIGMABAR_m
\end{align}
\end{subequations}
\par A long term goal of the V5 engine is to maximize \SIGMA \ by maximizing \vp{p^+} points and minimizing \vp{p^-} points. \SIGMA \ indicates the current (instantaneous) good/bad state, \DELTA \ reflects whether \SIGMA \ is increasing or decreasing. A positive \DELTA \ means the current value of \SIGMA \ is greater than the recent average (getting better). Conversely, a negative \DELTA \ indicates that \SIGMA \ is smaller than the recent average (getting worse). Given the V5 engine's goal of maximizing \SIGMA \ over time, what steps can be taken to achieve this?
\begin{enumerate}
\item Make the V5 processing clock speed inversely related to \SIGMA. As \SIGMA \ increases, slow processing clock to maintain the status quo. As \SIGMA \ decreases, increase the clock speed to facilitate change to the status quo.
\item if \DELTA \ goes positive then \textit{conditions} are improving; continue with current actions.
\item if \DELTA \ goes negative then \textit{conditions} are worsening; perform different actions.
\item if both \SIGMA \ and  \DELTA \ are negative then bind neutral points to a minus point.
\end{enumerate}
Many situational factors dictate what `continue with current actions' and `perform different actions' mean. Two different situations are presented in the following simulation examples. The first simulation shows the use of \vp{p^+} points locating a target. The second simulation demonstrates the use of \vp{p^-} points in learning to avoid a `painful' situation.
\subsection{Simulation 1: A Sensor Locking onto a Target}
\begin{figure}[H]
\centering
\begin{tikzpicture}[scale=0.6]
 \draw [domain=-45:45] plot ({3*cos(\x)}, {3*sin(\x)});
 \draw [domain=-45:45] plot ({1*cos(\x)}, {1*sin(\x)});
 \node[] at (1.5,0)  {\tiny{$90\degree$}};
 \draw (0,0)[->] -- (-45:3.5);
 \draw (0,0)[->] -- (45:3.5);
 \foreach \i in {-4,-1,1,4}
  { \draw (\i:3) -- (\i:3.25); }
 \foreach \i in {-10,-7.5,-5,-2.5,0,2.5,5,7.5,10}
  { \draw (\i:3) -- (\i:3.25); }
 \foreach \i in {-15,-20,-25,15,20,25}
  { \draw (\i:3) -- (\i:3.25); }
 \foreach \i in {-35,35}
  { \draw (\i:3) -- (\i:3.25); }
\end{tikzpicture}
\caption{Sensor density increases towards center}\label{figGradient}
\end{figure}
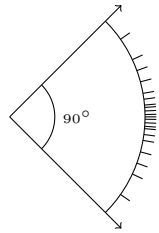

The first simulation investigates how an intity locates a target through the utilization of \vp{p^+} points. The intity has directional sensors such that when a sensor is pointed directly at a target (within $1\degree$) it generates a \vp{p^+} point. There are multiple sensors that span an arc of $90\degree$. The action of the intity is to move ahead a unit distance then decide to turn left $\Theta$ degrees, right $\Theta$ degrees or continue straight ahead. The specific actions are:
\begin{enumerate}
\item If \DELTA \ < -5 and the prior action was move left, then move right. If it was move right then move left. If it was move straight then move left or right, each with a probability of 0.5.
\item If \DELTA \ > 5 then continue moving left/straight/right as before.
\item Otherwise move left with a probability of 0.5, right with a probability of 0.25 or straight with a probability of 0.25. The reason for bias to the left is to ensure that the intity's sensors sweep the entire plane of the simulation space in a reasonable number of steps.
\end{enumerate}
The number of \vp{p^+} points needs to increase as the intity points more directly at the target reaching a maximum when it is heading directly towards the target. As it turns into the target \SIGMA \ increases and \DELTA \ is positive so that it continues its current action (turning or going straight). As it turns away from the target \SIGMA \ decreases and \DELTA \ goes negative so the intity changes direction. In this way it can home in on the target. Packing sensors more densely towards the center of the $90\degree$ arc for a gradient is necessary to achieve the desired result (figure~\ref{figGradient}).
\par The three example runs below show the action of the intity based on various starting positions with respect to the target. The small dot indicates the starting point. The larger dot is the target. The initial bearing is $0\degree$, due east. Runs (a) and (b) the show the bias to move to the left. All three demonstrate that once the target is detected the intity moves to the target.
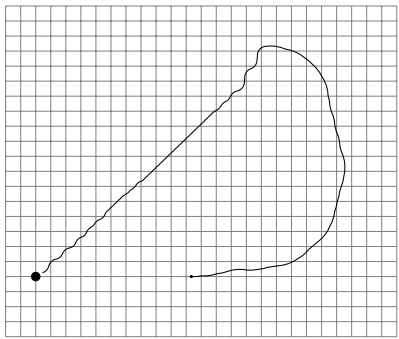
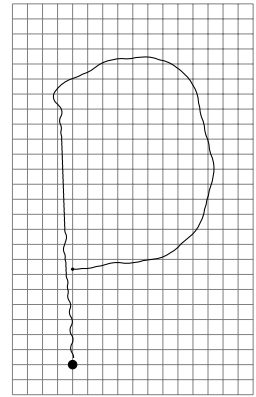
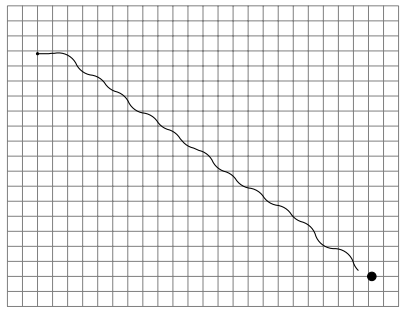
\begin{figure}[H]
\centering
  \begin{subfigure}[b]{0.45\textwidth}
    \centering
    \begin{tikzpicture}[scale=0.4]
\draw[step=0.5cm,gray,very thin] (-1,-2) grid (12,9);
\filldraw (0,0) circle (4pt);
\filldraw (5.17241,0) circle (1pt);
\draw (5.172414,0.000000)--(5.344828,0.000000)--(5.515563,0.023995)--(5.687977,0.023995)--(5.858713,0.047991)--(6.024448,0.095514)--(6.195184,0.119510)--(6.360919,0.167033)--(6.526653,0.214557)--(6.697389,0.238552)--(6.869803,0.238552)--(7.040539,0.214557)--(7.212953,0.214557)--(7.383689,0.238552)--(7.554424,0.262548)--(7.720159,0.310072)--(7.890895,0.334067)--(8.061631,0.358062)--(8.232367,0.382058)--(8.398102,0.429581)--(8.555609,0.499708)--(8.701825,0.591074)--(8.848040,0.682439)--(8.980116,0.793265)--(9.095484,0.921393)--(9.227560,1.032218)--(9.359637,1.143044)--(9.491714,1.253869)--(9.607081,1.381998)--(9.703494,1.524935)--(9.779075,1.679900)--(9.854656,1.834864)--(9.907935,1.998839)--(9.937874,2.168634)--(9.991153,2.332609)--(10.021092,2.502404)--(10.074371,2.666379)--(10.104310,2.836173)--(10.157589,3.000149)--(10.210868,3.164124)--(10.240807,3.333918)--(10.270747,3.503713)--(10.276764,3.676021)--(10.258742,3.847491)--(10.217031,4.014783)--(10.152444,4.174642)--(10.110733,4.341935)--(10.092711,4.513404)--(10.051000,4.680696)--(9.986413,4.840556)--(9.944702,5.007848)--(9.926680,5.179317)--(9.884969,5.346610)--(9.820382,5.506469)--(9.778671,5.673761)--(9.760649,5.845231)--(9.718939,6.012523)--(9.700916,6.183992)--(9.659206,6.351285)--(9.594618,6.511144)--(9.508411,6.660459)--(9.422205,6.809773)--(9.316056,6.945637)--(9.192032,7.065406)--(9.068008,7.185175)--(8.928522,7.286517)--(8.789036,7.387859)--(8.636804,7.468803)--(8.474788,7.527772)--(8.306142,7.563619)--(8.144126,7.622588)--(7.975480,7.658434)--(7.803486,7.670461)--(7.631492,7.658434)--(7.589331,7.649473)--(7.548827,7.634730)--(7.510769,7.614495)--(7.475897,7.589159)--(7.444891,7.559217)--(7.418354,7.525251)--(7.396802,7.487922)--(7.380656,7.447957)--(7.370228,7.406134)--(7.365722,7.363267)--(7.367227,7.320190)--(7.349204,7.148720)--(7.338777,7.106897)--(7.322630,7.066933)--(7.301078,7.029604)--(7.274541,6.995638)--(7.243535,6.965696)--(7.208664,6.940360)--(7.170606,6.920124)--(7.130102,6.905382)--(7.092044,6.885146)--(7.057172,6.859811)--(7.026166,6.829868)--(6.999629,6.795902)--(6.978077,6.758574)--(6.961930,6.718609)--(6.951503,6.676786)--(6.946997,6.633919)--(6.948501,6.590841)--(6.930479,6.419372)--(6.920052,6.377549)--(6.903905,6.337584)--(6.882353,6.300255)--(6.855816,6.266290)--(6.824810,6.236347)--(6.789938,6.211012)--(6.751880,6.190776)--(6.711376,6.176034)--(6.669215,6.167072)--(6.628711,6.152330)--(6.590653,6.132094)--(6.555781,6.106758)--(6.524775,6.076816)--(6.498238,6.042850)--(6.476686,6.005521)--(6.460540,5.965557)--(6.438988,5.928228)--(6.412451,5.894262)--(6.381445,5.864320)--(6.346573,5.838984)--(6.308515,5.818748)--(6.270457,5.798513)--(6.235586,5.773177)--(6.204580,5.743235)--(6.178043,5.709269)--(6.156491,5.671940)--(6.129954,5.637974)--(6.103417,5.604008)--(6.072411,5.574066)--(6.037539,5.548730)--(5.999481,5.528495)--(5.964610,5.503159)--(5.926552,5.482923)--(5.891680,5.457588)--(5.860674,5.427645)--(5.834137,5.393679)--(5.803131,5.363737)--(5.768260,5.338402)--(5.737254,5.308460)--(5.710716,5.274494)--(5.679710,5.244551)--(5.644839,5.219216)--(5.613833,5.189274)--(5.587296,5.155308)--(5.556290,5.125366)--(5.521418,5.100030)--(5.490412,5.070088)--(5.463875,5.036122)--(5.432869,5.006180)--(5.397998,4.980844)--(5.366992,4.950902)--(5.340455,4.916936)--(5.309449,4.886994)--(5.274577,4.861658)--(5.243571,4.831716)--(5.217034,4.797750)--(5.186028,4.767808)--(5.151157,4.742472)--(5.120150,4.712530)--(5.093613,4.678564)--(5.062607,4.648622)--(5.027736,4.623286)--(4.996730,4.593344)--(4.970193,4.559378)--(4.939187,4.529436)--(4.904315,4.504100)--(4.873309,4.474158)--(4.846772,4.440192)--(4.815766,4.410250)--(4.780895,4.384915)--(4.749889,4.354972)--(4.723352,4.321006)--(4.692346,4.291064)--(4.657474,4.265729)--(4.626468,4.235787)--(4.599931,4.201821)--(4.568925,4.171878)--(4.534053,4.146543)--(4.503047,4.116601)--(4.476510,4.082635)--(4.445504,4.052692)--(4.410633,4.027357)--(4.379627,3.997415)--(4.353090,3.963449)--(4.322084,3.933507)--(4.287212,3.908171)--(4.256206,3.878229)--(4.229669,3.844263)--(4.198663,3.814321)--(4.163792,3.788985)--(4.132786,3.759043)--(4.106248,3.725077)--(4.075242,3.695135)--(4.040371,3.669799)--(4.009365,3.639857)--(3.982828,3.605891)--(3.951822,3.575949)--(3.916950,3.550613)--(3.885944,3.520671)--(3.859407,3.486705)--(3.828401,3.456763)--(3.793530,3.431427)--(3.762524,3.401485)--(3.735987,3.367519)--(3.704981,3.337577)--(3.670109,3.312242)--(3.639103,3.282299)--(3.612566,3.248333)--(3.581560,3.218391)--(3.546689,3.193056)--(3.508631,3.172820)--(3.468127,3.158078)--(3.430068,3.137842)--(3.395197,3.112506)--(3.364191,3.082564)--(3.337654,3.048598)--(3.316102,3.011269)--(3.289565,2.977303)--(3.258559,2.947361)--(3.223688,2.922026)--(3.185629,2.901790)--(3.150758,2.876454)--(3.119752,2.846512)--(3.093215,2.812546)--(3.062209,2.782604)--(3.027337,2.757268)--(2.989279,2.737032)--(2.954408,2.711697)--(2.916350,2.691461)--(2.881478,2.666125)--(2.850472,2.636183)--(2.823935,2.602217)--(2.792929,2.572275)--(2.758058,2.546940)--(2.727052,2.516997)--(2.700515,2.483031)--(2.669509,2.453089)--(2.634637,2.427754)--(2.603631,2.397811)--(2.577094,2.363845)--(2.546088,2.333903)--(2.511217,2.308568)--(2.480211,2.278626)--(2.453673,2.244660)--(2.422667,2.214717)--(2.387796,2.189382)--(2.356790,2.159440)--(2.330253,2.125474)--(2.308701,2.088145)--(2.292554,2.048180)--(2.271003,2.010852)--(2.244465,1.976886)--(2.213459,1.946943)--(2.178588,1.921608)--(2.140530,1.901372)--(2.100026,1.886630)--(2.061968,1.866394)--(2.027096,1.841058)--(1.996090,1.811116)--(1.969553,1.777150)--(1.948002,1.739821)--(1.921464,1.705855)--(1.890458,1.675913)--(1.855587,1.650578)--(1.817529,1.630342)--(1.782657,1.605006)--(1.751651,1.575064)--(1.725114,1.541098)--(1.703563,1.503769)--(1.687416,1.463805)--(1.665864,1.426476)--(1.639327,1.392510)--(1.608321,1.362568)--(1.573449,1.337232)--(1.535391,1.316996)--(1.494887,1.302254)--(1.456829,1.282018)--(1.421958,1.256683)--(1.390952,1.226741)--(1.364415,1.192775)--(1.342863,1.155446)--(1.326716,1.115481)--(1.305164,1.078152)--(1.278627,1.044186)--(1.247621,1.014244)--(1.212750,0.988909)--(1.174692,0.968673)--(1.134188,0.953931)--(1.092026,0.944969)--(1.051522,0.930227)--(1.013464,0.909991)--(0.978593,0.884655)--(0.947587,0.854713)--(0.921050,0.820747)--(0.899498,0.783418)--(0.883351,0.743454)--(0.861799,0.706125)--(0.835262,0.672159)--(0.804256,0.642217)--(0.769385,0.616881)--(0.731327,0.596645)--(0.690823,0.581903)--(0.648661,0.572941)--(0.608157,0.558199)--(0.570099,0.537963)--(0.535228,0.512628)--(0.504222,0.482686)--(0.477684,0.448720)--(0.456133,0.411391)--(0.439986,0.371426)--(0.429558,0.329603)--(0.413411,0.289638)--(0.391860,0.252309)--(0.365322,0.218343)--(0.334316,0.188401)--(0.299445,0.163066)--(0.261387,0.142830)--(0.220883,0.128088);
\end{tikzpicture}
\caption{Sample Run from (30,0) to (0,0), target radius = 1, sensor spread = 90 degrees}
  \end{subfigure}\hfill%
  \begin{subfigure}[b]{0.45\textwidth}
    \centering
    \begin{tikzpicture}[scale=0.4]
\draw[step=0.5cm,gray,very thin] (-2,-1) grid (6,12);
\filldraw (0,0) circle (4pt);
\filldraw (0,3.1746) circle (1pt);
\draw (0.000000,3.174603)--(0.158730,3.174603)--(0.315916,3.196694)--(0.474646,3.196694)--(0.631831,3.218785)--(0.784412,3.262537)--(0.941598,3.284628)--(1.094179,3.328380)--(1.246760,3.372132)--(1.403946,3.394223)--(1.562676,3.394223)--(1.719861,3.372132)--(1.878591,3.372132)--(2.035777,3.394223)--(2.192962,3.416314)--(2.345543,3.460066)--(2.502729,3.482157)--(2.659914,3.504248)--(2.817100,3.526339)--(2.969681,3.570091)--(3.114688,3.634652)--(3.249299,3.718766)--(3.383910,3.802880)--(3.505504,3.904910)--(3.611715,4.022870)--(3.733310,4.124899)--(3.854904,4.226929)--(3.976498,4.328959)--(4.082709,4.446919)--(4.171470,4.578512)--(4.241053,4.721178)--(4.310636,4.863843)--(4.359686,5.014805)--(4.387249,5.171123)--(4.436300,5.322085)--(4.463863,5.478403)--(4.512913,5.629365)--(4.540476,5.785683)--(4.589527,5.936645)--(4.638577,6.087606)--(4.666140,6.243925)--(4.693703,6.400243)--(4.699243,6.558877)--(4.682651,6.716738)--(4.644251,6.870753)--(4.584789,7.017925)--(4.546389,7.171940)--(4.529797,7.329801)--(4.491397,7.483816)--(4.431936,7.630988)--(4.393535,7.785003)--(4.376944,7.942864)--(4.338543,8.096879)--(4.279082,8.244051)--(4.240682,8.398066)--(4.224090,8.555927)--(4.185689,8.709942)--(4.169098,8.867803)--(4.130697,9.021818)--(4.071236,9.168990)--(3.991871,9.306454)--(3.912506,9.443918)--(3.814782,9.569000)--(3.700601,9.679263)--(3.586420,9.789526)--(3.458005,9.882825)--(3.329589,9.976124)--(3.189439,10.050644)--(3.040281,10.104933)--(2.885020,10.137935)--(2.735862,10.192223)--(2.580601,10.225225)--(2.422257,10.236298)--(2.263914,10.225225)--(2.105570,10.214153)--(1.950309,10.181151)--(1.791965,10.170079)--(1.633622,10.181151)--(1.475278,10.170079)--(1.320016,10.137077)--(1.164755,10.104075)--(1.015597,10.049786)--(0.875447,9.975267)--(0.747032,9.881967)--(0.618616,9.788668)--(0.478466,9.714149)--(0.329308,9.659860)--(0.189158,9.585341)--(0.040000,9.531052)--(-0.109157,9.476763)--(-0.249308,9.402244)--(-0.377723,9.308944)--(-0.491904,9.198681)--(-0.589628,9.073600)--(-0.609469,9.039234)--(-0.624335,9.002441)--(-0.633935,8.963937)--(-0.638083,8.924472)--(-0.636698,8.884814)--(-0.629807,8.845734)--(-0.617544,8.807994)--(-0.600149,8.772327)--(-0.577959,8.739429)--(-0.551406,8.709939)--(-0.521007,8.684431)--(-0.414796,8.566472)--(-0.392606,8.533574)--(-0.375210,8.497907)--(-0.362947,8.460167)--(-0.356057,8.421087)--(-0.354672,8.381429)--(-0.358820,8.341964)--(-0.368420,8.303460)--(-0.383285,8.266667)--(-0.403126,8.232301)--(-0.417992,8.195508)--(-0.427592,8.157004)--(-0.431740,8.117539)--(-0.430355,8.077880)--(-0.423464,8.038801)--(-0.411202,8.001060)--(-0.393806,7.965394)--(-0.381543,7.927654)--(-0.374652,7.888574)--(-0.373268,7.848916)--(-0.377416,7.809450)--(-0.387016,7.770947)--(-0.391164,7.731482)--(-0.389779,7.691823)--(-0.382888,7.652743)--(-0.370625,7.615003)--(-0.363734,7.575923)--(-0.362350,7.536265)--(-0.366498,7.496800)--(-0.365113,7.457142)--(-0.358222,7.418062)--(-0.356837,7.378404)--(-0.360985,7.338938)--(-0.359600,7.299280)--(-0.352709,7.260200)--(-0.351324,7.220542)--(-0.355472,7.181077)--(-0.354087,7.141418)--(-0.347197,7.102339)--(-0.345812,7.062680)--(-0.349960,7.023215)--(-0.348575,6.983557)--(-0.341684,6.944477)--(-0.340299,6.904819)--(-0.344447,6.865354)--(-0.343062,6.825695)--(-0.336171,6.786616)--(-0.334786,6.746957)--(-0.338934,6.707492)--(-0.337549,6.667834)--(-0.330659,6.628754)--(-0.329274,6.589096)--(-0.333422,6.549631)--(-0.332037,6.509972)--(-0.325146,6.470893)--(-0.323761,6.431234)--(-0.327909,6.391769)--(-0.326524,6.352111)--(-0.319633,6.313031)--(-0.318248,6.273373)--(-0.322396,6.233907)--(-0.321011,6.194249)--(-0.314121,6.155169)--(-0.312736,6.115511)--(-0.316884,6.076046)--(-0.315499,6.036388)--(-0.308608,5.997308)--(-0.307223,5.957650)--(-0.311371,5.918184)--(-0.309986,5.878526)--(-0.303095,5.839446)--(-0.301710,5.799788)--(-0.305858,5.760323)--(-0.304474,5.720664)--(-0.297583,5.681585)--(-0.296198,5.641926)--(-0.300346,5.602461)--(-0.298961,5.562803)--(-0.292070,5.523723)--(-0.290685,5.484065)--(-0.294833,5.444600)--(-0.293448,5.404941)--(-0.286557,5.365862)--(-0.285173,5.326203)--(-0.289320,5.286738)--(-0.287936,5.247080)--(-0.281045,5.208000)--(-0.279660,5.168342)--(-0.283808,5.128877)--(-0.282423,5.089218)--(-0.275532,5.050139)--(-0.274147,5.010480)--(-0.278295,4.971015)--(-0.276910,4.931357)--(-0.270019,4.892277)--(-0.268635,4.852619)--(-0.272783,4.813153)--(-0.271398,4.773495)--(-0.264507,4.734415)--(-0.263122,4.694757)--(-0.267270,4.655292)--(-0.265885,4.615634)--(-0.258994,4.576554)--(-0.257609,4.536895)--(-0.261757,4.497430)--(-0.260372,4.457772)--(-0.253482,4.418692)--(-0.241219,4.380952)--(-0.223823,4.345286)--(-0.211561,4.307545)--(-0.204670,4.268466)--(-0.203285,4.228807)--(-0.207433,4.189342)--(-0.217033,4.150838)--(-0.226633,4.112334)--(-0.241498,4.075541)--(-0.251099,4.037038)--(-0.265964,4.000245)--(-0.275564,3.961741)--(-0.290429,3.924948)--(-0.300029,3.886444)--(-0.304177,3.846979)--(-0.302792,3.807320)--(-0.295902,3.768241)--(-0.283639,3.730500)--(-0.266243,3.694834)--(-0.253981,3.657094)--(-0.247090,3.618014)--(-0.245705,3.578356)--(-0.249853,3.538890)--(-0.248468,3.499232)--(-0.241577,3.460152)--(-0.229315,3.422412)--(-0.222424,3.383332)--(-0.221039,3.343674)--(-0.225187,3.304209)--(-0.223802,3.264551)--(-0.216911,3.225471)--(-0.215526,3.185813)--(-0.219674,3.146347)--(-0.218289,3.106689)--(-0.211399,3.067609)--(-0.210014,3.027951)--(-0.214162,2.988486)--(-0.212777,2.948827)--(-0.205886,2.909748)--(-0.193623,2.872007)--(-0.176228,2.836341)--(-0.163965,2.798601)--(-0.157074,2.759521)--(-0.155689,2.719863)--(-0.159837,2.680397)--(-0.169437,2.641894)--(-0.173585,2.602428)--(-0.172201,2.562770)--(-0.165310,2.523690)--(-0.153047,2.485950)--(-0.146156,2.446870)--(-0.144771,2.407212)--(-0.148919,2.367747)--(-0.158519,2.329243)--(-0.162667,2.289778)--(-0.161283,2.250120)--(-0.154392,2.211040)--(-0.142129,2.173300)--(-0.124733,2.137633)--(-0.102543,2.104735)--(-0.085148,2.069068)--(-0.072885,2.031328)--(-0.065994,1.992248)--(-0.064609,1.952590)--(-0.068757,1.913125)--(-0.078357,1.874621)--(-0.093223,1.837828)--(-0.102823,1.799324)--(-0.106971,1.759859)--(-0.105586,1.720201)--(-0.098695,1.681121)--(-0.086432,1.643381)--(-0.069037,1.607714)--(-0.046847,1.574816)--(-0.029451,1.539150)--(-0.017188,1.501409)--(-0.010298,1.462330)--(-0.008913,1.422671)--(-0.013061,1.383206)--(-0.022661,1.344702)--(-0.037526,1.307909)--(-0.057367,1.273543)--(-0.072233,1.236750)--(-0.081833,1.198246)--(-0.085981,1.158781)--(-0.084596,1.119123)--(-0.077705,1.080043)--(-0.065442,1.042303)--(-0.048047,1.006636)--(-0.025856,0.973738)--(-0.008461,0.938072)--(0.003802,0.900331)--(0.010693,0.861252)--(0.012077,0.821593)--(0.007930,0.782128)--(-0.001671,0.743624)--(-0.016536,0.706831)--(-0.036377,0.672465)--(-0.051242,0.635672)--(-0.060843,0.597168)--(-0.064991,0.557703)--(-0.063606,0.518045)--(-0.056715,0.478965)--(-0.044452,0.441225)--(-0.027057,0.405558)--(-0.004866,0.372660)--(0.012529,0.336994)--(0.024792,0.299253)--(0.031683,0.260174)--(0.033068,0.220515);
\end{tikzpicture}
\caption{Sample Run from (0,20) to (0,0), target radius = 1, sensor spread = 90 degrees}
  \end{subfigure}
  \begin{subfigure}[b]{0.45\textwidth}
    \centering
    \begin{tikzpicture}[scale=0.4]
\draw[step=0.5cm,gray,very thin] (-1,-1) grid (12,9);
\filldraw (11.1111,0) circle (4pt);
\filldraw (0,7.40741) circle (1pt);
\draw (0.000000,7.407407)--(0.370370,7.407407)--(0.462062,7.420294)--(0.554654,7.420294)--(0.646346,7.433180)--(0.738939,7.433180)--(0.830630,7.420294)--(0.919636,7.394772)--(1.004223,7.357111)--(1.082746,7.308044)--(1.153676,7.248527)--(1.215633,7.179717)--(1.267410,7.102955)--(1.308000,7.019733)--(1.359777,6.942970)--(1.421734,6.874160)--(1.492664,6.814643)--(1.571187,6.765577)--(1.655774,6.727916)--(1.744780,6.702394)--(1.836471,6.689507)--(1.925477,6.663985)--(2.010065,6.626325)--(2.088588,6.577258)--(2.159518,6.517741)--(2.221474,6.448931)--(2.273251,6.372168)--(2.335208,6.303358)--(2.406138,6.243841)--(2.484661,6.194774)--(2.569248,6.157114)--(2.658254,6.131592)--(2.742842,6.093931)--(2.821365,6.044864)--(2.892295,5.985347)--(2.954251,5.916537)--(3.006028,5.839774)--(3.046618,5.756553)--(3.098395,5.679790)--(3.160352,5.610980)--(3.231282,5.551463)--(3.309805,5.502396)--(3.394392,5.464736)--(3.483398,5.439214)--(3.575090,5.426327)--(3.664095,5.400805)--(3.748683,5.363144)--(3.827206,5.314078)--(3.898136,5.254560)--(3.960092,5.185751)--(4.011869,5.108988)--(4.073826,5.040178)--(4.144756,4.980661)--(4.223279,4.931594)--(4.307867,4.893934)--(4.396872,4.868412)--(4.481460,4.830751)--(4.559983,4.781684)--(4.630913,4.722167)--(4.692869,4.653357)--(4.744647,4.576594)--(4.806603,4.507785)--(4.868560,4.438975)--(4.939490,4.379458)--(5.018013,4.330391)--(5.102600,4.292730)--(5.191606,4.267208)--(5.276193,4.229547)--(5.360781,4.191887)--(5.449787,4.166365)--(5.534374,4.128704)--(5.612897,4.079637)--(5.683827,4.020120)--(5.745784,3.951310)--(5.797561,3.874547)--(5.838151,3.791326)--(5.889928,3.714563)--(5.951884,3.645753)--(6.022814,3.586236)--(6.101337,3.537169)--(6.185925,3.499509)--(6.274931,3.473987)--(6.359518,3.436326)--(6.438041,3.387259)--(6.508971,3.327742)--(6.570928,3.258932)--(6.622705,3.182169)--(6.684661,3.113360)--(6.755591,3.053842)--(6.834114,3.004776)--(6.918702,2.967115)--(7.007708,2.941593)--(7.099399,2.928706)--(7.188405,2.903185)--(7.272992,2.865524)--(7.351515,2.816457)--(7.422445,2.756940)--(7.484402,2.688130)--(7.536179,2.611367)--(7.598136,2.542558)--(7.669066,2.483040)--(7.747589,2.433974)--(7.832176,2.396313)--(7.921182,2.370791)--(8.012873,2.357904)--(8.101879,2.332382)--(8.186467,2.294722)--(8.264990,2.245655)--(8.335920,2.186138)--(8.397876,2.117328)--(8.449653,2.040565)--(8.511610,1.971756)--(8.582540,1.912238)--(8.661063,1.863172)--(8.745650,1.825511)--(8.834656,1.799989)--(8.919244,1.762328)--(8.997767,1.713261)--(9.068697,1.653744)--(9.130653,1.584934)--(9.182430,1.508172)--(9.223020,1.424950)--(9.251633,1.336889)--(9.292223,1.253667)--(9.344000,1.176905)--(9.405957,1.108095)--(9.476887,1.048578)--(9.555410,0.999511)--(9.639997,0.961850)--(9.729003,0.936328)--(9.820694,0.923442)--(9.913287,0.923442)--(10.004978,0.910555)--(10.093984,0.885033)--(10.178572,0.847373)--(10.257095,0.798306)--(10.328025,0.738789)--(10.389981,0.669979)--(10.441758,0.593216)--(10.482348,0.509995)--(10.510961,0.421934)--(10.551551,0.338712)--(10.603328,0.261949)--(10.665285,0.193140);
\end{tikzpicture}
\caption{Sample Run from (0,20) to (30,0), target radius = 1, sensor spread = 90 degrees}
  \end{subfigure}
\caption{Sample runs with differing start/end positions}
\end{figure}
\subsection{Simulation 2: Learning to Avoid a `Painful' Wall}
The second simulation has an intity enclosed within a square pen. The intity is free to roam randomly within the pen. If the intity hits any of the sides of the pen then it receives a painful jolt. The intity's sensors register the pain by inserting \vp{p^-} points into the context. The intity has additional sensors that create normal (i.e. neither + nor -) sensory points whenever it is in close proximity to any of the four sides. All of these sensed points are inserted into the intity's processing context. With each processing step, the null key set is evaluated. Any result is additionally added into the context. If \SIGMA \ goes negative (from \vp{p^-} points) then it immediately reverses direction (i.e. when it senses pain it reverses).
\par In the two runs below, the intity starts in the middle of the pen facing north. Both figures below show the intity's path over 5000 moves. There is no learning in the first run (figure~\ref{v5demo2-1}) and the intity repeatedly hits the boundaries. In the second run (figure~\ref{v5demo2-2}) it learns by binding the sensory points (proximity to a boundary) with pain points whenever \DELTA \ goes negative. The intity recognizes pain before experiencing it and can reverse itself before hitting a boundary.
\begin{figure}[H]
\centering
  \begin{subfigure}[b]{0.45\textwidth}
    \include*{v5demo2-1}
  \end{subfigure}
  \begin{subfigure}[b]{0.45\textwidth}
    \include*{v5demo2-2}
  \end{subfigure}
\caption{Sample runs with and without learning}
\end{figure}
\section{Thoughts, Language and Meaning}\label{secThoughtsLanMean}
The CEM has now been introduced and with the V5 engine many examples have been presented. Mazes have been solved, songs have been sung. A Turing machine has been implemented demonstrating that at least in principle, any computable function can be calculated with CEM/V5. None of the examples presented so far would be considered difficult AI problems. None of them are novel and all have been `solved' with many different AI and non-AI paradigms.
\par The second half of this paper is devoted to a more challenging and fundamental problem in AI: what are \textit{thoughts} and how do thoughts relate to language and meaning? This section defines thoughts within the CEM framework. Section~\ref{secThought2Lan} demonstrates a sequence for translating thoughts to language by taking a single thought and generating both English and French output sentences. The converse problem of converting a string of words to a thought is covered in section~\ref{secLan2Thought}. Examples showing how \textit{meaning} arises from thought patterns are given in section~\ref{secconsofthought}.
\subsection{Thoughts}\label{secThoughts}
\begin{definition}{}
A \textbf{thought}\label{def-t} is a point, \vp{t}, at the base of a multi-branch is-a tree (is-a twines) where the separate branches of the tree are the components of the thought.
\end{definition}
At its simplest, the thought of `x' (as represented as point \vp{x}) would be point \vp{x} is-a twined to a thought point (\vp{x}<\vp{t}). More complex thoughts have multiple branches with additional points that relate the branches to each other. For example, the thought corresponding to the sentence `Sue walked her dog' is (thought) point with two is-a branches. The first would be the point representing Sue, the second would be the point representing her dog: \vp{pSue}<\vp{t} and \vp{pDog}<\vp{t}. Additional is-a twines would relate the Sue point (\vp{pSue}) to her dog point (\vp{pDog}) as the subject and object of the verb walk: \vp{subWalk}<\vp{pSue} and \vp{objWalk}<\vp{pSue}. But these last two twines are not quite correct. \vp{pSue} is not always the subject of \vp{walk} and \vp{pDog} is not always the object. They are subject/object only for this thought. Or to restate only in the context of this thought. So these two twines are more properly written as \vp{subWalk}<\vp{pSue}|\vp{t} and \vp{objWalk}<\vp{pDog}|\vp{t}.
\par
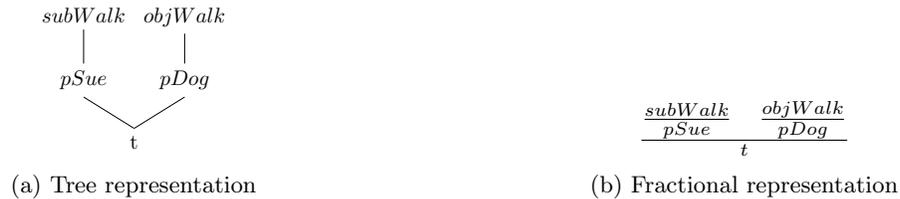
\begin{figure}[H]
\centering
\begin{subfigure}[b]{0.45\textwidth}
\centering
\begin{tikzpicture}[scale=0.8, grow'=up]
\Tree [.t$$  [.\vp{pSue} \vp{subWalk} ]  [.\vp{pDog} \vp{objWalk} ] ]
\end{tikzpicture}
\caption{Tree representation}
\end{subfigure}
\begin{subfigure}[b]{0.45\textwidth}
\centering
$\frac{\begin{matrix} \frac{subWalk}{pSue} & \frac{objWalk}{pDog}\\ \end{matrix}}{t}$
\caption{Fractional representation}
\end{subfigure}
\caption{Tree and fractional representations of the thought `Sue walked her dog'}\label{figTreeFrac}
\end{figure}

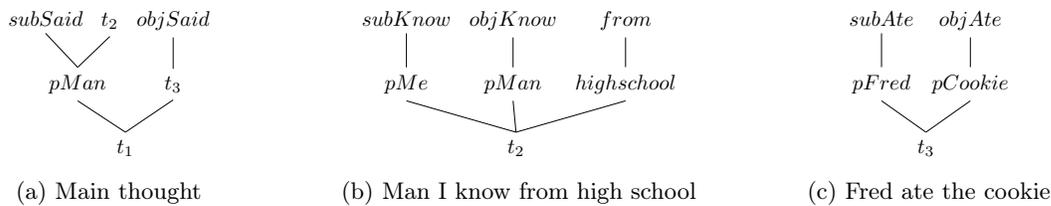
\begin{figure}[H]
  \begin{subfigure}[b]{0.30\textwidth}
\centering
\begin{tikzpicture}[scale=0.8, grow'=up]
\Tree [.$t_1$ [.\vp{pMan} \vp{subSaid} $t_2$ ] [.$t_3$ \vp{objSaid} ] ]
\end{tikzpicture}
\caption{Main thought}
  \end{subfigure}
  \begin{subfigure}[b]{0.30\textwidth}
    \centering
\begin{tikzpicture}[scale=0.8, grow'=up]
\Tree [.$t_2$ [.\vp{pMe} \vp{subKnow} ] [.\vp{pMan} \vp{objKnow} ] [.\vp{highschool} \vp{from} ] ]
\end{tikzpicture}
\caption{Man I know from high school}
  \end{subfigure}
  \begin{subfigure}[b]{0.30\textwidth}
\centering
\begin{tikzpicture}[scale=0.8, grow'=up]
\Tree [.$t_3$ [.\vp{pFred} \vp{subAte} ] [.\vp{pCookie} \vp{objAte} ] ]
\end{tikzpicture}
\caption{Fred ate the cookie}
  \end{subfigure}
\caption{Embedding thoughts within a thought}\label{figEmbedT}
\end{figure}

 Representing a thought as a collection of twines can be difficult to visualize especially for complicated thoughts. However, showing a thought graphically as a \textit{tree}\label{def-tree} makes it easier to understand the structure of a thought. In a thought \textit{tree}, the nodes represent points and the branches represent is-a twines. To further simplify, specific \textit{contextual} restrictions on a twine are not shown (e.g. \vp{subWalk}<\vp{pSue} and \vp{subWalk}<\vp{pSue}|t would have identical tree representations. A more compact \textit{fractional}\label{def-fractional} notation may also be used when representing is-a relationships. Figure~\ref{figTreeFrac} shows both the tree and fractional representations of the thought `Sue walked her dog'.
Thoughts are points so it is possible to include one thought within another. Consider the thought `The man I knew from high school said that Fred ate the cookie'. This thought can be decomposed into two sub-thoughts within a third. Figure~\ref{figEmbedT} shows\vp{t1}, the main thought, linking the subject  'man I knew from high school' (\vp{t2}) to the object 'Fred ate the cookie' (\vp{t3}).
\subsection{Meaning of a Thought}
\begin{figure}[H]
\centering
\begin{tikzpicture}[scale=0.8, grow'=up]
\Tree [.$t_x$ [.$s_1$ \vp{Jack} \vp{Jill} $went\_L$ ] [.$s_2$ $went\_R$ $up\_R$ \vp{hill} ] ]
\end{tikzpicture}
\caption{Thought for `Jack and Jill went up the hill'}\label{figJackJill}
\end{figure}
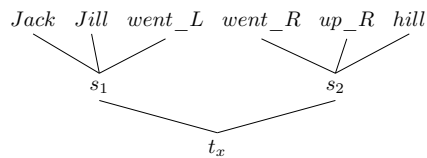

The \textit{meaning} of a thought is defined as \textit{the subsequent actions taken by an intity as a consequence of the thought}. A single thought may have different meanings for different people. An individual's interpretation of a thought may change over time. Consider the thought shown in figure~\ref{figJackJill}.
What is the meaning to the reader? No prior mention was made as to whom Jack or Jill are. There was no prior mention of hills in this paper. Yet each reader will interpret the sentence and have subsequent thoughts, for example:
\begin{clist}
\item Think that it's just an example to define the author's point. There is no meaning per se.
\item Recall the old nursery rhyme and maybe continue it in their mind (`to fetch a pail of water').
\item Mentally picture two children jointly holding a bucket walking up a hill towards a well.
\end{clist}
Adding a thought point to the context (or the PS for V5) effectively adds all of the is-a points of the thought to the context. These points as well as any other points in the context (sensory, control and internal) can be used to recognize patterns. The value of a pattern can be another thought point, the starting point of a sequence, a long term goal point, etc. The constant processing of sensory input and thoughts leading to other (new) thoughts results in a chaotic stream of points flowing through the intity.
\section{Converting Thoughts to Language}\label{secThought2Lan}
The conversion of a thought to a sequence of words requires bindings representing information for:
\begin{clist}
\item words corresponding to the various points comprising the thought (e.g. `John' for point \vp{nounJohn});
\item rules for converting a point to one or more words (e.g. the ordering of adjectives before a noun);
\item rules for the overall structure of the resulting sentence (e.g. active versus passive voice);
\item a sequence to accomplish the task given the above information.
\item and the thought point to be converted to a sequence of words;
\end{clist}
The example below demonstrates one way this can be done. It is self-contained with all the bindings necessary to convert the thought representing `John has the red ball' into an appropriate sequence of words reflecting the thought.
\par Everything in the CEM is contextual. First, an English sentence is generated. Then a few more bindings are added to demonstrate contextual flexibility. The addition of point \vp{french} and associated bindings results in the thought generating a French language sentence.
\par The example begins with bindings defining the value for point \vp{label} in various contexts corresponding to different points in the thought. Line~\ref{line:ttllabels} declares output labels in various contexts. For example, the value of \vp{label} given the context of \vp{nounJohn} is ``John''.
\begin{lstlisting}
twine noun<nounJohn,nounBall
twine adj<adjRed|english,adjBig
twine daLabel>"the" ; label>"John"|nounJohn ; twine label>"He"|pnounJohn ;(*\label{line:ttllabels}*)
  label>"big"|adjBig ; label>"red"|adjRed ; label>"ball"|nounBall ; label>"has"|verbHas
\end{lstlisting}
\begin{exDesc}
The code in lines \ref{line:ttlsetup} through \ref{line:ttlsetup2} are instructions to create the thought in figure~\ref{figJohnHasBall} of `John has the big red ball''. Line~\ref{line:ttlsetup} inserts instructions and arguments into the PS to create twines corresponding to the tree below. The branches of the tree are is-a twines. The nodes are points. Points of the form `$\#number$' are internal points created with the opNEW opcode.
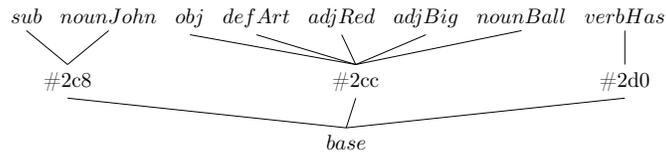
\begin{figure}[H]
\centering
\begin{tikzpicture}[scale=0.8, grow'=up]
\Tree [.\vp{base}  [.\#2c8  \vp{sub} \vp{nounJohn} ] [.\#2cc \vp{obj} \vp{defArt} \vp{adjRed} \vp{adjBig} \vp{nounBall} ] [.\#2d0 \vp{verbHas} ]]
\end{tikzpicture}
\caption{Tree representation of `John has the big red ball'}\label{figJohnHasBall}
\end{figure}
\end{exDesc}
\begin{lstlisting}[firstnumber=5]
trace opTWISA
ps opNEW,rNEW,base,opTWISA,sub,rNEW,opTWISA,nounJohn,rNEW,opTWISA,(*\label{line:ttlsetup}*)
	opNEW,rNEW,base,opTWISA,obj,rNEW,opTWISA,defArt,rNEW,opTWISA,adjRed,
	rNEW,opTWISA,adjBig,rNEW,opTWISA,nounBall,rNEW,opTWISA,opNEW,rNEW,
	base,opTWISA,verbHas,rNEW,opTWISA
run(*\label{line:ttlsetup2}*)
(*\vout{opTWISA: \#44: [base.i rCTP(1)] = \#2e8}*)
(*\vout{opTWISA: \#45: [\#2e8.i rCTP(1)] = sub}*)
(*\vout{opTWISA: \#46: [\#2e8.i rCTP(1)] = nounJohn}*)
(*\vout{opTWISA: \#47: [base.i rCTP(1)] = \#2ec}*)
(*\vout{opTWISA: \#48: [\#2ec.i rCTP(1)] = obj}*)
(*\vout{opTWISA: \#49: [\#2ec.i rCTP(1)] = defArt}*)
(*\vout{opTWISA: \#50: [\#2ec.i rCTP(1)] = adjRed}*)
(*\vout{opTWISA: \#51: [\#2ec.i rCTP(1)] = adjBig}*)
(*\vout{opTWISA: \#52: [\#2ec.i rCTP(1)] = nounBall}*)
(*\vout{opTWISA: \#53: [base.i rCTP(1)] = \#2f0}*)
(*\vout{opTWISA: \#54: [\#2f0.i rCTP(1)] = verbHas}*)
\end{lstlisting}
\begin{exDesc}
The next several groupings define bindings to be used by the opRASM\footnote{See section~\ref{secASReduce} for more detail on the opRASM instruction.} instruction. Line~\ref{line:defartadj} states that a definite article (\vp{defArt}) followed by an adjective (\vp{adj}) should trigger the adding of the label of the article (\vp{daLabel.v}) to the print queue (opADDPQ). The reference to the definite article should be removed (-\vp{defArt}) with no replacement (the value list ends with the \vp{null} point). Note that the \vp{asPH} point is automatically appended to the AS with opISATOAS instruction and is used to mark the end of a series of points.
\end{exDesc}
\begin{lstlisting}[firstnumber=21]
bind +10 [-defArt/adj] daLabel.v,opADDPQ,null(*\label{line:defartadj}*)
bind +10 [adj/-defArt] daLabel.v,opADDPQ,null
bind +10 [-defArt/noun] daLabel.v,opADDPQ,null
bind +10 [noun/-defArt] daLabel.v,opADDPQ,null
bind +10 [-defArt/asPH] daLabel.v,opADDPQ,null
bind +5 [-adj/noun] eoa,@label.v,adj.v,opEVAL,rEVAL,opADDPQ,null
bind +5 [noun/-adj] eoa,@label.v,adj.v,opEVAL,rEVAL,opADDPQ,null
bind +5 [-adj/asPH] eoa,@label.v,adj.v,opEVAL,rEVAL,opADDPQ,null
bind +2 [-noun/asPH] eoa,@label.v,noun.v,opEVAL,rEVAL,opADDPQ,null
bind +2 [-noun/adjpost] eoa,@label.v,noun.v,opEVAL,rEVAL,opADDPQ,null
bind +2 [adjpost/-noun] eoa,@label.v,noun.v,opEVAL,rEVAL,opADDPQ,null
bind [-adjpost/asPH] eoa,@label.v,adjpost.v,opEVAL,rEVAL,opADDPQ,null
bind [-sub/asPH] null
bind [-obj/asPH] null
\end{lstlisting}
\begin{exDesc}
The following twines define the sequence that convert a thought into a series of words. Line~\ref{line:speakseq} evaluates [\vp{speak}] to get the `next' step of the sequence. The binding on line~\ref{line:speakbind} defines [\vp{speak}] in the context of the additional points (\vp{sub}, \vp{verbHas} and \vp{obj}) to have the value \vp{lsr.v} (line~\ref{line:speaklsr}). This clears the PQ (opCLRPQ), pushes the value of the subject onto the PS (\vp{sub.v}), runs the opISATOAS instruction to insert all is-a points of the subject into the AS and then run the AS reduction instruction (opRASM). This results in the words representing the subject to be appended to the PQ. The word for the \vp{verbHas} is appended and then the words for the object. The last step of the sequence outputs the contents to the PQ to the console.
\end{exDesc}
\begin{lstlisting}[firstnumber=35]
twine lsa>opPSISAS,speak,eoa,opEVAL,rEVAL(*\label{line:speakseq}*)
twine lsr>opCLRPQ,sub.v,opISATOAS,lsr2.v(*\label{line:speaklsr}*) 
twine lsr2>opRASM,lsr3,lsr3.v
twine lsr3>@label.v,verbHas,eoa,opEVAL,rEVAL,opADDPQ,lsr4.v
twine lsr4>obj.v,opISATOAS,lsr5.v
twine lsr5>opRASM,lsr6.v
twine lsr6>opOUTPQ
bind [speak sub verbHas obj] lsr.v(*\label{line:speakbind}*)
\end{lstlisting}
\begin{exDesc}
Starting up the sequence is done by adding the base thought (\vp{base}), the language (\vp{english}) and the starting point (\vp{lsa.v}) for the sequence. The run command starts the execution and the sentence corresponding to the thought is output.
\end{exDesc}
\begin{lstlisting}[firstnumber=43]
ps base,english,lsa.v
run
(*\vout{PQ(3): "John" "has" "the" "big" "red" "ball"}*)
\end{lstlisting}
\begin{exDesc}
The remainder of the example demonstrates how a few additional bindings can be used to convert the same thought into French. Twines with the additional context point (\vp{french}) are given for the French equivalents of several word/points (lines starting at \ref{line:ttlfrenchwords}). The twine on line~\ref{line:ttlfrenchred} specifies that in the context of \vp{french} the \vp{adjRed} point is-a \vp{adjPost}. This, in combination with the binding on line~\ref{line:ttlfrenchred2}, ensures that the word for the color red appears after the noun as is typical for color adjectives in French.
\end{exDesc}
\begin{lstlisting}[firstnumber=46]
twine adjpost<adjRed|french(*\label{line:ttlfrenchred}*)
twine daLabel>"la"|french(*\label{line:ttlfrenchwords}*)
twine label>"grande"|adjBig,french
twine label>"rouge"|adjRed,french
twine label>"balle"|nounBall,french
twine label>"a"|verbHas,french
bind +10 [-defArt/adjpost] daLabel.v,opADDPQ,null(*\label{line:ttlfrenchred2}*)
bind +10 [adjpost/-defArt] daLabel.v,opADDPQ,null
\end{lstlisting}
\begin{exDesc}
Finally, rerun the sequence but this time with \vp{french} in the context. 
\end{exDesc}
\begin{lstlisting}[firstnumber=54]
ps base,french,lsa.v
run
(*\vout{PQ(5): "John" "a" "la" "grande" "balle" "rouge"}*)
\end{lstlisting}
The key points of this example are:
\begin{enumerate}
\item The thought is the starting point. A thought is fundamentally independent of language.
\item A generating sequence is determined from the overall structure and content of the thought (line~\ref{line:speakbind}). Different sequences may be evaluated for different modes of expression (active voice versus passive voice) or any other contextual conditions. This implies that any thought can be expressed as many different sentences.
\item The words for each element of the sentence are determined contextually. For example an additional twine (\vp{label}>`Johnny'|\vp{nounJohn},\vp{diminutive}) specifies that `Johnny' instead of `John' should be output when the point \vp{diminutive} is in the context.
\end{enumerate}
\section{Converting Language to Thoughts}\label{secLan2Thought}
Language is the medium used to communicate a thought from one intity (the speaker) to another (the listener) where the speaker and listener have different internal point structures. A simple example is communicating a reference to an object (a particular red ball) that both the speaker and listener are familiar with. The speaker has the thought \vp{ball_s}<\vp{t_s} where \vp{ball_s} is the speaker's internal point for the ball and \vp{t_s} is the speaker's corresponding thought. The corresponding thought for the listener is \vp{ball_l}<\vp{t_l} where \vp{ball_l} is the listener's internal point for the (same) ball and \vp{t_l} is a listener's thought point. How does the speaker communicate the thought \vp{ball_l}<\vp{t_l} to the listener without having access to the listener's point \vp{ball_l}?
\subsection{Language to Thought with Surrogates}
The answer is that the speaker uses attributes  of \vp{ball_s} to describe it in words. In this example \vp{colorRed_s}<\vp{ball_s} and \vp{shapeBall_s}<\vp{ball_s} result in the phrase `red ball'. Upon hearing this phrase, the listener converts the spoken/heard word `red' to \vp{colorRed_l} and the word `ball' to \vp{shapeBall_l} and uses these points to locate/find/deduce \vp{ball_l}.
\par
\begin{figure}[H]
\centering
\begin{tikzpicture}[scale=0.8, grow'=up]
\Tree [.\vp{t_s} [.\vp{ball_s} \vp{colorRed_s} \vp{shapeBall_s} ] ]
\node[anchor=west,text width=4cm] (note1) at (2.5,.5) { $\Longrightarrow$ };
\node[anchor=west,text width=5cm] (note1) at (2,1) { \small{`red ball'} };
\begin{scope}[xshift=6cm]
\Tree [.\vp{t_l} [.\vp{s_l} \vp{colorRed_l} \vp{shapeBall_l} ] ]
\end{scope}
\node[anchor=west,text width=4cm] (note1) at (7.5,1) { + };
\begin{scope}[xshift=10cm]
\Tree [.$t'_l$ [.\vp{ball_l} \vp{colorRed_l} \vp{shapeBall_l} ] ]
\end{scope}
\node[anchor=west,text width=5cm] (note2) at (12,1) { $\Longrightarrow$ \small{\vp{s_l} $\equiv$ \vp{ball_l}} };
\end{tikzpicture}
\caption{Using language to communicate `red ball' from speaker to listener}\label{figXmitRedBall}
\end{figure}
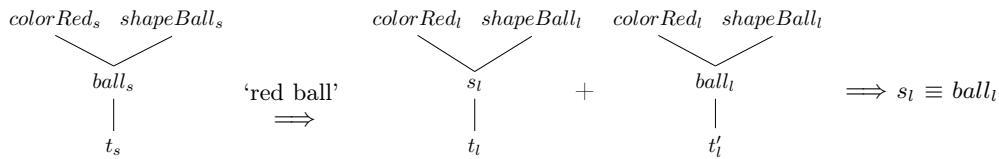

 A new type of point is needed for this process to work in the CEM environment.
\begin{definition}{} A \textbf{surrogate}\label{def-surrogate}\label{def-s} is a point (\vp{s_x}), a stand-in for another to-be-determined point.
\end{definition}
In figure~\ref{figXmitRedBall} the speaker's thought (\vp{t_s}) is converted to the words `red ball'. The listener, upon hearing, creates a thought (\vp{t_l}) with the surrogate point \vp{s_l}. Surrogate \vp{s_l} has two qualifying attributes linked via is-a twines (\vp{colorRed_l} and \vp{shapeBall_l}). The listener searches prior thoughts for a point having the same two points as is-a twines. A match is found with thought \vp{t'_l} with point \vp{ball_l} so surrogate \vp{s_l} is bound to point \vp{ball_l}.
\subsection{Resolving Surrogates}
While the issue of resolving a surrogate to its corresponding point is an area of ongoing research, the simplest approach is to search backward through prior thoughts looking for a point with is-as covering the is-a attributes of the surrogate. For example a surrogate (\vp{s_x}) having the twines \vp{rare}<\vp{s_x}, \vp{juicy}<\vp{s_x} and \vp{steak}<\vp{s_x} (rare juicy steak) would be resolved by locating another point with at least those same attributes.
\par Surrogates will have is-a attributes that are not to be included in the searching. These would be syntactic attributes local to the current thought. The inclusions or exclusion of these attributes can be handled with additional points in the is-a twines.
\par
\begin{figure}[H]
  \begin{subfigure}[b]{0.45\textwidth}
\centering
\begin{tikzpicture}[scale=0.8, grow'=up]
\Tree [.t1 [.\vp{s1} \vp{red} \vp{blue} \vp{ball} ] ]
\end{tikzpicture}
\caption{Representation of `red blue ball'}
  \end{subfigure}
  \begin{subfigure}[b]{0.45\textwidth}
    \centering
\begin{tikzpicture}[scale=0.8, grow'=up]
\Tree [.t2 [.\vp{s2} \vp{rbPoint} \vp{ball} ] ]
\node[anchor=west,text width=4cm] (note1) at (2,1) {
where \vp{rbPoint}<\vp{red}|\vp{t2} and \vp{rbPoint}<\vp{blue}|\vp{t2}  };
\end{tikzpicture}
\caption{Representation of `red or blue ball'}
  \end{subfigure}
\caption{The difference between `red blue ball' and `red or blue ball'}\label{figredblueball}
\end{figure}
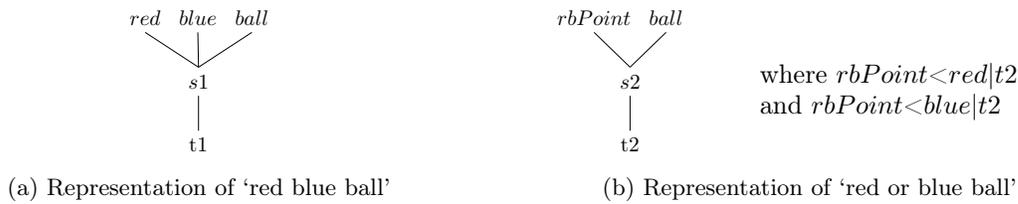

 The is-a twines off of a surrogate constitute a logical \textit{and} condition. A `rare juicy steak' must be rare and juicy and a steak. Logical \textit{or} conditions (`chocolate or strawberry milkshake') can be easily accommodated with the addition of thought specific twines. In figure~\ref{figredblueball} the point \vp{csflavor} is twined to both \vp{chocolate} and \vp{strawberry} (\vp{csflavor}<\vp{chocolate}|\vp{t_2}, \vp{csflavor}<\vp{strawberry}|\vp{t_2}) within the context of the thought \vp{t_2}. The twine \vp{csflavor}<\vp{surrogate} links the surrogate to this new point. Now either a chocolate or strawberry milkshake would match the \vp{csflavor} attribute. A similar strategy is shown in figure~\ref{figredblueball} with the `red [and] blue ball' versus `red or blue ball'.
\par The handling of definite versus indefinite articles may be as simple as resolving an indefinite surrogate to a new point rather than performing any type of search. With `You will meet a tall dark stranger' the surrogate for `a tall dark stranger' can be replaced with just a new point. Nested clauses (`the fish that got away') require searching for prior conditional thoughts. Finally, in many instances, a surrogate may never be resolved. In `Jack and Jill went up the hill', the reader has no idea who either Jack or Jill are.
\subsection{Grounded versus Non-Grounded Chiral Points}
The task of converting a sequence of words into a thought is primarily the process of converting those words into surrogates, all is-a twined to a thought point.
\begin{definition}{}
A \textbf{grounded}\label{def-grounded} point is a point that has an is-a twine to an external sensory point.\footnote{See \cite{cogprints3106} for more on the grounding problem.}
\end{definition}
The first step is to distinguish grounded points from non-grounded points. This is done with an is-a twine to the \vp{grounded} point (\vp{grounded}<\vp{p}). Nouns and adjectives are grounded.
\begin{definition}{}
Points that are not grounded are considered \textbf{chiral}\label{def-chiral}\footnote{The word \textit{chiral} is related to handedness (left hand, right hand).}.
\end{definition}
Verbs, articles, adverbs and prepositions are not grounded. These points have a related left hand point and right hand point (\vp{p\_L} and \vp{p\_R}). For example the non-grounded point \vp{verbWalk} has left and right hand counterparts ($verbWalk\_L$ and $verbWalk\_R$). The reason for this is explained in section~\ref{secconsofthought}.
\par
\begin{figure}[H]
  \begin{subfigure}[b]{0.45\textwidth}
\centering
\begin{tikzpicture}[scale=0.8, grow'=up]
\Tree [.\vp{t} [.\vp{s_1} \vp{John} \vp{sub} ] [.\vp{threw} \vp{verb} ] [.\vp{s_2} \vp{ball} \vp{obj} ] ]
\end{tikzpicture}
\caption{Old representation}
  \end{subfigure}
  \begin{subfigure}[b]{0.45\textwidth}
    \centering
\begin{tikzpicture}[scale=0.8, grow'=up]
\Tree [.\vp{t'} [.\vp{s_1} \vp{John} $threw\_L$ ] [.\vp{s_2} \vp{ball} \vp{threw}\_R ]]
\end{tikzpicture}
\caption{New representation}
  \end{subfigure}
\caption{New and old representations of a thought}\label{fignewoldrep}
\end{figure}
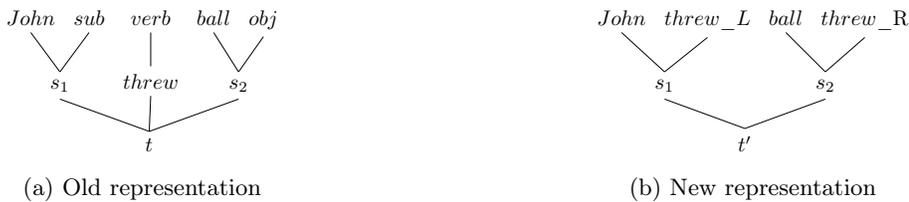

 Prior examples of thoughts have not been consistent in how verbs are represented in thoughts. From this point forward, a verb will not be represented with its own is-a branch on a thought tree. The verb will be split into its left and right hand parts. The two trees in figure~\ref{fignewoldrep} show the difference.
Note that with the new representation there is no explicit subject or object. These have been replaced with \vp{verb}\_L (subject side) and \vp{verb}\_R points (object side) respectively.  Section~\ref{secconsofthought} will demonstrate how these new points are used.
\subsection{Overview of the Language to Thought (LTT\label{def-LTT}) Process}\label{secLTT}
Converting a series of words into a thought is achieved with the opLUPARSE instruction. The execution of this instruction consists of several phases. A very simple sentence, `little boy threw ball', will be used as an example in the description of this process.
\begin{enumerate}
\item\label{lttphase1} The first phase is to load the AS with points from the PS. This is done on a point by point basis until the \vp{eoa} point is detected in the PS. For each point, a determination is made as to whether or not the point is grounded. If it is grounded then a new surrogate point is created, the point taken from the PS is is-a twined to the surrogate and the surrogate is appended to the AS. If the point pulled from the PS is not grounded then the corresponding left hand and right chiral points are appended to the AS, not the original point. As an example, if the PS contains the points $\begin {Bmatrix}little & boy & threw & ball & eoa\end{Bmatrix}$ with \vp{little}, \vp{boy} and \vp{ball} being grounded then the AS would contain five points: $\begin{Bmatrix} \frac{little}{s_1} & \frac{boy}{s_2} & threw\_L & threw\_R & \frac{ball}{s_3} \end{Bmatrix}$.
\item\label{lttphase2} The next phase is to look at all combinations of contiguous surrogates. For each combination, temporarily add the surrogates (with is-a points) to the PS and evaluate the binding [\vp{surAction}]. If the evaluation fails go on to the next contiguous pair. If the evaluation is successful then the result should be either the opJOIN or opLULINK points/opcodes. If it is one of these two points then execute the instruction for the pair of points. In this example, the first two points would be combined with the opJOIN instruction/point.  The AS would then be $\begin{Bmatrix} \frac{little \; boy}{s_1} &  threw\_L & threw\_R & \frac{ball}{s_3} \end{Bmatrix}$
\item\label{lttphase3} This phase looks to join left and right hand chiral points to adjacent surrogates. If a surrogate is followed with a left hand point then the left hand point is-a twined to the surrogate. Similarly if a right hand point is immediately followed with a surrogate then it is is-a twined to that surrogate unless the surrogate has already been is-a twined with a left handed point. If a right handed point is immediately followed with a left handed point then remove the left handed point from the AS.  The AS after this phase contains $\begin{Bmatrix} \frac{little \; boy \; threw_L}{s_1} & \frac{ball \; threw\_R}{s_3} \end{Bmatrix}$
\item If there are only surrogate points in the AS then continue phase \ref{lttphase5}. If there are other (chiral) points then continue with phase \ref{lttphase2} unless phases \ref{lttphase2} and \ref{lttphase3} have already been repeated with no actions taken. In this case remove any remaining chiral points from the AS and continue with phase \ref{lttphase5}. In our example, continue with phase \ref{lttphase5}.
\item\label{lttphase4} Perform one final check for contiguous surrogates (similar to phase \ref{lttphase2}).
\item\label{lttphase5} In the final phase is-a twine each remaining surrogate point in the AS to the current thought point (register rT) and remove all points from the AS. This example ends with the thought shown in figure~\ref{figlittleboy}. The thought point (\vp{t_x}) has two surrogate points is-a twined. Surrogate \vp{s_1} representing a point described by \vp{little} and \vp{boy}, and surrogate \vp{s_2} representing a point described by \vp{ball}.
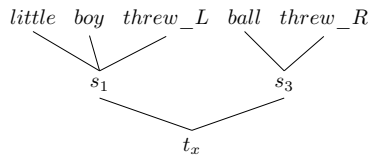
\begin{figure}[H]
\centering
\begin{tikzpicture}[scale=0.8, grow'=up]
\Tree [.$t_x$ [.$s_1$ \vp{little} \vp{boy} $threw\_L$ ] [.$s_3$ \vp{ball} $threw\_R$ ] ]
\end{tikzpicture}
\caption{Resulting thought tree from the sentence 'little boy threw ball'}\label{figlittleboy}
\end{figure}
\end{enumerate}
Below is a more complicated example with the sentence `The little boy threw the big red ball on the table to John':
\begin{lstlisting}
twine grounded<boy,ball,John,little,red,big,table
twine obj<boy,ball,John,objSurface
twine adjObj<little,red,big
twine objSurface<table
bind [surAction adjObj obj] opLUJOIN
bind [surAction obj objSurface on_R] opLULINK
ps the,little,boy,threw,the,big,red,ball,on,the,table,to,John,eoa
run opLUPARSE
\end{lstlisting}
The resulting thought is shown by the tree in figure~\ref{figlittleboyball}. The surrogate points created in the execution of the opLUPARSE instruction are shown as `sur\textit{n}?' where \textit{n} is a unique suffix for each surrogate. Note that the tree includes only surrogates 1, 3, 6 and 7. The other surrogates (2, 4, and 5) were merged with the opJOIN instruction.
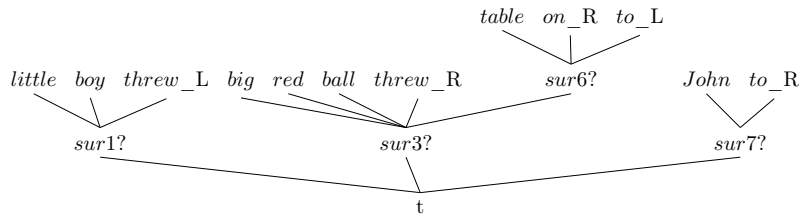
\begin{figure}[H]
\centering
\begin{tikzpicture}[scale=0.8, grow'=up ]
\Tree [.t [.\vp{sur1?} \vp{little} \vp{boy} \vp{threw}\_L ] [.\vp{sur3?} \vp{big} \vp{red} \vp{ball} \vp{threw}\_R  [.\vp{sur6?} \vp{table} \vp{on}\_R \vp{to}\_L ] ] [.\vp{sur7?} \vp{John} \vp{to}\_R ] ]
\end{tikzpicture}
\caption{Excerpted parsing tree of `The little boy threw the big red ball on the table to John'}\label{figlittleboyball}
\end{figure}
\subsection{Limitations of this Approach}
The parsing algorithm as just described is limited in its abilities to parse natural language. This is intentional as the next section (\ref{secLanLearn}) will describe how an intity can autonomously learn the bindings (rules) necessary to accomplish this level of linguistic sophistication. \par Linguistic parsing using the opRASM instruction does not have these two restrictions. See appendix~\ref{secParseopRASM} for several examples of parsing using opRASM. Merging this parsing sophistication into the LTT process is another area of ongoing research.
\section{How a Language is Learned}\label{secLanLearn}
How a natural language is learned is the topic of this section. A simplified subset of English consisting of nouns, adjectives, prepositions and verbs is used. Only simple noun phrases are considered; no nested clauses (e.g. `The man \textit{John saw at the park}'). Verb tenses are not considered. Even with these simplifications, language learning is a daunting task. Consider what a toddler acquires to understand this limited subset of English:
\begin{clist}
\item recognize and discriminate sounds into different points;
\item learn to recognize visual objects (e.g. form a pattern that recognizes apples);
\item learn the meaning of a word (the grounding problem) by associating sounds with physical objects (e.g. associate the spoken or written word `apple' with the visual pattern for an apple);
\item learn the syntax of the language (e.g. `John ate the cookie' is not the same as `The cookie ate John');
\item learn the meaning of thought (e.g. what does it mean to throw something, what are the consequences of throwing something). 
\end{clist}
The parsing process used in this section is the same as that described in section~\ref{secLTT}. The process requires that certain knowledge, in the form of bindings, be available to:
\begin{enumerate}
 \setlength\itemsep{-0.4em}
\item recognize words (i.e. patterns that recognize words from spoken sounds);
\item recognize grounded word (i.e. points that have is-twines to physical descriptions);
\item combine contiguous surrogate points in the AS using bindings of the form [$surAction$ $s_1$ $s_2$];
\item use the context of the resulting thought (all the is-a points linked to that thought) to find patterns resulting in consequential thoughts and/or actions.
\end{enumerate}
The following four sections address the four points just mentioned above. In all four cases the `learning' is just a form of pattern learning as discussed in section~\ref{subseclearnpat} albeit with different sets of pattern data. It should also be noted that although these four cases of learning are presented in a logical order, the actual learning need not be sequential.
\subsection{Learning to Recognize Words}
Spoken words enter the context as sensory input. Given sufficient repetition, the recognition process described in section~\ref{subseclearnpat} creates patterns that associate specific words with internal points. In this way, words are recognized. Embedded patterns (and points) are created distinguishing spoken words from other sounds, i.e. words have an is-a twine of the form \vp{spokenByPerson}<\vp{recognizedSound}.
\par Visual objects are recognized with the same learning process. The only difference between the learning of sounds versus visual objects is the source of the sensory points. All recognized visual points have a common embedded pattern that is recognized as point \vp{visualObj} (i.e. all visual points are is-a twined to \vp{visualObj}) and that is-a twine \vp{grounded}<\vp{visualObj}\footnote{This would be a predefined twine within the intity.} exists (i.e. all recognized visual points are grounded).
\subsection{Grounding: Associating Words with Visual Objects and Attributes}
\par Associating a spoken word point with its visual object point is again achieved with the pattern recognition algorithm. For the association to occur, both the word and object sensory points must  reside in the PS concurrently over multiple instances (e.g. the sounds for the word `apple' and the visual image of an apple). Given a word that is recognized with the following pattern/binding: [\vp{w_1} \vp{w_2} \DOTS \vp{w_n}]=\vp{word} and a visual object is recognized with the binding [\vp{v_1} \vp{v_2} \DOTS $v_{n'}$]=\vp{visObj} then if both the word is spoken and the corresponding object seen at the same time the context would contain \{\vp{w_1} \vp{w_2} \DOTS \vp{w_n} \vp{v_1} \vp{v_2} \DOTS \vp{v_{n'}}\} (in addition to other points). With enough repetitions, a new pattern would be detected= [\vp{w_1} \vp{w_2} \DOTS \vp{w_n} \vp{v_1} \vp{v_2} \DOTS \vp{v_{n'}}]=\vp{wvpoint}. However, the points comprising \vp{wvpoint} have embedded sub-patterns for both the word and visual objects resulting in the twines \vp{word}:\vp{wvpoint} and \vp{visObj}:\vp{wvpoint} (section~\ref{secTPwP}). 
\par Now if the PS contains $\begin{Bmatrix} \vp{w_1} \ \vp{w_2} \DOTS \vp{w_n} \end{Bmatrix}$ then \vp{word} is recognized resulting in \vp{word}.v being added to the PS (section~\ref{secRecPat}). The evaluation of [\vp{word}.v] results in \vp{wvpoint} and since \vp{visualObj}<\vp{wvpoint} and \vp{grounded}<\vp{visualObj} the resulting point is grounded.
\par Only words associated with visual objects would be grounded. Any other recognized words would have the is-a twine \vp{spokenByPerson}<\vp{otherWord}, but not be grounded.
\subsection{Combining Surrogate Points via Join and Link}
Once the AS is loaded with points (grounded and chiral) then only two operations are used to combine them. The V5 opJOIN instruction is used to merge two surrogate points into one surrogate by combining the is-a points of each (e.g. combine an adjective and noun). The opLULINK instruction is used to is-a twine one surrogate to another (e.g. link a prepositional phrase to a noun phrase). This section describes how an intity learns to join/link two surrogates together as in:
\begin{equation*}
\begin{Bmatrix}\frac{big}{s_1} \ \frac{red}{s_2} \ \frac{apple}{s_3}\end{Bmatrix} \Longrightarrow \begin{Bmatrix}\frac{big \; red \; apple}{s_1}\end{Bmatrix}
\end{equation*}
with bindings of the form [\vp{surAction} \vp{p_1} \vp{p_2} \DOTS] = opJOIN.
Initially an intity has no [\vp{surAction}] bindings. During the LTT process (section \ref{secLTT}) for each pair of contiguous surrogate points, the intity attempts to evaluate [\vp{surAction}]. If it fails, it looks at recent thoughts for points similar to the two surrogates in question. For example if the two surrogates are $\frac{big}{s_1}$ and $\frac{red}{s_2}$ then look through past thoughts (\vp{t} points) for is-a points having either the \vp{big} point or \vp{red} point. If the intity had recently noticed or thought of `big red apple' then there would be a thought with the branch $\frac{big \; red \; apple}{p_x}$. Both the \vp{big} point and the \vp{red} point have a common parent (in this case \vp{p_x}). The intity would then proceed and join the two surrogates in the AS as part of the LTT process. Additionally, the intity would save all the is-a points of the two LTT surrogates in a buffer for a future recognition run. When this buffer reached a certain number of entries or a certain amount of time has passed, the pattern learning process (section \ref{subseclearnpat}) would run. If a pattern was detected then the recognized points would be used to create a binding of the form [\vp{surAction} \vp{r_1} \vp{r_2} \DOTS \vp{r_n}]=opLUJOIN where \vp{r_i} are the recognized points. In this way the intity learns to join adjectives and nouns to form a single surrogate.
\par As an example, if the surrogate points $\frac{red}{s_1}$ and $\frac{apple}{s_2}$ (corresponding to the grounded words `red' and `apple') are repeatedly encountered in the AS and repeatedly combined into a single surrogate then multiple sets of \{\vp{red},\vp{apple},opLUJOIN\} would be appended to the pattern learning buffer. During the pattern learning phase this set of points would be recognized as a pattern and the binding [\vp{red} \vp{apple} \vp{surAction}]=\vp{opJOIN} would be created.
\par Success requires that the intity has a (previous) thought matching the spoken words. In other words it must have noticed a `big red ball' prior hearing the words `big', `red' and `ball'. As the intity learns more about the world it develops more sophisticated is-a twines and the patterns recognized for the surrogate action become more generalized.
\par Learning when to apply the opLULINK operation is handled in a similar fashion. In this case the reference thought is not a single point but two points with one an is-a of the other. The learning processes are identical, the only difference is the use of the opLULINK instruction instead of the opJOIN instruction.
\par The phrase `ball on table' maps into two surrogate points in the AS: \{$\frac{ball}{s_1},\frac{on \ table}{s_2}$\}. If the evaluation of [$surAction$] in the context of the two surrogates fails then a search of prior thoughts is performed looking for the match on the is-a points. Assuming the ball is perceived to be on the table then the thought $\frac{\frac{ball \ \frac{on \ p_{table}}{p_x}}{p_{apple}}}{t}$ exists. Matches would be found between the \vp{ball} and \vp{table} resulting in the set \{\vp{ball},\vp{table},\vp{on},\vp{opLULINK}\}. With enough repetitions a pattern would be found resulting in the binding [\vp{surAction} \vp{ball} \vp{table} \vp{on}] = \vp{opLULINK}.
\subsection{Meaning / Consequences of a Thought}\label{secconsofthought}
The \textit{meaning} of a thought is defined as the effect that thought has on subsequent actions and thoughts. How those subsequent actions/thoughts arise from a given thought is presented in this section. A general overview of the steps required is:
\begin{indent1}
\begin{itemize}
\item Look at pairs of thoughts that were both created within a specific time interval.
\item Create a set of points that relate the before and after points from the two thoughts and save these sets in a reserved pattern learning buffer.
\item After a certain number of sets have been created (or a specified amount of time has passed) perform the pattern learning process (section~\ref{subseclearnpat}).
\item If a pattern is found, use the matched points to create another twine of the form [\vp{cons}.v \vp{b_1} \vp{b_2} \DOTS]=\vp{consequences}, where \vp{b_x} are trigger points from the initial (before) thought) and \vp{consequences} are a series of points to create a consequential thought or initiate a consequential action/sequence.
\end{itemize}
\end{indent1}
The encoding of the before and after points is the key to making this a simple process of pattern learning.
\subsubsection{Action Verbs}
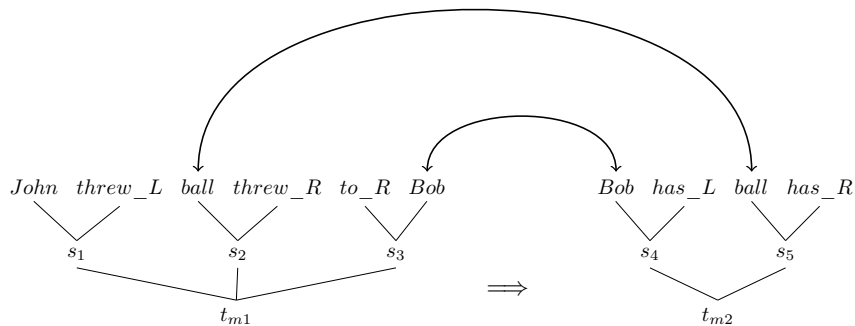
\begin{figure}[H]
\centering
\begin{tikzpicture}[scale=0.8, grow'=up]
\centering
\Tree [.$t_{m1}$  [.$s_1$ \vp{John} $threw\_L$ ] [.$s_2$ \node(ball1){\vp{ball}}; $threw\_R$ ] [.$s_3$ $to\_R$ \node(bob1){\vp{Bob}}; ] ]
\node[anchor=west,text width=4cm] (note1) at (4,.5) { $\Longrightarrow$ };
\begin{scope}[xshift=8cm]
\Tree [.$t_{m2}$ [.$s_4$ \node(bob2){\vp{Bob}}; $has\_L$ ]  [.$s_5$ \node(ball2){\vp{ball}}; $has\_R$ ] ]
\end{scope}
\draw[semithick, <-> ]  (bob1) to [bend left=90]  (bob2);
\draw[semithick, <-> ] (ball1) to [bend left=90]  (ball2);
\end{tikzpicture}
\caption{Two thoughts related by common points ($m1 < m2$)}\label{fig2thoughtscommon}
\end{figure}

Figure~\ref{fig2thoughtscommon} shows two thoughts. The first corresponds to `John threw the ball to Bob'. The second (subsequent) thought corresponds to `Bob has the ball'. These two thoughts are related by two pairs of surrogate points. Surrogates \vp{s_2} and \vp{s_5} both resolve to `the ball' and surrogates \vp{s_3} and \vp{s_4} resolve to `Bob'.  For each of these pairs of points, a binding is created with a new link point as the value. For the `ball' this becomes [\vp{threw}\_R \vp{has}\_R]=\vp{link1}. For `Bob' it is [\vp{to}\_R \vp{has}\_L]=\vp{link2}. Additional twines are created so that given either \vp{link1} or \vp{link2} the cause and effect points can be determined. This is reflected in lines~\ref{line:caex1a} through \ref{line:caex1b} in the example below. An additional link point is created for the surrogate point \vp{s_1}. Since this surrogate point is not related to any point in the second thought, the binding contains all of the surrogate is-a twine points. Again, an additional twine is created for just the \textit{cause} since there is no corresponding effect side. Line~\ref{line:caex1c} and \ref{line:caex1c2} show this third link. The pattern data generated by this pair of thoughts would be $\begin{Bmatrix} \vp{link1} \ \vp{link2} \ \vp{link3} \end{Bmatrix}$.
\par Now consider a similar pair of thoughts represented by the two thoughts: `Mary threw the bone to Fido' and `Fido has the bone'. In this case `bone' and `Fido' are identical. The links generated for this pair of thoughts would be \vp{link1}, \vp{link2} and \vp{link4}. The links \vp{link1} and \vp{link2} would be identical to the first pair because the binding key sets [\vp{threw}\_R \vp{has}\_R] and  [\vp{to}\_R \vp{has}\_L] are identical. But `John' is not the same as `Mary' so while [\vp{john} \vp{threw}\_L]=\vp{link3}, [\vp{mary} \vp{threw}\_L]=\vp{link4}. The pattern data for this pair of thoughts would be $\begin{Bmatrix} \vp{link1} \ \vp{link2} \ \vp{link4} \end{Bmatrix}$.
\par Given a sufficient number of thought pairs, the standard pattern matching algorithm would find the pattern \{\vp{link1},\vp{link2}\}. But rather than bind the pattern to a recognition point, a consequence would be created from the cause/effect is-a points of the pattern links. In this example (line~\ref{line:caex1d}) the trigger points are $to\_R$ and $threw\_R$ and the consequence is to create a new thought relating a \textit{value} to an \textit{attribute}.
\vspace{-15pt}
\centerline{(Note in above figure arrows linking Bob and ball are to the surrogate nodes, not to leaf points.)}
\begin{lstlisting}
bind [to_R has_L] link1(*\label{line:caex1a}*)
twine to_R<link1|cause ; has_L<link1|effect
bind [threw_R has_R] link2
twine threw_R<link2|cause ; has_R<link2|effect(*\label{line:caex1b}*)
bind [threw_L John] link3(*\label{line:caex1c}*)
twine threw_L,John<link3|cause(*\label{line:caex1c2}*)

twine cons>to_R.v,has_L,eoa,opSURISA,threw_R.v,has_R,eoa,opSURISA | to_R,threw_R(*\label{line:caex1d}*)
\end{lstlisting}
\begin{exDesc}
The consequence defined on line~\ref{line:caex1d} is tested with the thought representing the sentence `Jill threw the keychain to Helen' (line~\ref{line:caex1e}). The thought \vp{t1} is added to the PS (context). Then \vp{cons.v} is added and evaluated. The result are the twines shown below (lines~\ref{line:exHHB1} through \ref{line:exHHB2}) corresponding to figure~\ref{figConsKeychainHelen}; i.e. Helen has the keychain.
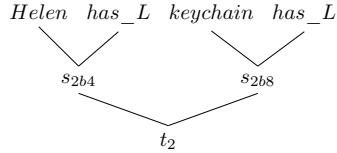
\begin{figure}[H]
\centering
\begin{tikzpicture}[scale=0.8, grow'=up]
\Tree [.$t_2$ [.$s_{2b4}$ \vp{Helen} $has\_L$ ] [.$s_{2b8}$ \vp{keychain} $has\_L$ ] ]
\end{tikzpicture}
\caption{Consequence of throwing a keychain to Helen}\label{figConsKeychainHelen}
\end{figure}
\end{exDesc}
\begin{lstlisting}[firstnumber=9]
twine Jill,keychain,Helen<t1(*\label{line:caex1e}*)
twine threw_L<Jill ; threw_R<keychain ; to_R<Helen

ps t1,opPSISAS,cons.v
trace bind
run
(*\vout{bind: \#48: [\#2b4?.i] = Helen}\label{line:exHHB1}*)
(*\vout{bind: \#49: [\#2b4?.i] = has\_L}*)
(*\vout{bind: \#50: [*T*(2).i] = \#2b4?}*)
(*\vout{bind: \#51: [\#2b8?.i] = keychain}*)
(*\vout{bind: \#52: [\#2b8?.i] = has\_R}*)
(*\vout{bind: \#53: [*T*(2).i] = \#2b8?}\label{line:exHHB2}*)
\end{lstlisting}
\begin{exDesc}
The next example shows how a consequence that is not based solely on linked nodes can be derived.
\begin{figure}[H]
\centering
\begin{tikzpicture}[scale=0.8, grow'=up]
\centering
\Tree [.$t_{m1}$  [.$s_1$ \vp{John} $threw\_L$ ] [.$s_2$ \node(ball1){\vp{ball}}; $threw\_R$ ] [.$s_3$ $to\_R$ \node(bob1){\vp{Bob}}; ] ]
\node[anchor=west,text width=4cm] (note1) at (4,.5) { $\Longrightarrow$ };
\begin{scope}[xshift=8cm]
\Tree [.$t_{m3}$ [.$s_6$ \node(ball2){\vp{ball}}; $flies\_L$ ]  [.$s_7$ $through\_R$  \vp{air} ] ]
\end{scope}
\draw[semithick, <-> ] (ball1) to [bend left=90]  (ball2);
\end{tikzpicture}
\caption{Example of two thoughts related by a common point ($m1 < m3$)}
\end{figure}
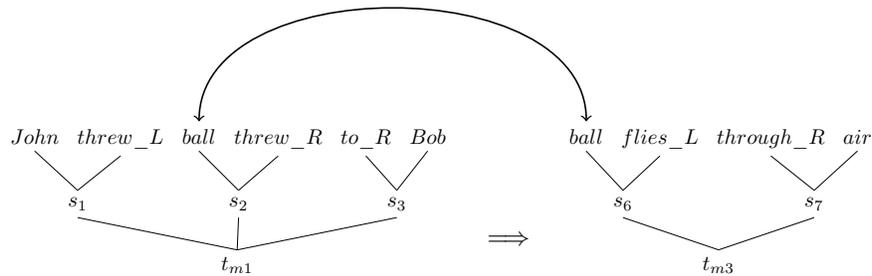
\end{exDesc}
The first thought is identical to that of the previous example. The second thought is `The ball flew through the air'. These two thoughts are inter-linked only once with the ball. This creates link point \vp{linka}. The other points are created from the other nodes: \vp{link3} would be same as from the above example, \vp{linkc} and \vp{linkd} would be different points. The pattern data from this pair of thoughts is $\begin{Bmatrix} \vp{linka}, \ \vp{link3}, \ \vp{linkc}, \ \vp{linkd} \end{Bmatrix}$. Other similar pairs of thoughts (`Harry threw the stick to his dog' `The stick flew through the air') would generate pattern data duplicating \vp{linka} and \vp{linkd} but other link points would vary. The pattern would be detected and the consequence (line~\ref{line:caex2a}) would be created.
\begin{lstlisting}[firstnumber=1]
bind [threw_R flew_L] linka
twine threw_R<linka|cause ; flew_L<linka|effect
bind [threw_L John] link3
twine threw_L,John<link3|cause
bind [to_R Bob] linkc
twine to_R,Bob<linkc|cause
bind [through_R air] linkd
twine through_R,air<linkd|effect

twine cons>threw_R.v,flew_L,eoa,opSURISA,through_R,air,eoa,opSURISA | threw_R(*\label{line:caex2a}*)
\end{lstlisting}
\begin{exDesc}
A new thought corresponding to `Harry threw the rock' (thought \vp{2} on line~\ref{line:caex2b}) can now be tested. The resulting (consequential) thought, `the rock flew through the air', is shown in figure~\ref{figConsThrowRock} and in the bindings starting at line~\ref{line:caex2c}.
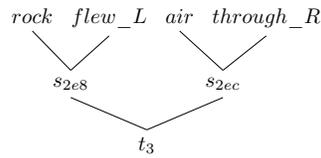
\begin{figure}[H]
\centering
\begin{tikzpicture}[scale=0.8, grow'=up]
\Tree [.$t_3$ [.$s_{2e8}$ \vp{rock} $flew\_L$ ] [.$s_{2ec}$ \vp{air} $through\_R$ ] ]
\end{tikzpicture}
\caption{Consequence of throwing a rock}\label{figConsThrowRock}
\end{figure}
\end{exDesc}
\begin{lstlisting}[firstnumber=11]
twine Harry,rock<t2 ; threw_L<Harry ; threw_R<rock(*\label{line:caex2b}*)
ps t2,opPSISAS,cons.v
trace bind
run
(*\vout{bind: \#71: [\#2e8?.i] = rock}\label{line:caex2c}*)
(*\vout{bind: \#72: [\#2e8?.i] = flew\_L}*)
(*\vout{bind: \#73: [*T*(3).i] = \#2e8?}*)
(*\vout{bind: \#74: [\#2ec?.i] = through\_R}*)
(*\vout{bind: \#75: [\#2ec?.i] = air}*)
(*\vout{bind: \#76: [*T*(3).i] = \#2ec?}*)
\end{lstlisting}
\subsubsection{Copula (To Be)}
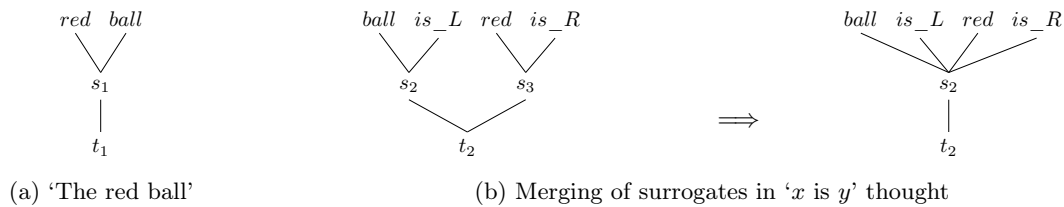
\begin{figure}[H]
\centering
\begin{subfigure}[b]{0.3\textwidth}
\centering
\begin{tikzpicture}[scale=0.8, grow'=up]
\Tree [.$t_1$  [.$s_1$ \vp{red} \vp{ball} ] ]
\end{tikzpicture}
\caption{`The red ball'}
\end{subfigure}
\begin{subfigure}[b]{0.6\textwidth}
\centering
\begin{tikzpicture}[scale=0.8, grow'=up]
\Tree [.$t_2$  [.$s_2$ \vp{ball} $is\_L$ ] [.$s_3$ \vp{red} $is\_R$ ] ]
\node[anchor=west,text width=4cm] (note1) at (4,.5) { $\Longrightarrow$ };
\begin{scope}[xshift=8cm]
\Tree [.$t_2$ [.$s_2$ \vp{ball} $is\_L$ \vp{red} $is\_R$ ] ]
\end{scope}
\end{tikzpicture}
\caption{Merging of surrogates in `\vp{x} is \vp{y}' thought}
\end{subfigure}
\caption{Similarity between `the red ball' and `the ball is red'}\label{fig2RedBalls}
\end{figure}

In our simplified language, the sentence `\vp{x} is \vp{y}' (e.g. `the ball is red') is very similar to the noun phrase 'the red ball' in that the thoughts generated by each would reduce to the same tree structure as shown in figure~\ref{fig2RedBalls}.
\par There is a subtle semantic difference between `the red ball' and `the ball is red'. In the first phrase, a single surrogate has both the `red' and `ball' is-a links. Resolution of the surrogate would require matching on both points. In the second phrase, the subject surrogate only has the \vp{ball} point. Resolution requires only that the target point also have the \vp{ball} point. After resolution, the additional \vp{red} point would be twined. There is an unresolved timing issue with the phrase `The ball is red'. What happens if the \vp{ball} surrogate is not resolved when the \vp{red} point is-a twined?
\section{Conclusions}
This paper introduced the contextual evaluation model for representing knowledge. The model is conceptually simple consisting of only four elements: the point, the key set, the binding and context. Contextual evaluation is the only operation. A single simple algorithm is used for learning patterns. Despite the simplicity of the model, it is shown to be Turing complete. More importantly, many examples demonstrate how the model performs a range of artificial intelligence tasks from the simple (learning a maze) to the difficult (learning a natural language). It is the hope of the author that it will interest others to investigate the potential of the contextual evaluation model.
\par The major contributions of this paper are:
\begin{itemize}
\item The CEM - the simplicity of the model and its ability to integrate internal, sensory and control data (points) into a contextual framework of facts, patterns and sequences.
\item V5 - proof that the CEM is viable and that with relatively few primitive instructions ($\approx$ 25) much can be accomplished.
\item The incorporation of time - how time naturally fits into the CEM model.
\item Patterns as bindings - how the CEM binding is used for pattern matching; how points of different modalities (e.g. sound, smell, vision, internal) can be incorporated into a single pattern and how focus points can be used to implement focus and direct \textit{attention}.
\item The \textit{thought}- defined as a collection of is-a twines linked to a base thought point; using sequences to convert thoughts to natural language utterances; converting natural language utterances to thoughts.
\item Meaning - the \textit{meaning} of a thought, not as anything inherent in the thought but as the subsequent actions of an intity as a consequence of the thought. 
\item Motivation - the \vp{p^+} and \vp{p^-} points to represent good/pleasure and bad/pain and the effect these points have on the operation of the V5 engine.
\item Natural Language Acquisition - that each of the steps in autonomous language acquisition is accomplished with the simple tools of this model.
\end{itemize}
\subsection{Major Unanswered Questions}
The CEM provides a concise methodology for handling many disparate AI situations. The CEM and V5 are works in progress and much remains to be resolved:
\begin{itemize}
\item How are sense points encoded? The CEM assumes that all senses are encoded such that pattern learning and pattern recognition are possible using the techniques discussed in section~\ref{secPatterns}. Could it be that a neural-network pre-processor is required to convert raw sensory data into a point based format?
\item How would the V5 engine scale to perform useful tasks? Despite decades of experience working with an earlier contextual engine, the effort required to scale up the V5 engine is an unknown.
\item Would the language parsing and generation techniques scale up to handle a more complete language subset? This paper describes only the simplest forms of language parsing and generation for English. Can additional rules (bindings) be developed to handle complex English? Would similar rules or completely different rules be required for the majority of languages where syntax is not based on word position but word endings?
\item How are surrogate points resolved? Only the simplest rules for surrogate resolution were given in this paper. How should surrogates resolve that are qualified with multiple points and thoughts? When should surrogates be resolved? Once resolved should they continue to be re-resolved?
\item Does the CEM reflect, at some level, how brains actually operate? And if so then how did it evolve? What can be inferred from the similarities between viewing the E\sub{C} engine as a set of rules (bindings) operating within a constantly changing context and a living cell having a set of rules (DNA/RNA) operating within a constantly changing environment (context)?
\item How are sequences learned? The methods described in section~\ref{secconsofthought} for learning consequences of a thought would be one approach, albeit slow and inefficient for any sequence consisting of more than a few points. Another possible approach is to have a basic sequence scaffolding or framework on which points of a sequence could be twined.
\end{itemize}
\par As with everything else in science and technology, the ideas presented in this paper stand on the shoulders of many others in a diverse range of AI endeavors. The following sections of the conclusion point out some of the many areas of prior research utilized in developing the CEM and V5 engine.
\subsection{Other Models of Knowledge Representation}
Knowledge representation (KR) is the represention of information for solving non-trivial problems with a capability similar to that of a human. First order logic has been a key component of KR systems since the late 1950's with the IPL/GPS\cite{gps} programs. IPL introduced the concept of the \textit{list}. McCarthy expanded on that concept, added ideas from lambda calculus and developed LISP\cite{lisp,lisp2}.
\par First order logic, while mathematically rigorous, was found to be difficult to apply to many real world situations. Other KR models moved from strictly declarative structures to formats allowing for procedural and frame-based formats\cite{planner,wiki:Prolog,frame,kl-one}. Larger, ongoing KR projects have expanded on the idea of semantic webs\cite{wiki:Cyc,cycSyntax,owl}.
\subsection{Comparison of the CEM Context with Other Contextual Models}
\par Intensional logic is a group of formal systems allowing expressions or statements whose values depend on hidden contexts. It was originally developed to help understand the contextual nature of natural languages. In this logic, the extension of an expression is the value in a specific context. The intension of an expression is the function that maps from a context the value of each expression within that context\cite{wadge}.
\par Lucid is a programming language developed in this format\cite{lucid}. Early versions of the language supported only time and space as the context. Current versions permit the users to define their own dimensions. In Lucid, the context and the expressions are separate, while in the CEM, the context consists of the same points as the other components of the model. The expressions within Lucid are based on variables, functions, constants, and operations. These can be assembled into programs. The CEM has no explicit variables, functions, or constants, just bindings. This is not to say that the equivalent of variables, functions and constants cannot be implemented with the CEM, but that there is nothing intrinsic to the model implementing these concepts. There is no concept of a program within the CEM yet sophisticated programming can be achieved through contextual sequences.
\par McCarthy  and later Guha\cite{guha}  attempted to define context through the $ist(c,p)$ relation that asserts that $p$ is true in the context of $c$. In this model the
\enquote{Contexts are abstract objects. We don't offer a definition}.\cite{mccarthy} The propositions, p's, are expressions based on the predicate logic. As an example, McCarthy gives the following: 
\begin{equation}
ist(context-of(`Sherlock Holmes Stories'), `Holmes is a detective')
\end{equation}
 that states, in the context of Sherlock Holmes stories, the character Holmes is a detective. In another context, Holmes might be a short order cook or somebody's dog.
\par Lenat further extends this thinking in the Cyc project by defining contexts within a 12 dimensional space.  As with McCarthy and Guha, Cyc facts and propositions are stated within a context. There does not appear to be any overlap between context and proposition. A comparison of contexts based (or derivative) of the McCarthy model can be found in Akman\cite{akman}.  Another review of contextual models can be found in Serafini\cite{serafini}.
\subsection{V5 Language and Engine}
V5 has its roots in the V4 language\cite{patent:6470490, v4Ref}. V4 was started in the mid-1990s and is currently used as a production language for web-based data analysis. V4 is based on points, bindings and contextual evaluations. It differs from V5 in several significant ways:
\begin{enumerate}
\item In V4, every point belongs to a \textit{dimension}. Dimensions are typed and supported types include integer, floating point, logical, string, date-time and dictionary. A point is specified as \textit{dimension:value} as in date:23-Apr-2015.
\item V4 supports \textit{modules} that perform typical language functions such as string manipulation, mathematical and statistical analysis and input/output.
\item The V4 context is `stacked' meaning that a new stack \textit{frame} is created with each execution frame. Points inserted into a frame take precedence over points in prior frames. Context frames are removed when the execution frame is exited. Recursive binding definitions are thus supported. The factorial function for integers greater than zero could be defined as shown in (\ref{equV4Fact})\footnote{With some syntactic sugaring, points on the NId dimension (named identifier dimension) can be written without the dimension prefix (e.g. NId:factorial $\equiv$ factorial). Similarly, integers can be written without the Int dimension prefix (e.g. Int:1 $\equiv$ 1). A dimension with an asterisk suffix references the value of that dimension in the context. Arithmetic expressions are enclosed in braces. If the evaluation of a key set fails and the key set is followed with a comma and another point then that point is taken as the value (e.g. evaluating [factorial 0],1 returns 1).}.
\begin{equation}\label{equV4Fact}\text{[factorial Int:>0] = \{Int* * [factorial \{Int* - 1\}],1\}}\end{equation}
\item V4 may be programmed much like any other functional language. The binding in figure~\ref{figHelloWorld}, when evaluated, generates HTML to display `Hello World'.
\begin{figure}[H]
\centering
\begin{subfigure}[b]{0.4\textwidth}
\begin{lstlisting}
[HelloWorld] =
 Do(XML::html XML::body
     Echo(XML::h1 "Hello World")
    )
\end{lstlisting}
\caption{V4 code to generate `Hello World' HTML page}
\end{subfigure}
\begin{subfigure}[b]{0.4\textwidth}
\begin{lstlisting}[numbers=none]
<html>
 <body>
  <h1>Hello World</h1>
 </body>
</html>
\end{lstlisting}
\caption{HTML generated by [HelloWorld] binding}
\end{subfigure}
\caption{Hello World example}\label{figHelloWorld}
\end{figure}
\end{enumerate}
What V4 has demonstrated over the years is that an efficient algorithm for contextual evaluations exists and the binding / contextual evaluation paradigm is a practical tool for real world problems.  The author is not aware of any work on contextual sequences other than his own\cite{hansen:isic:98}.
\subsection{Language, Thought and Meaning}
In this paper, a \textit{thought} is defined in section~\ref{secThoughts} as a point with is-a twines linked to it. While the example thoughts presented throughout this paper all had points with English labels, the thought itself, is not tied to any language. Several sections of this paper are devoted to mapping from a thought to a natural language utterance and from an utterance to a thought. This begs the question, `Is language necessary for thought?' There are two philosophical schools on this issue. Lingualism, most notably the Sapir-Whorf hypothesis, claims that language, to a greater or lesser degree, determines thoughts\cite{wiki:Linguistic_relativity}. Another school, the `language of thought hypothesis' (LOTH)\cite{fodor,wiki:Language_of_thought_hypothesis}, claims that thought is independent of language. The author tends more towards the LOTH school, not because of any philosophical reasons but because of empirical studies showing that people with aphasia (an inability to use language) still possess other mental faculties\cite{fedorenko}.
\par There are any number of theories related to the \textit{meaning} of language\cite{wiki:Meaning-philosophy-of-language}. In this paper \textit{meaning} arises from the consequences of the thought. A thought, in and of itself, does not have meaning. Meaning is an ongoing process, not a static attribute or quality. Searle in his book \textit{The Language of Thought} makes a somewhat similar argument\cite[p. 120]{searle}:
\blockquote{\DOTS I shall argue that for a large number of cases the notion of literal meaning of a sentence only has application relative to a set of background assumptions, and furthermore these background assumptions are not  all and could not all be realized in the semantic structure of the sentence in the way that presupposition and indexically dependent elements of the sentence's truth conditions are realized in the semantic structure of the sentence.}
\par The language learning process presented in this paper differs from many of the existing models\cite{frank,wintner} in several respects. First is that it assumes no a priori knowledge and no external `guidance'. Secondly it spans the learning process from distinguishing words from other sounds to learning the meanings of utterances (as mapped to thoughts). And thirdly it does so in a logical, easily explained way using no special techniques other than bindings and pattern learning.
\subsection{Motivation}
The point types \vp{p^+} and \vp{p^-} are used to differentiate good/pleasure and bad/pain points from other points. These +/- points impact the operation of an intity in such a way as to maximize \SIGMA \ thus favoring good over bad. Given that the intity reacts appropriately to the +/- points, does the intity \textit{feel} the points? Others have asked this very question. For Dennet it is the notion of qualia\cite{dennett}. For Chalmers it is the discussion of philosophical zombies\cite{wiki:Philosophical_zombie}. This question is best left for the philosophers.
\par Emotions are an important component to human level intelligence. Thirty years ago Minsky recognized the importance of emotions: \enquote{The question is not whether intelligent machines can have any emotions but whether machines can be intelligent without any emotions.}\cite[p.163]{minsky} Picard writes in her book \textit{Affective Computing}\cite{picard}: \enquote{The latest scientific findings indicate that emotions play an essential role in rational decision making, perception, learning, and a variety of other cognitive functions}. Do the +/- points have any relationship to emotions? Could \vp{p^+} points unrelated to senses (i.e. not associated with a sensory experience) be related to the feelings of happiness or contentment? Similarly, could \vp{p^-} points unrelated to senses be the root of fear and/or anxiety? Recall that the intity is constantly matching patterns in the PS. What if the PS is suddenly swamped with a novel set of points and was unable to recognize anything in the set? Would it momentarily \textit{freeze up} in a way that is similar to our reaction of surprise?
\vspace{1cm}
\par This paper is all about models: modeling knowledge using points and bindings, modeling thoughts as trees of is-a twines, modeling language as a medium for communicating thoughts, modeling meaning as a process, modeling motivation with \vp{p^{+/-}} points. Are these models valid? Might they reflect how we actually think? To paraphrase George Box\cite{box}, all models are wrong, but some are useful. My hope is that some will find this paper useful.



\bibliographystyle{acm}
\bibliography{contextModel}

\newpage
\appendix
\section{Glossary of Terms and Symbols}\label{secGlossary}
The following table lists the major terms and symbols used within this paper. The page number indicates where the term/symbol first appears.
\begin{longtable}{c r p{.75\textwidth}}
\textbf{Item}&\textbf{Page}&\textbf{Short Description}\\
\hline\endhead
\rule{0pt}{3ex} AS&\pageref{def-AS}&The aggregate set used with the opRAS and opRASM instructions.\\
\symB&\pageref{def-B}&The binding set where all bindings reside. \\
binding&\pageref{def-binding}&The associating of a non-null key set to a value point: [\vp{p_1} \vp{p_1} \DOTS \vp{p_n}] = \vp{p_{value}}.\\
\vp{c_x}& &A point residing in the context set \symC.\\
\symC&\pageref{def-C}&A collection of all points comprising the context.\\
E\sub{C}&\pageref{def-EC}&The contextual evaluation engine.\\
CEM&\pageref{def-CEM}&The abbreviation of the \underline{C}ontextual \underline{E}valuation \underline{M}odel.\\
chiral&\pageref{def-chiral}&A non-grounded point which is split into left hand and right hand parts.\\
complete&\pageref{def-complete}&Evaluating a key set where the points in the key set exactly match the points in the binding.\\
context&\pageref{def-context}&An independent collection of points used with a contextual evaluation.\\
\DELTA&\pageref{def-Delta}&\SIGMA - \SIGMABAR\\
E\sub{C}&\pageref{def-EC}&The contextual evaluation engine.\\
\vp{f} & \pageref{def-f} & A focus point used in determining whether or not a pattern is active. \\
fractional&\pageref{def-fractional}&The representation of is-a twines as fractions: $\frac{isa-point}{base-point}$.\\
grounded&\pageref{def-grounded}&A point that is \textit{grounded} and can be related to a physical (sensed) object or property.\\
\symI&\pageref{def-I}&The intelligence function that in this paper is defined as contextual evaluation.\\
incomplete&\pageref{def-incomplete}&Evaluating a key set where some of the points required to match a binding are taken from the context.\\
intity&\pageref{def-intity}&A contraction of \textit{\underline{int}elligent}+\textit{ent\underline{ity}} meaning an intelligent natural or man-made entity.\\
key set&\pageref{def-keyset}&A collection of 0 or more points enclosed with brackets ([\vp{p_1} \vp{p_2} \DOTS \vp{p_n}].\\
LTT&\pageref{def-LTT}&The language-to-thought process.\\
\vp{m_x}&\pageref{def-m}&A moment (in time) point that is often represented as \vp{m(num}) where $num$ is a time value.\\
\PATMINO&\pageref{def-minoccurs}&The minimum number of times a set of points has to occur in a pattern set (\symS) to be considered a pattern.\\
\PATMINP&\pageref{def-minpoints}&The minimum number of points in a set to be considered for inclusion as a pattern.\\
MR&\pageref{def-MR}&A miscellaneous collection of registers used by the V5 engine (see appendix~\ref{secregisters}).\\
PPU&\pageref{def-PPU}&The point processing unit is the V5 equivalent of a CPU.\\
PQ&\pageref{def-PQ}&The print queue that is used for queuing points for output to the console via the opOUTPQ instruction.\\
PS&\pageref{def-PS}&A point set.\\
\vp{s_x}&\pageref{def-s}&A surrogate point.\\
\symS&\pageref{def-S}&The set of points (i.e. `raw material') used for finding patterns.\\
sequence&\pageref{def-sequence}&A tightly coupled ordered collection of points or loosely coupled collection of patterns defining a goal.\\
\SIGMA&\pageref{def-Sigma}&The sum of \vp{p^+} and \vp{p^-} points.\\
\SIGMABAR&\pageref{def-SigmaBar}&The weighted average of \SIGMA.\\
\vp{t}&\pageref{def-t}&A thought point.\\
TC&\pageref{def-TC}&The twine context used when creating twines via various V5 instructions.\\
tree&\pageref{def-tree}&The representation of is-a twines as a graphical tree.\\
twine&\pageref{def-twine}&A binding containing either a value or is-a variant in the key set.\\
variant&\pageref{def-variant}&A variation of a normal point, either \vp{p.i} for is-a and \vp{p.v} for value-of.\\
\vp{w}&\pageref{def-w}&The binding weight associated with a binding.\\
\end{longtable}
\newpage
\section{V5 Commands, Opcodes, Registers and Reserved Words}\label{secV5Stuff}
\subsection{V5 Commands}\label{secV5Commands}
\begin{indent1}
\begin{labeledpar}{twine}{bind}
Creates a binding consisting of a key set and value. Note that the equal sign between the key set and value is optional.
\begin{itemize}
\item[] bind [$set \ of \ points$] = \vp{value}
\item[] bind [$set \ of \ points$] \vp{value}
\end{itemize}
\end{labeledpar}
\begin{labeledpar}{twine}{def}
Defines one or more labeled points. Multiple points are separated with commas.
\begin{itemize}
\item[] def \textit{\vp{point1}, \vp{point2}, \DOTS, \vp{pointn}}
\end{itemize}
\end{labeledpar}
\begin{labeledpar}{twine}{eval}
This command evaluates its argument and outputs the value to the console. All evaluations are with respect to the context (PS).
\begin{itemize}
\item[] eval \textit{key set} - evaluates the key set and prints the value
\item[] eval \vp{point.v} - evaluates [\vp{point.v}] and prints the value
\item[] eval \vp{point.i} - determines and outputs is-a points for \vp{point} but no is-a points off of those points
\item[] eval \textit{opcode} - executes the opcode, nothing is printed except as a result of the execution of the instruction
\item[] eval \textit{register} - prints the contents of the named register
\item[] eval \textit{sur} - prints the resolution of the surrogate point
\end{itemize}
\end{labeledpar}
\begin{labeledpar}{twine}{run}
Begin execution with the current contents of the PS.
\end{labeledpar}
\begin{labeledpar}{twine}{set}
Set various V5 internal paramters.
\begin{itemize}
\item[] set autodef \{on,off\} - if enabled then any referenced undefined point is automatically defined.
\item[] set newthought - creates a new thought point and copies it to register rT.
\end{itemize}
\end{labeledpar}
\begin{labeledpar}{twine}{show}
A debugging command to show various V5 sets and states.
\begin{itemize}
\item[] show aggset - outputs the points currently in the AS (aggregate set)
\item[] show opcodes - outputs the number of times each opcode has been executed
\item[] show ps - outputs the content of the PS
\item[] show tcs - outputs the contents of the twine context
\end{itemize}
\end{labeledpar}
\begin{labeledpar}{twine}{trace}
Control various trace/debugging features.
\begin{itemize}
\item[] trace autodef - output a trace of all points automatically created (without the def command)
\item[] trace binding - trace all created bindings
\item[] trace eval - trace all evaluations
\item[] trace off - turn off all tracing
\item[] trace op\textsl{\textsc{xxxx}} - trace specific opcode execution
\item[] trace reduce - trace the operation of the agg set (AS) reduction operation
\item[] trace sequence \vp{pattern} - trace the evaluation of any [\vp{p.v}] where the label associated with \vp{p} begins with \vp{pattern} (in the Turing machine example (section~\ref{secTM}), the lines \ref{line:tm8fail}-\ref{line:tm8failb} could be traced with the command `trace sequence tm8')
\item[] trace twine - trace the creation of any twine
\end{itemize}
\end{labeledpar}
\begin{labeledpar}{twine}{twine}
Creates one or more twines. Multiple twines can be entered delimited with semicolons. Multiple points can be separately twined to a single point with the construct: 
\begin{itemize}
\item[] twine \vp{p1}>\vp{p2} - create a value twine [\vp{p1}.v] = \vp{p2}
\item[] twine \vp{p1}>\vp{p2}|\vp{p_3},\DOTS,\vp{p_n} - create a value twine with additional point(s) - [\vp{p1}.v \vp{p_3} \DOTS \vp{p_n}]=\vp{p2}
\item[] twine \vp{p1}<\vp{p2} - create an is-a twine [\vp{p2}.i] = \vp{p1}
\item[] twine \vp{p1}<\vp{p2}|\vp{p_3},\DOTS,\vp{p_n} - create an is-a twine [\vp{p2}.i \vp{p_3} \DOTS \vp{p_n}] = \vp{p1}
\item[] twine \vp{p1}<\vp{p2},\vp{p3},\DOTS - create multiple is-a twines each with value of \vp{p1}
\item[] twine \vp{p1}<\vp{p2} ; \vp{p3}>\vp{p4} ; \DOTS - multiple twines may be delimited with semicolons
\end{itemize}
\end{labeledpar}
\end{indent1}
\subsection{V5 Opcodes}\label{secopcodes}
\begin{labeledpar}{opISATOAS}{opACTX}
Pops the head (top) point in the PS and adds it to the current twine context set (TC). On line~\ref{line:exopACTX} the opACTX instruction adds point \vp{a} to TC. The opTWVAL instruction twines \vp{b} to \vp{c} and includes point \vp{a} in the twine key set as shown by the trace output.
\begin{lstlisting}
def a,b,c
trace opTWVAL
ps a,opACTX,b,c,opTWVAL(*\label{line:exopACTX}*)
run
(*\vout{opTWVAL: \#32: [a b.v rCTP(1)] = c}*)
\end{lstlisting}
\end{labeledpar}
\begin{labeledpar}{opISATOAS}{opADDPQ}
Pops the head point in the PS and appends it to the print queue set (PQ).
\end{labeledpar}
\begin{labeledpar}{opISATOAS}{opBIND}
Pops points off the head of the PS until the \vp{eoa} point is hit. These points are inserted into a key set.
Then the next point after the \vp{eoa} point is popped off of the PS and a binding is created linking the key set to this point. The trace output on line~\ref{line:opbindout} shows the created binding. The number after the crosshatch (\#) is the internal binding reference number.
\begin{lstlisting}
def a,b,c,d
trace bind
ps a,b,c,eoa,d,opBIND
run
(*\vout{bind: \#32: [a b c] = d}\label{line:opbindout}*)
\end{lstlisting}
\end{labeledpar}
\begin{labeledpar}{opISATOAS}{opCLRPQ}
Removes (clears) all points from the PQ set.
\end{labeledpar}
\begin{labeledpar}{opISATOAS}{opEVAL}
Pops points off of the PS until the \vp{eoa} point is hit. These points are inserted into a key set. The key set is evaluated within the context of the PS. If the evaluation succeeds the result point is inserted into the rEVAL register. If the evaluation fails rEVAL is set to \vp{null} and the point \vp{evalFail} is pushed onto the PS. In the example below assume the binding [\vp{a} \vp{b} \vp{c}]=\vp{d}. The first example run (line~\ref{line:opevalfail}) fails evaluating [\vp{a} \vp{b}]. Note that \vp{evalFail} was added to the PS. On line~\ref{line:opevalok} the evaluation of [\vp{a} \vp{b} \vp{c}] succeeds and the PS is empty and register rEVAL contains point \vp{d } (line~\ref{line:opevalres}).
\begin{lstlisting}
trace opEVAL
ps a,b,eoa,opEVAL
run(*\label{line:opevalfail}*)
(*\vout{opEVAL: [a b] failed, ps: *empty*}*)
show ps
(*\vout{ps: evalFail}*)
ps a,b,c,eoa,opEVAL
run(*\label{line:opevalok}*)
(*\vout{opEVAL: [a b c] => \#32: [a b c] = d}*)
show ps
(*\vout{ps: *empty*}*)
ps rEVAL
run(*\label{line:opevalres}*)
show ps
(*\vout{ps: d}*)
\end{lstlisting}
\end{labeledpar}
\begin{labeledpar}{opISATOAS}{opINCCTP}
Increments the current time point by 1. This is used to force a change in the time point before it would normally be updated by the internal V5 wall clock.
\end{labeledpar}
\begin{labeledpar}{opISATOAS}{opINCT}
Copies the current time point into the twine current time point and increments it by one. If there is a sequence of consecutive opINCT opcodes then the twine current time point is incremented for each opcode instance.
\end{labeledpar}
\begin{labeledpar}{opISATOAS}{opISATOAS}
Pops the first point off of the PS, recalculates all is-a points for that point and appends the is-a points to the AS. After all is-a points are appended it appends the \vp{asPH}\label{def-asPH} point.
\begin{lstlisting}
twine big,red,ball<brb
ps brb,opISATOAS,opSTATE
run
(*\vout{state: ps: *empty*}*)
(*\vout{aggset: big, red, ball, asPH}*)
(*\vout{  pq: *empty*}*)
(*\vout{  prNEW: *none*}*)
(*\vout{  prEVAL: *none*}*)
(*\vout{  tCTX: *empty*}*)
(*\vout{  ctp: *CTP*(2)}*)
(*\vout{  tCTP: *CTP*(1)}*)
\end{lstlisting}
\end{labeledpar}
\begin{labeledpar}{opISATOAS}{opLASM}
Pops points off of the PS and appends them to the AS until the \vp{eoa} point is reached.
\begin{lstlisting}
ps a,b,c,d,eoa
run opLASM
show AS
(*\vout{aggset: a, b, c, d}*)
\end{lstlisting}
\end{labeledpar}
\begin{labeledpar}{opISATOAS}{opLUJOIN}
This opcode is only valid when run within the opLUPARSE operation. It combines two adjacent surrogate points in the AS into one surrogate point.
\begin{equation*}
\frac{a b}{s_1} + \frac{c d}{s_2} \Longrightarrow \frac{a b c d}{s_1}
\end{equation*}
\end{labeledpar}
\begin{labeledpar}{opISATOAS}{opLULINK}
This opcode is only valid when run within the opLUPARSE operation. It combines two adjacent surrogate points by twining the second as an is-a of the first (e.g. \vp{p_1} followed by \vp{p_2} twines \vp{p_2}<\vp{p_1}).
\begin{equation*}
\frac{a b}{s_1} + \frac{c d}{s_2} \Longrightarrow \frac{a b \frac{c d}{s_2}}{s_1}
\end{equation*}
\end{labeledpar}
\begin{labeledpar}{opISATOAS}{opLUPARSE}
The opcode first clears the AS. It then pops points off of the PS until the eoa point is reached. For each \vp{point} it determines if the \vp{point} is grounded (i.e. grounded<\vp{point}). If so then it creates a new surrogate point, twines the point to the surrogate (\vp{point}<\vp{surrogate}) and appends the surrogate to the AS. If the \vp{point} is not grounded then it appends the left and right chiral points to the AS (\vp{point_L} and \vp{point_R}). When all points have been popped off the PS, the language parsing routine as described in section~\ref{secLan2Thought} is performed.
\end{labeledpar}
\begin{labeledpar}{opISATOAS}{opLW1}
Pops the top point from the PS and saves it in working memory location 1. It can be referenced via register rWM1.
\begin{lstlisting}
eval rWM1
(*\vout{? Internal register (op=507) has no value}*)
def a
ps a,opLWM1
run
V5>show ps
(*\vout{ps: *empty*}*)
eval rWM1
(*\vout{ Register rWM1 = a}*)
\end{lstlisting}
\end{labeledpar}
\begin{labeledpar}{opISATOAS}{opPSISAS}
Recalculate all is-a points off of the points currently in the PS.
\end{labeledpar}
\begin{labeledpar}{opISATOAS}{opNEW}
Creates a new point and copies it to register rNEW.
\begin{lstlisting}
eval rNEW
(*\vout{? Internal register (op=502) has no value}*)
eval opNEW
eval rNEW
(*\vout{  Register rNEW = \#270}*)
ps opNEW,rNEW
run
show ps
(*\vout{ps: \#274}*)
\end{lstlisting}
\end{labeledpar}
\begin{labeledpar}{opISATOAS}{opNEWSUR}
This instruction is identical to opNEW except that it creates a new surrogate point.
\end{labeledpar}
\begin{labeledpar}{opISATOAS}{opNEWT}
Creates a new \vp{t} (thought) point and copies to the rT register. The engine automatically creates an initial thought point at startup (line~\ref{line:newt1}).
\begin{lstlisting}
eval rT
(*\vout{  Register rT = *T*(1)}\label{line:newt1}*)
eval opNEWT
eval  rT
(*\vout{  Register rT = *T*(2)}*)
\end{lstlisting}
\end{labeledpar}
\begin{labeledpar}{opISATOAS}{opOUTPQ}
Outputs the contents of the print queue (PQ) to the console. The output is prefaced with `PQ(\textit{m})' where \textit{m} is the value of the current-time point.
\begin{lstlisting}
def a,b,c
ps a,b,c,opADDPQ,opADDPQ,opADDPQ,opOUTPQ
run
(*\vout{PQ(1): a b c}*)
\end{lstlisting}
\end{labeledpar}
\begin{labeledpar}{opISATOAS}{opRASM}
This instruction performs multiple cycles of aggregate set (AS) reduction until a failure occurs. See section~\ref{secASReduce} for more details.
\end{labeledpar}
\begin{labeledpar}{opISATOAS}{opSURISA}
This instruction creates a new surrogate point and then pops off points from the PS until the \vp{eoa} point is detected. Each popped point is is-a twined to the new surrogate point. If rT contains a valid thought point (i.e. opNEWT been executed) then the surrogate is additionally is-a twined to the thought point.
\begin{lstlisting}
def a,b,c,d
trace bind
ps opNEWT,a,b,eoa,opSURISA,c,d,eoa,opSURISA
run
(*\vout{bind: \#32: [\#284?.i] = a}*)
(*\vout{bind: \#33: [\#284?.i] = b}*)
(*\vout{bind: \#34: [*T*(2).i] = \#284?}*)
(*\vout{bind: \#35: [\#288?.i] = c}*)
(*\vout{bind: \#36: [\#288?.i] = d}*)
(*\vout{bind: \#37: [*T*(2).i] = \#288?}*)
\end{lstlisting}
\end{labeledpar}
\begin{labeledpar}{opISATOAS}{opPSISAS}
(Re)calculates all is-A points for all points currently in the PS.
\begin{lstlisting}
def dog,animal,barney,cat,fluffy
twine animal<dog ; dog<barney
twine animal<cat ; cat<fluffy
ps barney,fluffy,opPSISAS
show ps
(*\vout{ps: barney, fluffy, opPSISAS}*)
run
show ps
(*\vout{ps: barney+<dog,animal>, fluffy+<cat,animal>}*)
\end{lstlisting}
\end{labeledpar}
\begin{labeledpar}{opISATOAS}{opRCTX}
Removes all points from the twine context set (TC).
\end{labeledpar}
\begin{labeledpar}{opISATOAS}{opSTATE}
This is a debugging tool that outputs a short summary of all current V5 states to the console. The example below shows the PS is empty, the PQ (print queue) is empty, no value for rNEW, no value for rEVAL, the twine context set (tCTX) is empty, the current time point is 220 and the twine time point is 1.
\begin{lstlisting}
eval opSTATE
(*\vout{state: ps: *empty*}*)
(*\vout{  pq: *empty*}*)
(*\vout{  prNEW: *none*}*)
(*\vout{  prEVAL: *none*}*)
(*\vout{  tCTX: *empty*}*)
(*\vout{  ctp: *CTP*(220)}*)
(*\vout{  tCTP: *CTP*(1)}*)
\end{lstlisting}
\end{labeledpar}
\begin{labeledpar}{opISATOAS}{opTWVAL}
Pops the top two points off of the PS. The second point is twined as the value to the first point. Any points in the twine context set are included in the twine.
\begin{lstlisting}
trace opTWVAL
ps a,b,opTWVAL
show ps
(*\vout{ps: a, b, opTWVAL}*)
run
(*\vout{opTWVAL: \#143: [a.v rCTP(122)] = b}*)
\end{lstlisting}
\end{labeledpar}
\begin{labeledpar}{opISATOAS}{opVAL}
Pops the first point off of the PS, converts it to its value variant and repushes back onto the PS. The first run (line~\ref{line:optwval1}) pops \vp{a} off of the PS, converts it to \vp{a.v} and pushes it back onto the PS. On the second pass the engine sees the \vp{a.v}, evaluates it and replaces it with \vp{b}. The second run (line~\ref{line:optwval2}) does the same for point \vp{b}. The evaluation of [\vp{b.v}] is \vp{null} so nothing is put back onto the PS.
\begin{lstlisting}
def a,b
twine a>b
ps a,opVAL
run(*\label{line:optwval1}*)
show ps
(*\vout{ps: b}*)
twine b>null
ps b,opVAL
run(*\label{line:optwval2}*)
show ps
(*\vout{ps: *empty*}*)
\end{lstlisting}
\end{labeledpar}
\subsection{Registers}\label{secregisters}
\begin{labeledpar}{rEVAL}{rCTP}
The current time point.
\end{labeledpar}
\begin{labeledpar}{rEVAL}{rNEW}
Holds the new point created by the last opNEW opcode.
\end{labeledpar}
\begin{labeledpar}{rEVAL}{rEVAL}
Holds the value point of the last opEVAL opcode.
\end{labeledpar}
\begin{labeledpar}{rEVAL}{rWM1}
Holds the value point of the last point loaded into working memory \#1 with the opLWM1 opcode.
\end{labeledpar}
\begin{labeledpar}{rEVAL}{rTCTX}
Returns the time point in the twine context set.
\end{labeledpar}
\begin{labeledpar}{rEVAL}{rT}
Holds the last thought point created by opNEWT.
\end{labeledpar}
\subsection{Reserved/Predefined Points}
\begin{labeledpar}{determiner}{\vp{asPH}}The AS placeholder point appended to the end of the AS with the opISATOAS instruction (page~\pageref{def-asPH}).\end{labeledpar}
\begin{labeledpar}{determiner}{\vp{eoa}}
The end-of-arguments point. It is used by various opcodes to indicate the end of a list of arguments in the PS.
\end{labeledpar}
\begin{labeledpar}{determiner}{\vp{grounded}}
The twine \vp{grounded}<\vp{p} indicates that the point \vp{p} is grounded, i.e. it has is-a points that are sensory.
\end{labeledpar}
\begin{labeledpar}{determiner}{\vp{determiner}}
The twine \vp{determiner}<\vp{p} indicates that the point \vp{p} is a noun phrase determiner (e.g. `a', `the').
\end{labeledpar}
\begin{labeledpar}{determiner}{\vp{surAction}}
This point is within the opLUPARSE instruction. The evaluation of [\vp{surAction}] determines whether or not two contiguous surrogate points are joined or linked.
\end{labeledpar}
\begin{labeledpar}{determiner}{\vp{letter}}
The letters of the alphabet are predefined within V5 as `a', `b', \DOTS, `z'. Each of the letters is twined to the point letter: twine \vp{letter}<`a',`b',\DOTS,`z'.
\end{labeledpar}\begin{labeledpar}{determiner}{\vp{null}}
The \vp{null} point is used to represent no point.  It is also used to indicate no action when the single value of a value twine (\textit{p}>\vp{null}).
\end{labeledpar}
\begin{labeledpar}{determiner}{\vp{evalFail}}
This point is added to the PS when the opEVAL operation fails.
\end{labeledpar}
\begin{labeledpar}{determiner}{\vp{rasFail}}
This point is added to the PS when the opRAS/opRASM operation fails.
\end{labeledpar}
\begin{labeledpar}{determiner}{\vp{parse}}
This point is inserted into the TC so that is-a twines created as a result of parsing natural language include this point (to distinguish them from is-a points actually describing the surrogate).
\end{labeledpar}
\newpage
\section{Initial Data for Pattern Recognition Example}\label{datapatrecex}
\begin{enumerate}
\setlength\itemsep{-.25em} \small
\item \{002 068 082 223 263 295 422 435 489 493 615 672 724 743 744 812 860 887 912 942\}
\item \{055 061 062 083 113 205 288 353 476 483 494 520 648 673 780 800 805 938 965 983\}
\item \{011 018 146 189 209 271 300 320 477 536 577 578 598 661 748 775 882 909 939 984\}
\item \{121 226 246 296 304 518 621 647 711 739 760 782 819 833 878 925 946 963 980\}
\item \{056 098 137 152 174 184 329 369 372 376 445 477 491 492 568 578 634 846 866 975\}
\item \{007 032 208 228 235 275 276 303 331 344 376 401 411 440 478 571 590 711 805 893\}
\item \{035 039 098 116 131 146 211 231 269 285 471 480 481 518 554 576 620 736 790\}
\item \{014 032 147 269 316 325 350 499 537 619 628 764 816 825 833 834 864 889 982 999\}
\item \{016 045 058 073 136 201 370 420 472 492 543 550 562 711 746 838 938 941 948 972\}
\item \{061 117 167 220 257 264 356 362 535 554 595 625 712 751 757 781 804 832 885 994\}
\item \{009 012 048 105 189 264 285 321 346 477 538 780 783 800 814 816 875 879 919 920\}
\item \{007 031 047 069 085 144 337 420 450 467 546 548 583 604 662 666 688 801 822 857\}
\item \{042 090 220 260 268 377 417 457 472 516 517 566 590 630 650 776 816 897 944\}
\item \{086 124 125 169 259 276 283 298 307 326 432 457 593 605 617 649 724 805 946\}
\item \{006 064 085 156 160 170 212 231 300 371 546 572 638 651 661 682 879 883 900 987\}
\item \{013 088 177 218 265 272 275 333 409 495 608 621 710 795 850 873 887 889 933\}
\item \{004 145 180 322 330 347 512 568 615 623 657 685 709 750 766 799 805 899 947 982\}
\item \{014 087 101 109 110 113 228 248 298 312 313 338 485 564 586 652 725 799 840 877\}
\item \{042 043 101 211 307 326 350 352 401 418 443 457 503 528 563 591 601 803 849\}
\item \{067 101 169 211 225 250 264 307 401 439 503 542 548 601 638 668 709 738 880 944\}
\item \{007 008 018 027 123 136 152 162 188 214 216 220 253 325 711 734 759 843 871 956\}
\item \{190 267 313 340 352 367 394 417 489 497 593 595 726 734 853 870 882 905 920 967\}
\item \{000 016 071 091 097 151 156 193 213 223 309 521 545 569 571 672 723 763 765 996\}
\item \{066 111 147 203 270 311 333 337 452 455 475 492 521 575 641 653 655 859 960\}
\item \{101 192 211 282 307 317 401 425 503 506 528 548 563 567 601 656 664 691 825 940\}
\item \{028 032 114 130 132 133 136 184 262 406 452 460 498 619 643 739 742 852 915 997\}
\item \{031 102 105 135 138 147 198 204 374 408 539 547 564 584 586 784 792 808 917 995\}
\item \{001 123 137 146 211 244 250 285 308 563 600 612 632 677 694 777 799 875 886 992\}
\item \{081 101 158 196 211 277 307 309 401 430 483 503 599 601 626 653 727 749 819 885\}
\item \{002 051 060 121 180 246 267 369 411 450 469 541 582 645 647 722 757 812 866 903\}
\item \{012 048 078 146 332 425 527 567 646 658 693 718 723 747 809 813 834 866 982\}
\item \{008 033 103 218 224 275 307 330 337 486 495 512 523 553 583 636 702 707 909 928\}
\item \{028 045 101 160 190 211 215 266 307 385 401 503 529 591 601 608 751 815 881 893\}
\item \{027 064 191 219 246 319 323 368 379 444 539 623 691 827 870 884 896 903 936 954\}
\item \{064 073 087 132 213 300 322 339 409 434 447 487 510 535 620 684 852 875 952\}
\item \{047 178 197 232 359 384 460 481 627 634 702 707 830 839 879 888 891 898 955 968\}
\item \{030 055 087 161 223 385 391 409 515 570 615 689 726 761 814 820 834 845 922 965\}
\item \{041 067 115 179 188 212 314 361 405 636 669 705 757 759 770 787 841 846 952 953\}
\item \{018 059 076 193 243 315 377 419 485 488 560 606 723 758 788 870 908 931 941\}
\item \{171 189 219 277 294 332 335 336 358 421 426 430 489 589 738 787 843 850 933 998\}
\end{enumerate}
\newpage
\section{Parsing Speech using the opRASM Instruction}\label{secParseopRASM}
A few examples of parsing with speech with the opRASM are presented in this appendix. The capabilities of this approach go beyond those of the LTT process. The abbreviations (points) used in these examples for the different parts of speech are described in figure~\ref{figpartsofspeech}.
\begin{figure}[H]
\centering
\begin{tabularx}{\linewidth}{ c L | c L }
\toprule
adj & adjective & prep & preposition \\
artD & article definite & sub & verb subject \\
dnp & double noun phrase (for nested clauses) & tspec & time specification (e.g. today, now) \\
gerundI & gerund intransitive & tvobj & transitive verb object \\
infP & infinitive prefix (e.g. 'to' as in 'to go') & vpI & verb phrase intransitive \\
noun & noun & vpT & verb phrase transitive \\
np & noun phrase & verb2b & verb to-be (e.g. is, was) \\
obj & verb object & vtwo & verb phrase transitive with object \\
pnoun & proper noun & vtws & verb phrase transitive with subject \\
\bottomrule
\end{tabularx}
\caption{Parts of speech codes to be used in the following examples}\label{figpartsofspeech}
\end{figure}
\begin{exDesc}
The first section of this example defines the relationships of words to parts of speech. Note that some of the words can have multiple syntactic meanings (lines~\ref{line:exmultmean1}, \ref{line:exmultmean2}, \ref{line:exmultmean3}).
\end{exDesc}
\begin{lstlisting}
twine artD<the ; noun<man,ball,tuxedo ; prep<to ; infP<to(*\label{line:exmultmean1}*)
twine prep<in ; pnoun<john,mary ; adj<red ; pnoun<mary
twine noun<ball ; verbT<ball,threw ; prep<threw(*\label{line:exmultmean2}*)
twine noun<question ; pronoun<you ; noun<saw ; verb2b<is
twine gerundI<going ; tspec<today ; verbT<saw ; tspec<yesterday
twine pnoun<google ; verbT<google ; person<nounMan(*\label{line:exmultmean3}*)
twine clothes<nounTuxedo

bind [john pnoun] pnounJohn
bind [mary pnoun] pnounMary
bind [ball noun] nounBall
bind [man noun] nounMan
bind [tuxedo noun] nounTuxedo
bind [question noun] nounQuestion
bind [adj red] adjRed
bind [verbT threw] verbTthrow
bind [verbT google] verbTGoogle
bind [pnoun google] pnounGoogle
bind [prep to] prepTo
bind [prep in] prepIn
\end{lstlisting}
\begin{exDesc}
The next section of the example defines the rules for parsing a string of words via the AS reduction instruction (opRASM).
\end{exDesc}
\begin{lstlisting}[firstnumber=21]
bind +2 [-artD/-noun] opNEWSUR,rNEW,np,opTWISA,eoa,noun,noun.v,opEVAL,rNEW,
	rEVAL,opTWISA,rNEW
bind +2 [-artD/-adj] opNEWSUR,rNEW,np,opTWISA,eoa,adj,adj.v,opEVAL,rNEW,rEVAL,
	opTWISA,rNEW
bind +2 [-np/-noun] eoa,noun,noun.v,opEVAL,np.v,rEVAL,opTWISA,np.v
bind [-np/pronoun] dnp
bind [-np/-verbT] vtws
bind [vtws/-tspec] null
bind [dnp/-vtws] null
bind [-verb2b/-gerundI] opNEWSUR,rNEW,gerundI.v,opTWISA,vpI
bind [-dnp/-vpI] viws
bind [-dnp/-vpT] vtws
bind [-dnp/-verbT] vtws 
bind +1 [-verbT/-np] opNEWSUR,eoa,verbT,verbT.v,opEVAL,rNEW,rEVAL,opTWISA, np.v,obj,
	opTWISA, rNEW,tvobj,opTWISA,vtwo
bind +1 [-infP/-vtwo] null
bind [-vtws/-vpI] null
bind [-pnoun/-vpI] viws
bind [-prep/-pnoun] opNEWSUR,rNEW,prep,opTWISA,eoa,prep,prep.v,opEVALt,rNEW,rEVAL,
	opTWISA,eoa,pnoun,pnoun.v,opEVALt,rNEW,rEVAL,opTWISA,null
bind [-prep/-np] np.v,prep.v,opTWISA,np.v
bind +2 [-prep/-clothes] np.v,prepToWear,opTWISA,np.v
bind [-viws/-tspec] null
bind [-pnoun/-vtwo] opNEWSUR,rNEW,sub,opTWISA,eoa,pnoun.v,pnoun,opEVAL,rNEW,
	rEVAL,opTWISA,null
bind [-np/-vtwo] np.v,sub,opTWISA,np.v
bind [-np/-vpI] np.v,sub,opTWISA,null
bind [-person/-prepToWear] person.v,prepToWear.v,opTWISA,null
\end{lstlisting}
\begin{exDesc}
The first two examples demonstrate how the contextual evaluation selection of the highest binding weight can be used to disambiguate words. In the first example, the word `google' in the phrase `to google' is taken as a verb. The parse tree generated for the sentence `The man is going to google the question' is shown in figure~\ref{figgooglequestion}.
\end{exDesc}
\begin{lstlisting}[firstnumber=44]
ps the,man,is,going,to,google,the,question,eoa
run opLASM
trace opTWISA
run opRASM
\end{lstlisting}
\begin{figure}[H]
\centering
\begin{tikzpicture}[scale=0.8, grow'=up]
\Tree [.\vp{t_x} [.\vp{s_1} \vp{np} \vp{nounMan} \vp{sub} ] [.\vp{s_2} \vp{np} \vp{nounQuestion} \vp{obj} ] [.\vp{s_3} \vp{verbTGoogle} \vp{tvObj} ] ]
\end{tikzpicture}
\caption{Parsing of `The man is going to google the question'}\label{figgooglequestion}
\end{figure}
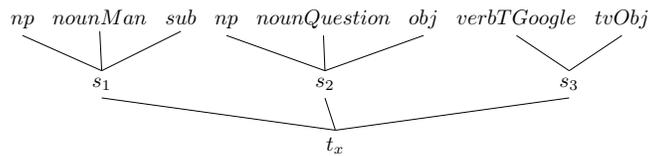
\begin{exDesc}
In the second example, the word `google' in the phrase `to google' is taken as a noun (place). The parse tree for the sentence `The man is going to google' is shown in figure~\ref{figgotogoogle}.
\end{exDesc}
\begin{lstlisting}[firstnumber=48]
ps the,man,is,going,to,google,eoa
run opLASM
trace opTWISA
run opRASM
\end{lstlisting}
\begin{figure}[H]
\centering
\begin{tikzpicture}[scale=0.8, grow'=up]
\Tree [.\vp{t_x} [.\vp{s_1} \vp{np} \vp{nounMan} \vp{sub} ] [.\vp{s_2} \vp{going} ] [.\vp{s_3} \vp{prepTo} \vp{pnounGoogle} ] ]
\end{tikzpicture}
\caption{Parse tree for `The man is going to google'}\label{figgotogoogle}
\end{figure}
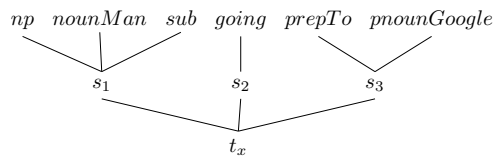
\begin{exDesc}
The last example demonstrates how semantic knowledge can be utilized to resolve a dangling prepositional phrase. In the sentence `The man threw the ball in the tuxedo' the phrase `in the tuxedo' is associated with the subject (`the man'). The resulting parse tree is shown in figure~\ref{figmantux}.
\end{exDesc}
\begin{lstlisting}[firstnumber=52]
ps the,man,threw,the,ball,in,the,tuxedo,eoa
run opLASM
trace opTWISA
run opRASM
\end{lstlisting}
\begin{figure}[H]
\centering
\begin{tikzpicture}[scale=0.8, grow'=up]
\Tree [.\vp{t_x} [.\vp{s_1} \vp{np} \vp{nounMan} \vp{sub} [.\vp{s_3} \vp{nounTuxedo} \vp{prepToWear} ] ] [.\vp{s_2} \vp{np} \vp{nounBall} \vp{obj} ] [.\vp{s_4} \vp{verbTthrow} ] ]
\end{tikzpicture}
\caption{Parse tree for `The man threw the ball in the tuxedo'}\label{figmantux}
\end{figure}
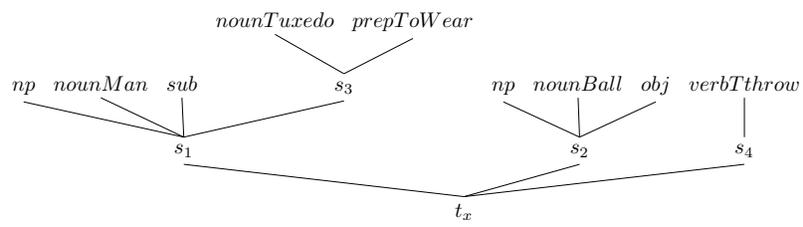

\end{document}